\documentclass[lettersize,journal]{IEEEtran}
\usepackage{tikz}
\usepackage{everypage}
\AddThispageHook{%
  \begin{tikzpicture}[remember picture,overlay]
    \node[above left, inner sep=0pt] at
         ([xshift=-1.35cm,yshift=-1.5cm]current page.north east) 
         {\includegraphics[height=1cm]{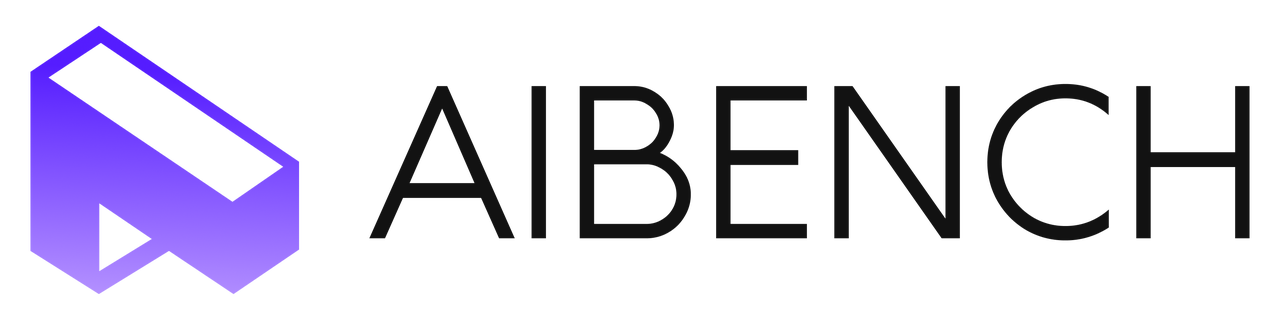}};             
  \end{tikzpicture}%
}
\usepackage{amsmath,amsfonts}
\usepackage{algorithmic}
\usepackage{algorithm}
\usepackage{array}
\usepackage{textcomp}
\usepackage{stfloats}
\usepackage{url}
\usepackage{verbatim}
\usepackage{graphicx}
\usepackage{cite}
\usepackage{multirow,subfigure}
\usepackage{color}
\usepackage{booktabs}
\usepackage{arydshln}
\usepackage{caption}
\usepackage{lipsum}
\usepackage{float}
\usepackage{hyperref,MnSymbol}
\usepackage{xcolor}
\usepackage{fontawesome}
\usepackage{lipsum}
\usepackage{pgf} 
\usepackage{pgffor}
\usepackage{multicol}
\usepackage{calc}
\usepackage{colortbl}

\usepackage{cuted}

\usepackage{subcaption}

\usepackage{kantlipsum}

\usepackage{etoolbox}

\let\OLDthebibliography\thebibliography
\renewcommand\thebibliography[1]{
  \OLDthebibliography{#1}
  \setlength{\parskip}{0pt}
  \setlength{\itemsep}{0pt plus 0.3ex}
}

\definecolor{lightgray}{gray}{0.9}

\newcommand{\CLA}[1]{{\color[HTML]{4472c4} \textbf{#1}}}
\newcommand{\CLB}[1]{{\color[HTML]{E76254} \textbf{#1}}}

\newcommand{\GroupA}[1]{{\color[HTML]{92ca2c} \textbf{#1}}}
\newcommand{\GroupB}[1]{{\color[HTML]{3c7daa} \textbf{#1}}}
\newcommand{\GroupC}[1]{{\color[HTML]{e64f04} \textbf{#1}}}
\newcommand{\GroupD}[1]{{\color[HTML]{f2c400} \textbf{#1}}}
\newcommand{\GroupE}[1]{{\color[HTML]{a757a7} \textbf{#1}}}

\begin{document}

\title{Using GUI Agent for Electronic Design Automation}

\author{Chunyi Li, Longfei Li, Zicheng Zhang, Xiaohong Liu, Min Tang,~\IEEEmembership{Senior Member,~IEEE} \\ 
Weisi Lin,~\IEEEmembership{Fellow,~IEEE}, 
Guangtao Zhai,~\IEEEmembership{Fellow,~IEEE}
\thanks{The work was supported by the National Natural Science Foundation of China under Grants 625B2118, 62225112, 62301310, 62572317, 623B2073, in part by the Singapore Ministry of Education under Grant ZDSYS20220527171406015. Chunyi Li and Longfei Li contributed equally to this work. Corresponding author: Min Tang, Guangtao Zhai.}
\thanks{Chunyi Li and Weisi Lin are with the College of Computing and Data Science, Nanyang Technological University, Singapore 639798, Singapore (email: lich0076@e.ntu.edu.sg, wslin@ntu.edu.sg)}
\thanks{Zicheng Zhang and Guangtao Zhai are with the Center of AI Evaluation, Shanghai AI Laboratory, Shanghai 200232, China (email: {zhangzicheng, zhaiguangtao}@pjlab.org.cn)}
\thanks{Longfei Li and Min Tang are with the School of Integrated Circuit Design, Shanghai Jiao Tong University, Shanghai 200240, China (email: {longfeili, tm222}@sjtu.edu.cn)}
\thanks{Xiaohong Liu is with the John Hopcroft Center, Shanghai Jiao Tong University, Shanghai 200240, China (email: xiaohongliu@sjtu.edu.cn)}

}

\markboth{Journal of \LaTeX\ Class Files,~Vol.~1, No.~1, Feb~2024}%
{Shell \MakeLowercase{\textit{et al.}}: A Sample Article Using IEEEtran.cls for IEEE Journals}


\maketitle

\begin{abstract}
Graphical User Interface (GUI) agents adopt an end-to-end paradigm that maps a screenshot to an action sequence, thereby automating repetitive tasks in virtual environments. However, existing GUI agents are evaluated almost exclusively on commodity software such as Microsoft Word and Excel. Professional Computer-Aided Design (CAD) suites promise an order-of-magnitude higher economic return, yet remain the weakest performance domain for existing agents and are still far from replacing expert Electronic-Design-Automation (EDA) engineers. We therefore present the first systematic study that deploys GUI agents for EDA workflows. Our contributions are: (1) a large-scale dataset named GUI-EDA, including 5 CAD tools and 5 physical domains, comprising 2,000+ high-quality screenshot-answer-action pairs recorded by EDA scientists and engineers during real-world component design; (2) a comprehensive benchmark that evaluates 30+ mainstream GUI agents, demonstrating that EDA tasks constitute a major, unsolved challenge; and (3) an EDA-specialized metric named EDAgent, equipped with a reflection mechanism that achieves reliable performance on industrial CAD software and, for the first time, outperforms Ph.D. students majored in Electrical Engineering. This work extends GUI agents from generic office automation to specialized, high-value engineering domains and offers a new avenue for advancing EDA productivity. The dataset will be released at: https://github.com/aiben-ch/GUI-EDA.
\end{abstract}



\section{Introduction}

The rapid evolution of Multimodal Large Language Models (MLLMs) now spans Text-to-Text (e.g., DeepSeek \cite{intro:deepseek}, InternLM \cite{intro:internlm}), Image-to-Text (e.g., Qwen-VL \cite{intro:qwen}, InternVL \cite{intro:internvl}), Text-to-Image (e.g., SDXL \cite{intro:xl}, DALL-E \cite{intro:dalle}), and Text-to-Video (e.g., Sora \cite{intro:sora}). While these models are widely deployed for creative tasks such as poetry generation and film synthesis, users frequently voice the concern: `I want AI to do my laundry and dishes so that I can do art and writing, not vice versa.'
Therefore, the paradigm of MLLM-as-Agent has emerged. Its objective is to delegate labor-intensive, repetitive work to autonomous agents operating in virtual environments, thereby liberating human productivity. 

\begin{figure}[t]
    \centering
    \includegraphics[width=\linewidth]{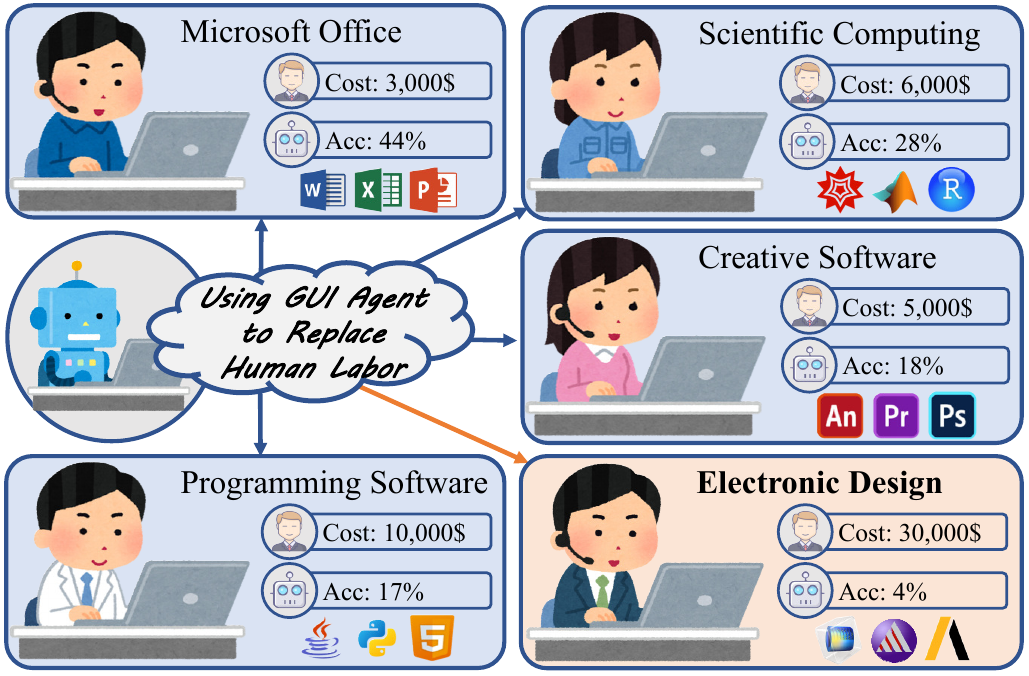}
    \caption{The accuracy of using GUI Agent to complete various tasks, and the monthly cost of hiring corresponding human labor. Electronic Design Automation (EDA) engineers have the highest cost, and the current Agent performs the worst, highlighting the significance of an EDA-specific GUI Agent.}
    \label{fig:spotlight}
    \vspace{-5mm}
\end{figure}

Such agents translate the decisions of an MLLM into concrete actions and interact with external tools—APIs, databases, or physical devices. Existing toolsets, however, exhibit clear limitations: for example, an agent that merely outputs Python code cannot initiate low-level network requests to send an e-mail on behalf of the user.
Because humans solve most digital tasks via vision alone, an end-to-end paradigm that maps a Graphical User Interface (GUI) directly to interaction commands has the potential to subsume all agent tasks. Consequently, the GUI Agent is a critical route toward Artificial General Intelligence (AGI) and promises substantial economic impact by automating assembly-line occupations.
For any concrete task, an effective GUI Agent must satisfy two criteria:
\begin{itemize}
    \item Usability: high success rate when the agent is deployed.
    \item Utility: high human-labor cost for the task above.
\end{itemize}
Figure \ref{fig:spotlight} summarizes the prospective return of mainstream GUI Agents across various applications. Success rates are drawn from ScreenSpot-Pro \cite{relate:screenspotpro}, and labor costs are taken from the U.S. occupational wage \cite{intro:salary} medians. The results indicate that current GUI Agents perform well on common software, yet the corresponding human labor is inexpensive. Substituting high-wage occupations would multiply the economic benefit.

Guided by these principles, \textbf{Electronic Design Automation (EDA) should be a separate branch beyond general GUI Agent tasks.}
For utility, a Computer-Aided Design (CAD) engineer costs roughly ten times more than an Office user, implying a strong demand for an automatic paradigm. However, for usability, GUI Agents achieve only a 4\% success rate on CAD workflows. Multiple GUI-Agent benchmarks corroborate that CAD consistently exhibits the lowest success rate of all tested domains, trailing far behind Web pages, MATLAB, and Photoshop. Therefore, CAD is both the most challenging and the most economically valuable scenario, representing the critical requirement that future GUI Agents must address.

\begin{table*}[t]
\centering
    \renewcommand\arraystretch{1.25}
    \renewcommand\tabcolsep{8pt}
    \belowrulesep=0pt\aboverulesep=0pt
    \caption{Previous GUI Agent benchmarks for general tasks and the GUI-EDA we proposed specifically for EDA tasks.}
    \label{tab:relate}
    \vspace{-5pt}
    \resizebox{\linewidth}{!}{
    \begin{tabular}{llrrrcc}
    \toprule
\multicolumn{1}{c}{Dataset} & \multicolumn{1}{c}{Source} & \multicolumn{1}{c}{Size} & \multicolumn{1}{c}{Resolution} & \multicolumn{1}{c}{CAD Task} & Labels           & Validation           \\ \midrule
Mind2Web \cite{relate:mind2web}    & WebCanvas 2023             & 2,350    & 768p $\sim$1080p               & 0\%               & Action         &  \\
Mind2Web-Live \cite{relate:mind2weblive}              & WebCanvas 2024             & 542  & 768p $\sim$1080p               & 0\%               & Action         & OS              \\
Multimodal-Mind2Web \cite{relate:Mind2Web-Multimodal}        & OSU-NLP 2024               & 2,350    & 1080p      & 0\%               & Action         &  \\
Explorer-Web  \cite{relate:Explorer-Web}              & MSR-OSU 2025               & 94,000   & 1080p      & 0\%               & Answer, Action  & OS               \\
ScreenSpot \cite{relate:Screenspot}  & NJU 2024   & 1,200    & 1080p      & 0\%               & Action         & \multicolumn{1}{l}{} \\
TongUI \cite{relate:tongui}  & BIGAI 2025                 & 1,430    & 1080p-2k & 0\%               & Answer, Action & OS               \\
GUI-Reflection \cite{relate:gui-reflection}             & NTU-MMLab 2025             & 63,353   & 1080p-4k & 0\%               & Answer, Action & OS              \\
OS-World \cite{relate:osworld}   & HKU-NLP 2024               & 412  & 1080p      & 13.30\%           & Answer, Action  & OS               \\
ScreenSpot-Pro \cite{relate:screenspotpro}             & NUS-Next++ 2025            & 1,581    & 1080p $\sim$4k                 & 19.11\%           & Action         &  \\
MMBench-GUI \cite{relate:mmbenchgui}                & SH-AILab 2025              & 1,536    & 1080p $\sim$2k                 & 17.64\%           & Answer, Action  & OS              \\
\textbf{GUI-EDA}     & \textbf{Ours}   & \textbf{2,082}    & \textbf{320p $\sim$4k}  & \textbf{100\%}             & \textbf{Answer, Action}  & \textbf{Real-World}  \\ \bottomrule        
\end{tabular}
}
    \vspace{-3mm}
\end{table*}

However, addressing EDA tasks with GUI Agents confronts two fundamental challenges. First, EDA demands domain expertise rather than common-sense knowledge. Every device design invokes multiple physical domains—thermal conduction, electromagnetic induction, optical reflection—requiring integrated conceptual comprehension. Second, the GUI of EDA-grade CAD software is arranged according to conventions that diverge sharply from everyday applications. Even when an agent has correctly inferred the required action, executing it still hinges on the capability to locate and activate the precise control element. As neuroscience distinguishes between System 1 (fast, execution-oriented) and System 2 (slow, comprehension-oriented) cognition, a GUI Agent must simultaneously master both faculties—an inherent dilemma. These factors jointly account for the pronounced under-performance of current GUI Agents on EDA tasks.
Therefore, we conducted the first comprehensive investigation on employing GUI Agent for EDA, aiming to promote the application of agentic intelligence in this challenging field and enable GUI Agents to better replace human labor. Our contributions can be summarized as follows:

\begin{itemize}
    \item Dataset construction. We introduce GUI-EDA, the first large-scale benchmark for GUI Agents in EDA, including 5 physical fields, 5 industry-standard CAD softwares, rendered at multiple resolutions. Guided by fine-grained labels from certified CAD engineers, the dataset establishes precise optimization targets for future agents.
    \item Empirical evaluation. We benchmark 20+ existing agents, including general-purpose MLLMs and specialized GUI Agents on EDA tasks, decomposing performance into comprehension and execution dimensions. Experiment shows none of the agents can reliably solve EDA tasks.
    \item Methodology implementation. We propose EDAgent, a GUI agent tailored to EDA. By integrating comprehension and execution within a unified framework, EDAgent raises execution accuracy on GUI-EDA by 16\% that surpassing human-expert for the first time, demonstrating strong practical usage in the EDA application.
\end{itemize}

\section{Related Works}

\subsection{AI Agent for EDA}

Early attempts to introduce artificial intelligence into EDA have pursued three dominant interaction paradigms.

Natural-language: Recent Large-Language-Model (LLM) works let users describe a circuit in free English and then generate Verilog code or instruction directly.  ChipNeMo, RTLLM, VerilogEval, ChipGPT and VeriGen \cite{eda:chipnemo,eda:rtllm,eda:verilogeval,eda:chipgpt,eda:2023verigen} show that fine-tuned LLMs can reach 90\% syntax correctness, but still suffer from 75\% functional errors even on small HDLBits benchmarks.  Because language is ambiguous with respect to clocking, reset, power intent, etc., the resulting Register Transfer Level (RTL) almost always needs manual repair.

Script-level assistance: Instead of RTL, many tools target the control layer—Tcl/Python scripts that drive commercial EDA flows.  ChatEDA \cite{eda:chateda} and the EDA-script experiments in ChipNeMo \cite{eda:chipnemo} demonstrate that LLMs can autogenerate placement constraints, synthesis directives or Innovus scripts. Although elegant, the agent issues shell commands that still bypass the graphical cockpit in which human designers actually debug (e.g., highlighting a hot-spot in a congestion map). When the script fails, the user must mentally reconstruct what the black-box command sequence would have looked like on the screen—an error-prone reverse-engineering task.

Open-ended LLM-as-Agent: Inspired by AutoGPT-style loops, the lateset works wrap an LLM inside a ReAct agent that can call EDA tools, parse their logs and self-correct.  AutoChip, RTLFixer and VerilogReader \cite{eda:autochip,eda:rtlfixer,eda:verilogreader} chain compilation, simulation and formal checks in a while-loop until the code passes.  However, these agents are used in different scenarios, such as RTL, synthesis, and physical design. They target different CAD software and require different physical considerations for components. Therefore, these specialized agents lack the generalizability to handle multiple EDA tasks.

Across all three paradigms the visual interface is treated as an after-thought: either it is completely ignored, or it is accessed only through textual scripts rather than original images.  Consequently, designers still spend most of their time pointing, clicking, zooming and cross-probing to validate what the AI has produced. To the best of our knowledge, no prior EDA work has trained an agent that perceives the pixel stream of the commercial GUI and produces mouse-keyboard actions exactly like a human would. Since images are the golden representation of external information, we therefore advocate a fourth paradigm—an EDA-specific GUI agent that reasons over screenshots, clicks on menus, and thereby stays implicitly consistent with the ever-changing visual state of the platform.

\begin{figure*}[t]
    \centering
    \includegraphics[width=\linewidth]{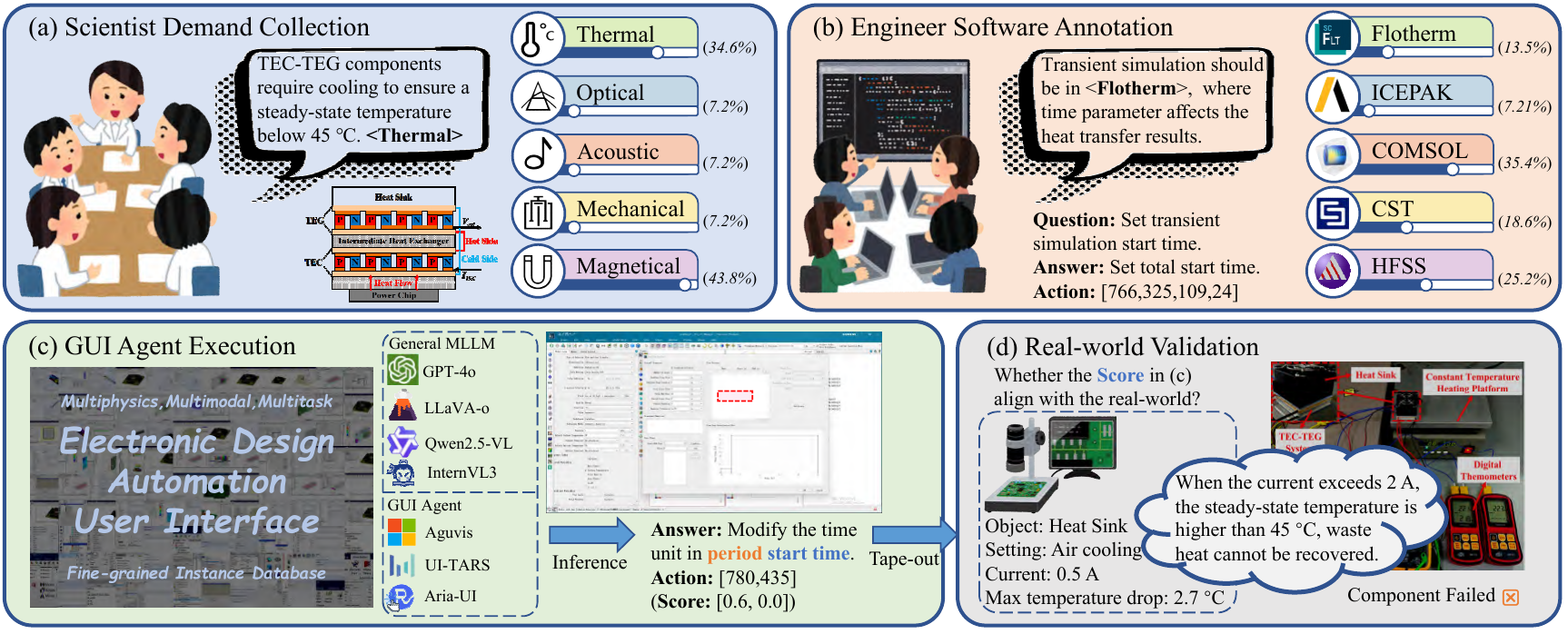}
    \caption{The construction of GUI-EDA benchmark. Scientists raise questions in real EDA tasks and then solved by engineers using the corresponding software. In virtual CAD software, the performance is evaluated by comparing the differences between the solution from engineers and GUI Agents, and validated by constructing electronic components in the Real-world.}
    \label{fig:dataset}
    \vspace{-2mm}
\end{figure*}

\subsection{GUI Agent}

Despite the explosive growth of GUI Agents, Table \ref{tab:relate} reveals three fundamental gaps that keep them outsiders in professional EDA workflows.
First, there is almost no real EDA task. Mainstream benchmarks such as ScreenSpot \cite{relate:Screenspot} operate entirely on lightweight suites where professional software is simply absent. Though few corpora takes EDA into consideration, like OS-World \cite{relate:osworld} and MMBench-GUI \cite{relate:mmbenchgui}, they still devote less than 20\% of their tasks. Moreover, even those tasks are only operated by COMSOL Multiphysics, while the sign-off tools that dominate CAD cycles (CST, HFSS) remain untouched. 
Some leading GUI benchmarks \cite{relate:scienceboard,relate:cerebrum,relate:computeruse,relate:guiworld,relate:gui-actor} consider professional software, usually in Astronomy, Algebra, Biology, Chemistry, and Geography tasks. Though it may replace certain human labor, as we analyze in Figure \ref{fig:spotlight}, EDA tasks offers greater finical benefits.
Second, existing datasets suffer from a severe modality bias: most corpora supply only (x, y) click coordinates (Action) and omit natural-language rationales (Answer) that human designers constantly exchange when debugging congestion maps; agents evaluated on such skewed annotation inevitably degenerate into blind clickers instead of articulate collaborators,
where an explicit `Answer' in engineer-style is needed to explain why the parameter is changed and what physical effect is expected.
Finally, validation remains embarrassingly open-loop: success is declared if the predicted widget is clicked.
Some recent work has deployed GUI agents directly in the operating system to interact with the screen for online verification. However, given the specificity of EDA tasks, its success depends on the availability of real design components, rather than simulation results from CAD software.
Thus, an EDA benchmark validated by silicon proof is needed: the same agent that clicks must later produce a tape-out component, which could close the cyber-physical loop that solved the real needs of CAD engineers.

\section{Benchmark Construction}

\subsection{Scientist Demand Collection}
\label{sec:dataset-a}

Our data-collection pipeline proceeds in two stages—(a) scientists specify macro-level objectives and (b) engineers solve the concrete problem—as illustrated in Figure \ref{fig:dataset}.
First, we collect and filter out 7,000+ project reports from EDA courses, laboratory notebooks, and industrial design files. From these documents, we extract fluent, domain-specific, and non-trivial natural-language constraints such as `Heat-sink temperature must remain below 45 °C' or `Transducer sound-pressure level $\leq$ 120 dB'.
Next, scientists assign the dominant physical domains required to satisfy each constraint, providing engineers with an unambiguous design context. The taxonomy comprises five categories: Acoustic, Optical, Mechanical, Electro-Thermal, and Electro-Magnetical.
Since most EDA components are intrinsically Electrical\footnote{The prefix `Electro' for these two fields is omitted in the following text.}, we subdivide the `Electro' class to preserve balance, including Electro-Thermal (e.g., heat generation/dissipation) beyond 40 \% in total, and Electro-Magnetical (e.g., computational blocks) beyond 30 \%. The remaining Acoustic, Optical, and Mechanical are usually off-chip sensor modules, accounting for about 20\%.
We perform stratified sampling over the resulting triplets to assemble a `scientist demand' pool with 1,000 valid instances. Such [requirement, threshold, fields] triplets are both domain-balanced and representative of real-world design constraints, thereby supplying the GUI-EDA dataset with authentic, objective, and high-quality instruction.

\subsection{Engineer Software Annotation}
\label{sec:dataset-b}
\begin{figure}[t]
    \includegraphics[width=\linewidth]{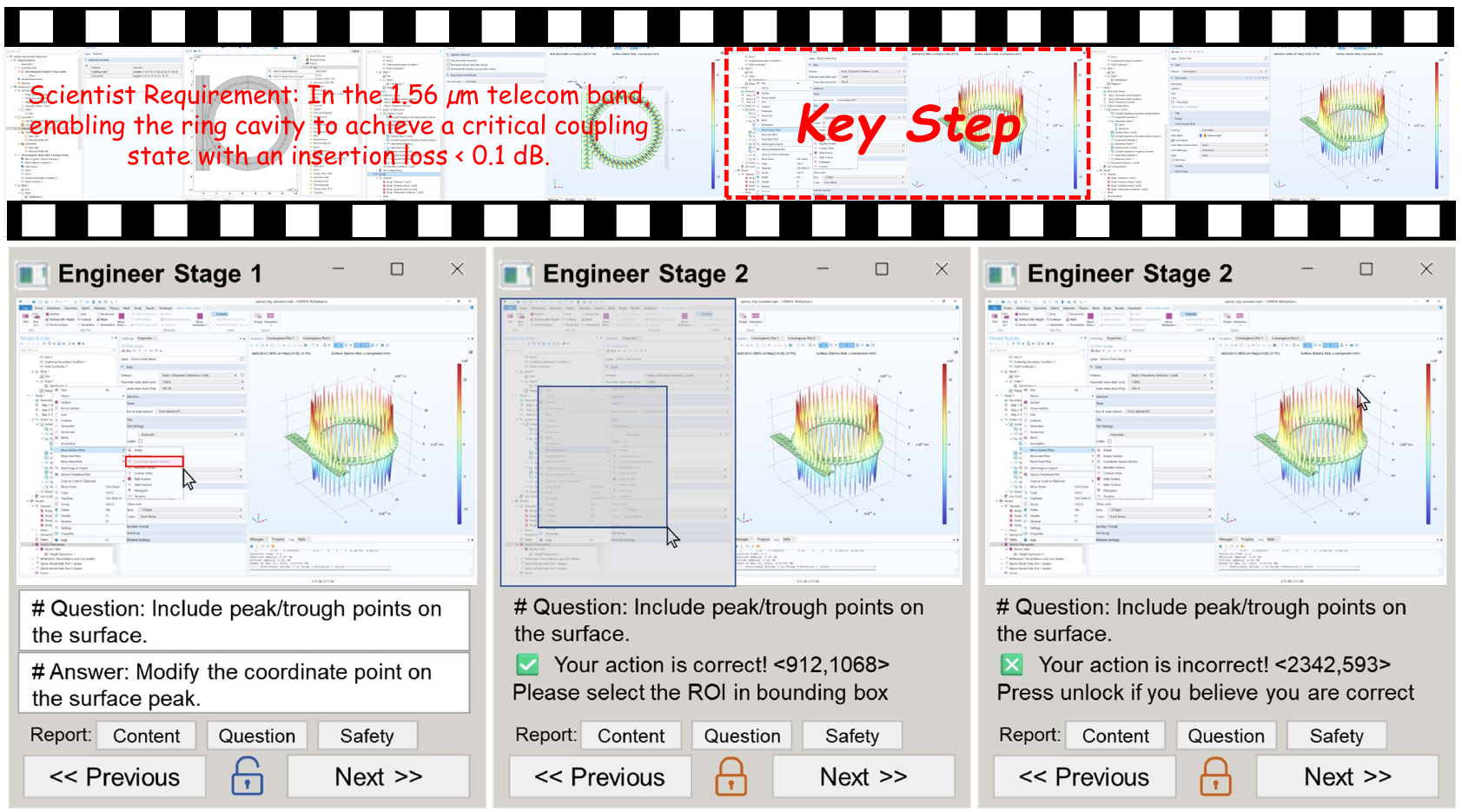}
    \caption{Annotation interface for engineers. After operating a CAD software according to scientist demand, the engineer will perform Stage-1: designing Question, Answer, and Actions for the Key Step; Stage-2: adding fine-grained annotation of valid samples in Stage-1, and discarding controversial samples.}
    \label{fig:interface}
    \vspace{-2mm}
\end{figure}

After the scientist demand pool is established, we invite 10 senior EDA engineers to perform `live' design sessions inside the native CAD software toolkits, producing \textbf{Stage-1} coarse annotations.
For each triplet, the engineer selects the appropriate CAD package, sets parameters, and launches a simulation.
Specifically, we provided 5 CAD software: COMSOL, Flotherm, ICEPAK, CST, and HFSS. For 5 multiphysic fields from scientists,
COMSOL is used for Acoustic, Optical, Mechanical, and a minority of Thermal tasks;
Flotherm and ICEPAK handle the remaining Thermal tasks;
and CST and HFSS cover all Magnetical tasks.
This yields 8 distinct software–field\footnote{Hereafter in the main text, we denote each sample category by the software acronym with field, e.g., CO-Acoustic and Fl-Thermal.} combinations.
If the simulated result meets the specified threshold, the trial is marked successful; the engineer then nominates the single most informative frame of the session as the Key Step.
Concretely, the scientist general requirement is re-cast as the immediate objective of that step, and recorded as the Question.
For this Key Step, the engineer provides Ground Truth (GT) at two granularities:
\begin{itemize}
    \item Action: an acceptable click region as a bounding box;
    \item Answer: a concise natural-language solution description.
\end{itemize}

After one engineer completes the above procedure, a second senior engineer performs a double-check as \textbf{Stage-2} fine-grained annotation.
Specifically, the second engineer must re-execute the action described in the Question on the Key Step; the sample is accepted only if the mouse coordinates fall inside the Action bounding box provided in Stage 1. \textbf{($\romannumeral1$)} If accepted,
motivated by recent GUI-Agent studies \cite{relate:gui-reflection,relate:screenspotpro} showing that operating on a sub-region of the GUI is more reliable than processing the full screen, the engineer further annotates: A sub-region covering roughly 50\% of the GUI area; and a tighter sub-region containing only the option bar.
Together with the original Key Step, these three crops constitute the Large/Middle/Small multi-resolution set, enabling us to systematically investigate resolution sensitivity on CAD software. \textbf{($\romannumeral2$)} 
If the mouse-click coordinates fall outside the pre-defined bounding box, the two annotating engineers must reach a consensus; otherwise the sample is discarded as ambiguous.
In addition, engineers may discard a sample when any of the following conditions hold:

\begin{itemize}
    \item Content mismatch: the software shown cannot solve the assigned physics (e.g., Magnetical task in Flotherm).
    \item Trivial question: the goal is obvious which does not require any image comprehension (e.g., Click the close button in the upper-right corner).
    \item Safety risk: the action would irreversibly damage the project (e.g., Delete all components).
\end{itemize}

In the above two stages, the mouse coordinates are recorded throughout the process by a self-developed logging plug-in, as shown in Figure \ref{fig:interface}. After labeling and verification, we obtain 2,082 valid samples in total, consists of the key Steps [Image, Question, Answer, Action], which serve as the multiphysics, multimodal, and multitask GUI-EDA datasets.


\begin{figure}[t]
    \centering
    \subfigure[Answer (Field correlated)]{
    \begin{minipage}[t]{0.45\linewidth}
    \includegraphics[width=\linewidth]{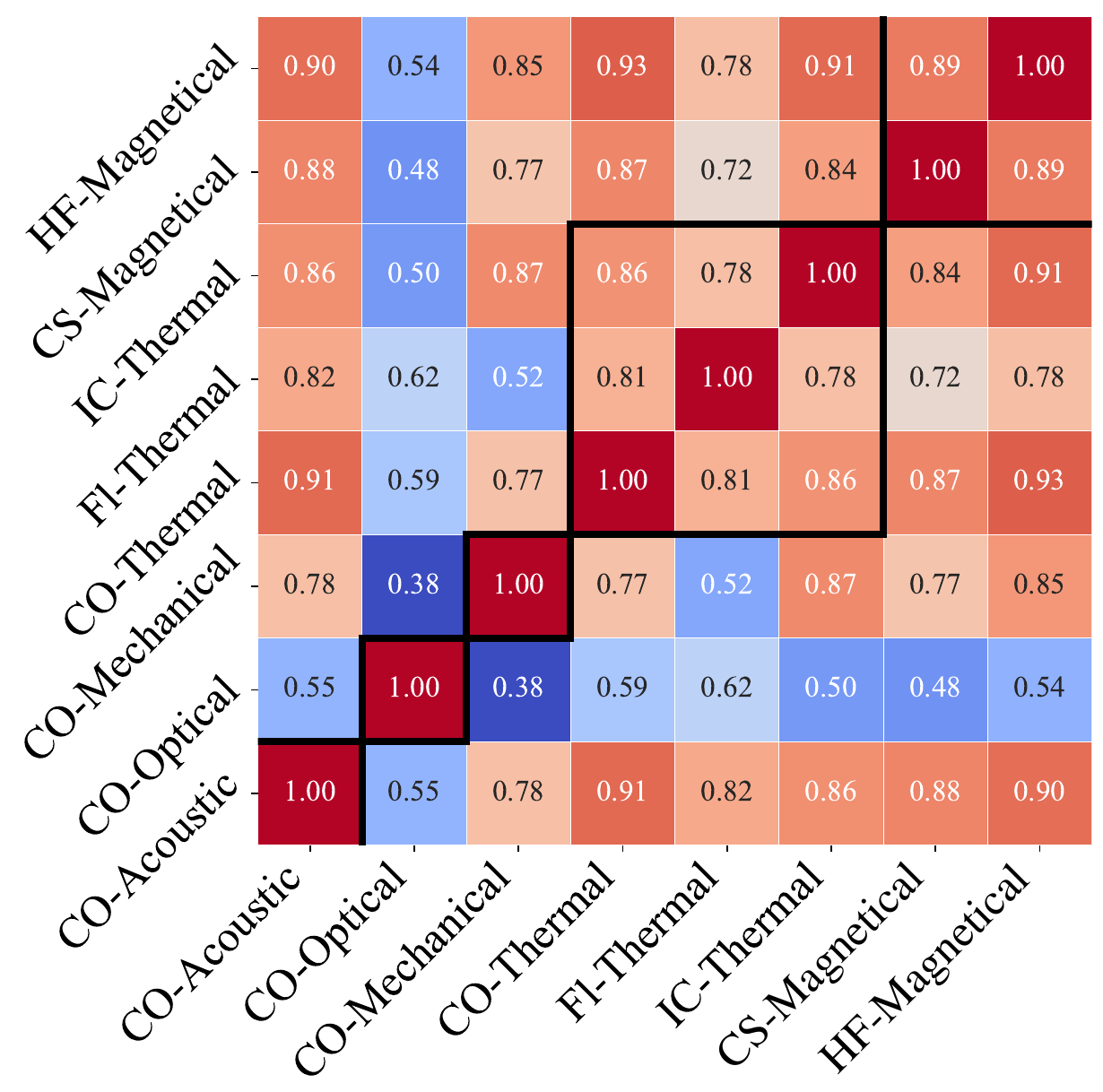}
    \centering
    \vspace{-1mm}
    \end{minipage}%
    }%
    \subfigure[Action (Software correlated)]{
    \begin{minipage}[t]{0.52\linewidth}
    \includegraphics[width=\linewidth]{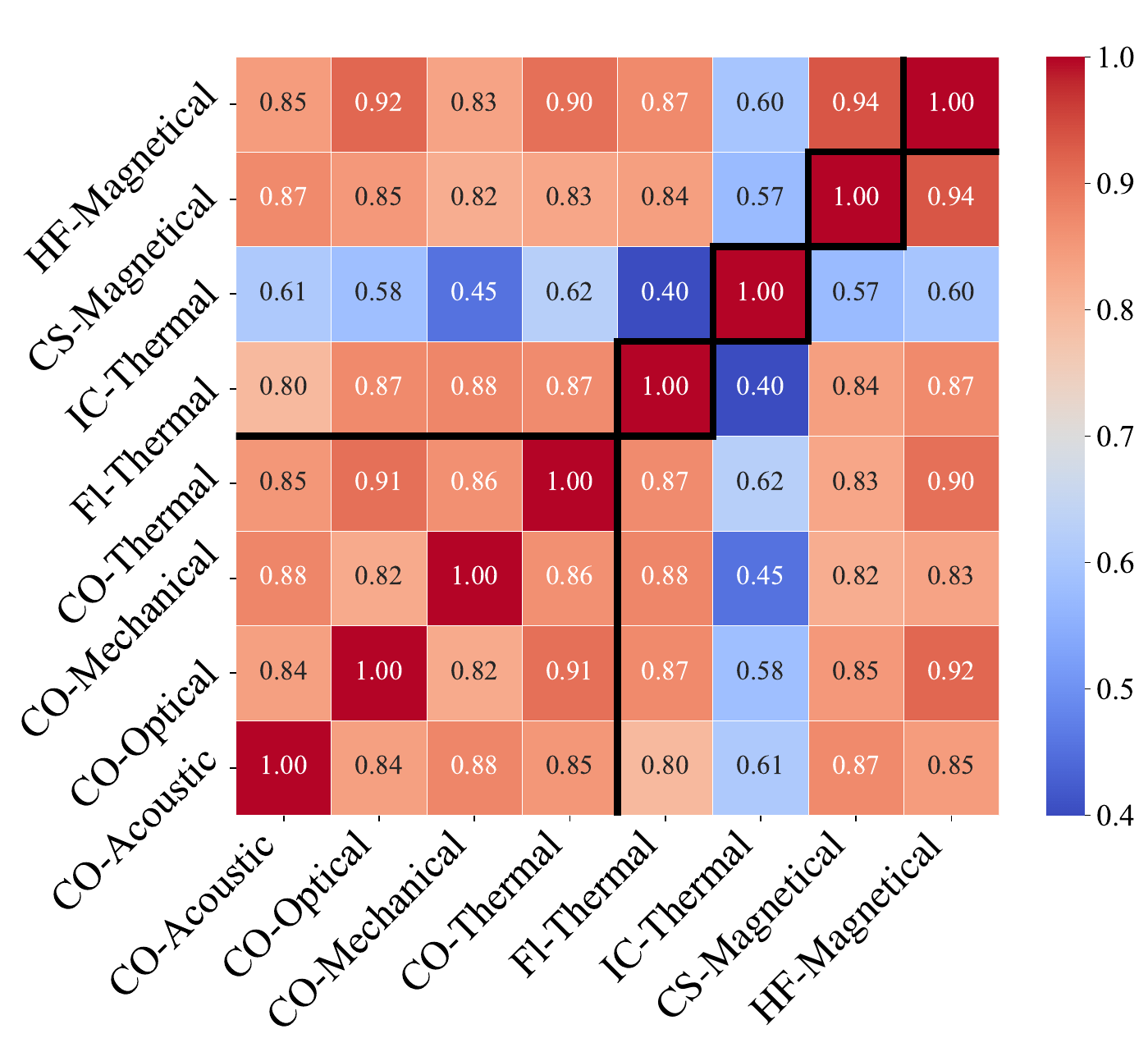}
    \centering
    \vspace{-1mm}
    \end{minipage}%
    }%
    \vspace{-2mm}
    \caption{Inner correlation matrix between 8 Software + Field combination, indicating both Answer and Action are indispensable, since they measure different ability dimensions.}
    \label{fig:corr}
    \vspace{-2mm}
\end{figure}

\subsection{GUI Agent Execution}
\label{sec:dataset-c}


After completing GUI-EDA, we benchmark existing agentic models—both general MLLMs and specialized GUI agents, the specific models used are listed in Section 6.2. For Answer score, each model is asked to produce a concise 5–10 word operational description. Considering traditional metrics like BLUE/CIDEr are limited to the word level and cannot represent complex semantic information in EDA scenarios (e.g. the ambiguous meaning of the word `chip'), we use the LLM-as-a-Judge paradigm to provide three categories: full compliance (1), partial compliance (0.5), and non-compliance (0).
Precision-oriented and recall-oriented prompts are run separately; the average of the two scores is reported as the Answer score.
For Action score, the model must output absolute (x, y) coordinates that receive (1) if the point falls inside the GT bounding box and (0) otherwise.

Unlike conventional GUI-agent benchmarks that split data only by software and evaluate  only Action score, the field+software taxonomy of GUI-EDA obliges us to score both Action and Answer. We run every agentic model on the samples of all 8 field+software subsets, average the scores, and compute across subsets; through the average of Spearman's rand-order correlation coefficient (SRCC) and Pearson linear correlation coefficient (PLCC), the resulting correlation matrices are shown in Figure \ref{fig:corr}. For Answer, subsets belonging to the same physics domain are strongly correlated (CS-Magnetical vs. HF-Magnetical yields SRCC = 0.89), indicating models that lead on one subset will also perform satisfactorily on another; whereas same-software pairs are not (CO-Optical vs. CO-Mechanical, SRCC = 0.38). For Action the pattern reverses: same-software subsets correlate highly while same-field pairs do not. Because Answer tests macroscopic semantic comprehension of the underlying physics, hence depends on field, whereas Action tests microscopic pixel-level manipulation of interface layout, hence depends on software. In conclusion, for EDA tasks, both scores are complementary since they measure distinct dimensions of capability.

\begin{figure*}[t]
\centering
\subfigure[COMSOL Acoustic]{
\begin{minipage}[t]{0.23\linewidth}
\includegraphics[width=\linewidth]{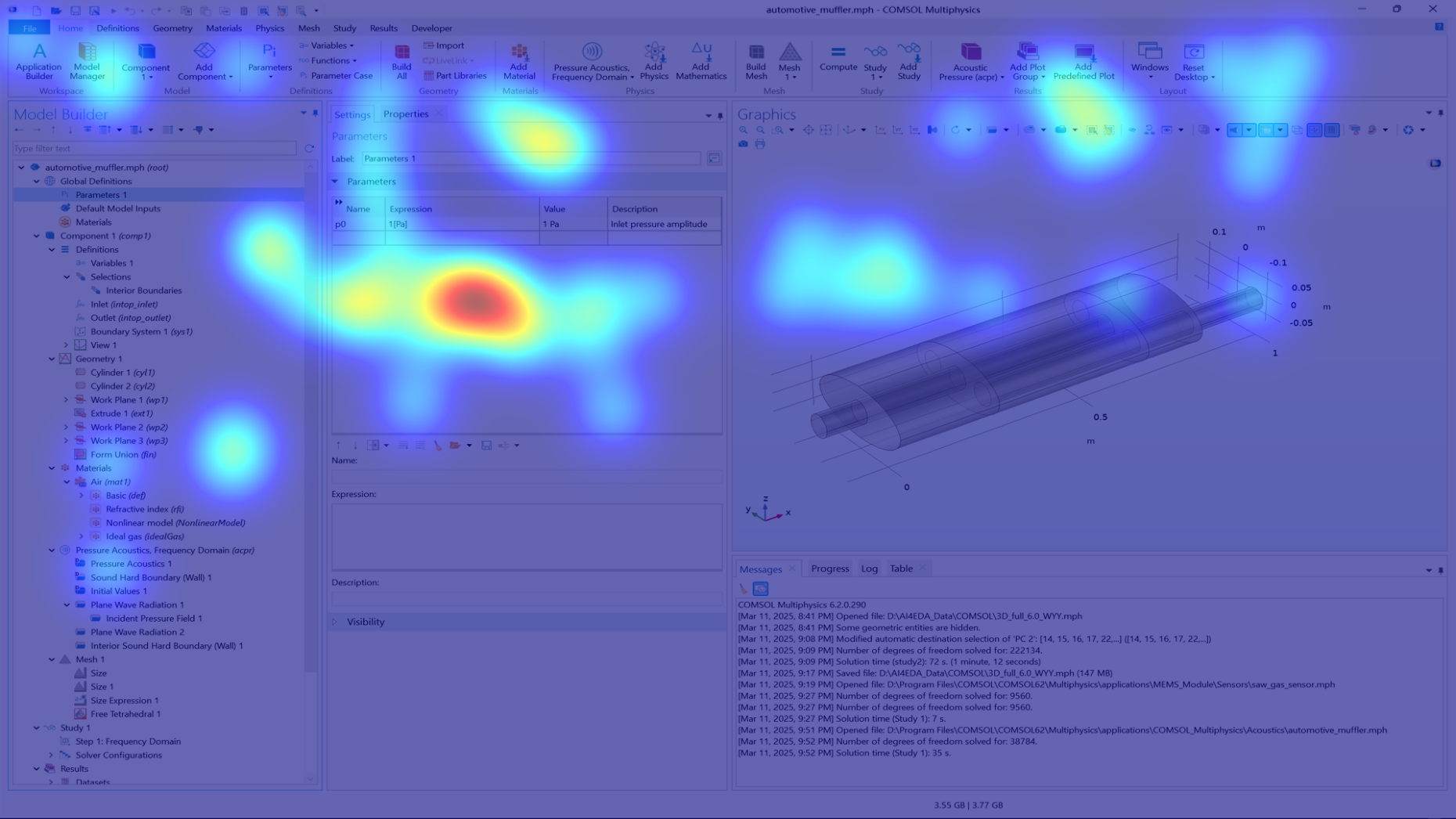}
\centering
\end{minipage}%
}%
\subfigure[COMSOL Optical]{
\begin{minipage}[t]{0.23\linewidth}
\includegraphics[width=\linewidth]{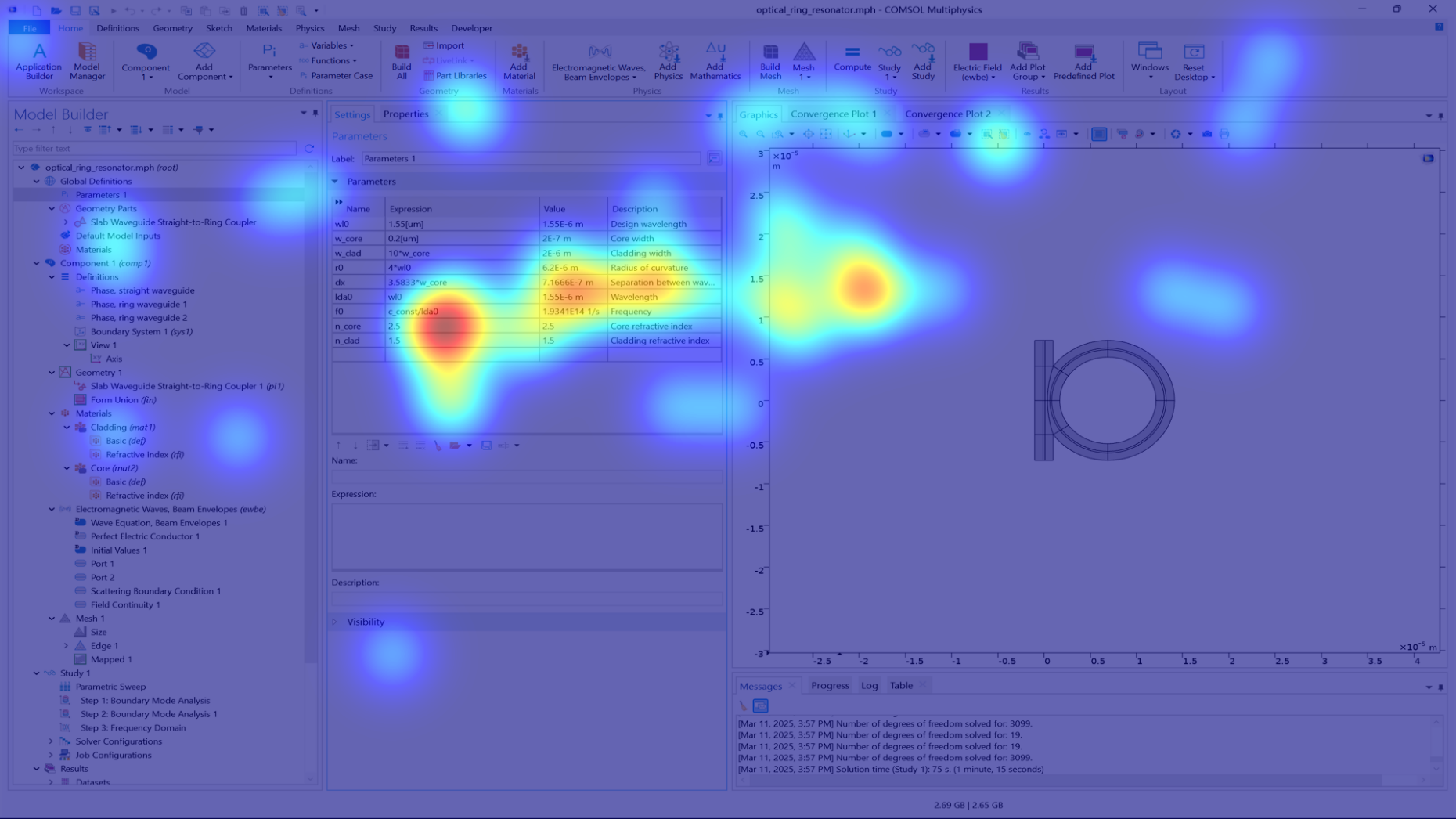}
\centering
\end{minipage}%
}%
\subfigure[COMSOL Mechanical]{
\begin{minipage}[t]{0.23\linewidth}
\includegraphics[width=\linewidth]{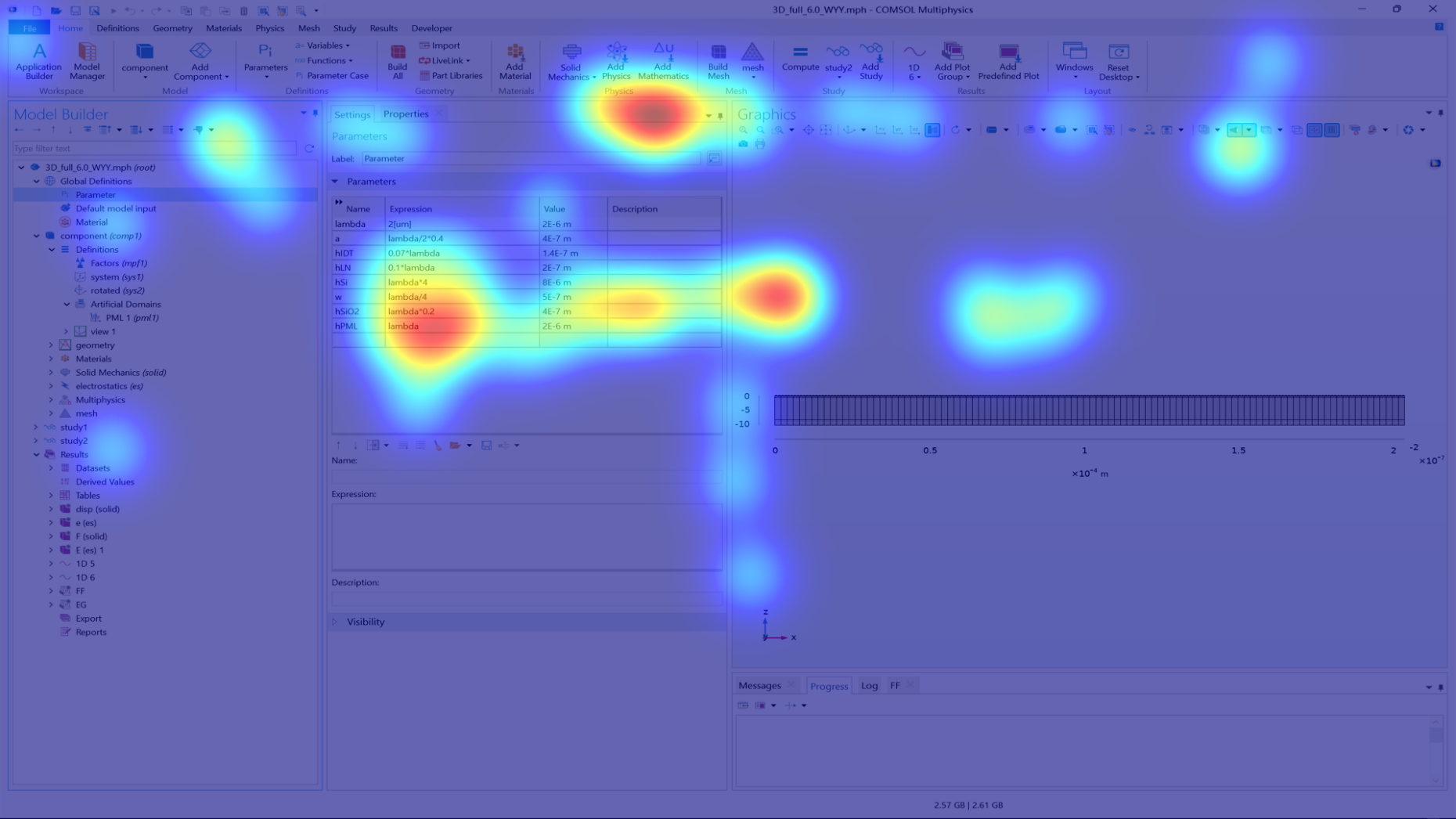}
\centering
\end{minipage}%
}%
\subfigure[COMSOL Thermal]{
\begin{minipage}[t]{0.23\linewidth}
\includegraphics[width=\linewidth]{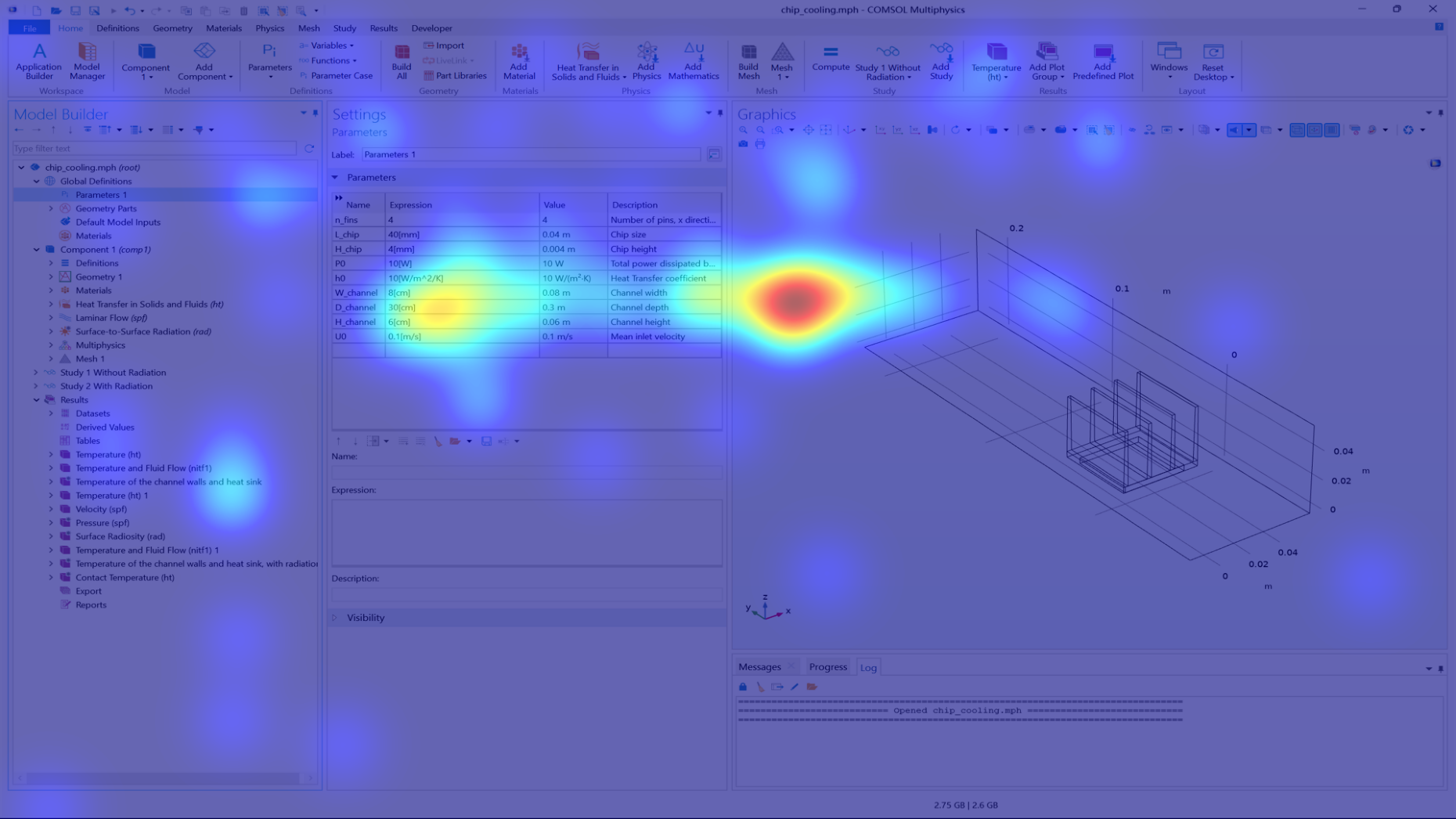}
\centering
\end{minipage}%
}%
\vspace{-2mm}

\subfigure[Flotherm Thermal]{
\begin{minipage}[t]{0.23\linewidth}
\includegraphics[width=\linewidth]{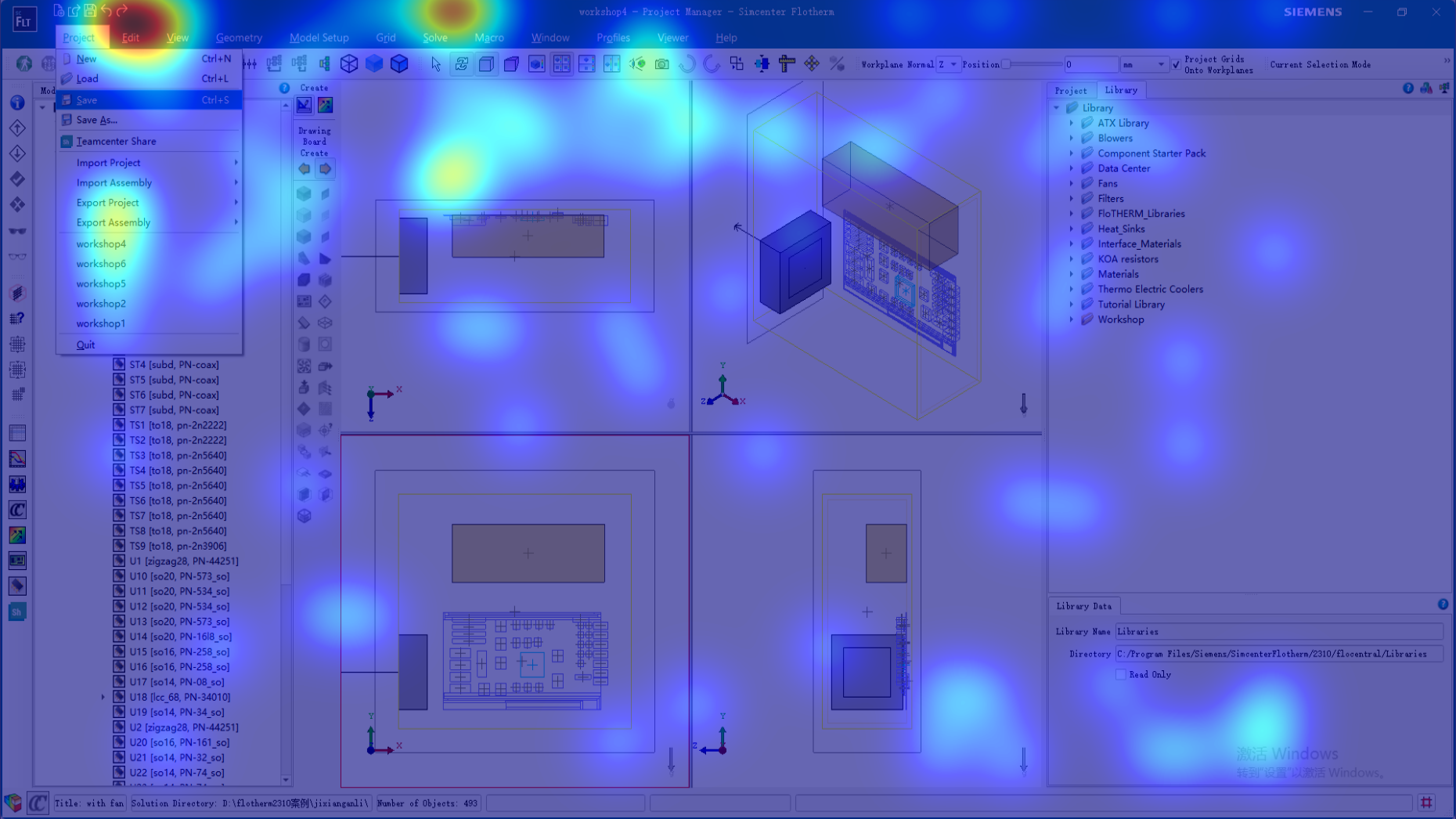}
\centering
\end{minipage}%
}%
\subfigure[ICEPAK Thermal]{
\begin{minipage}[t]{0.23\linewidth}
\includegraphics[width=\linewidth]{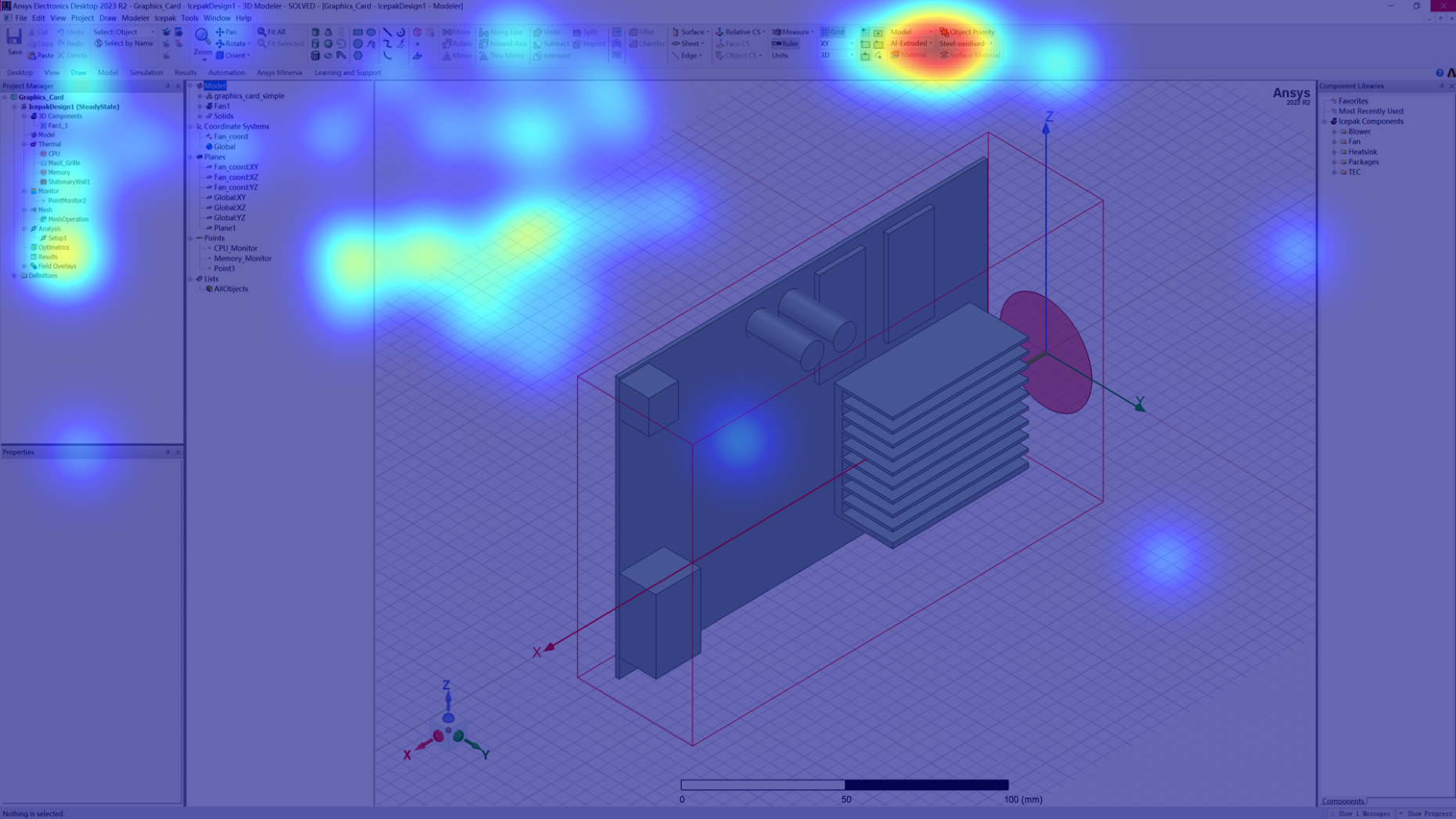}
\centering
\end{minipage}%
}%
\subfigure[CST Magnetical]{
\begin{minipage}[t]{0.23\linewidth}
\includegraphics[width=\linewidth]{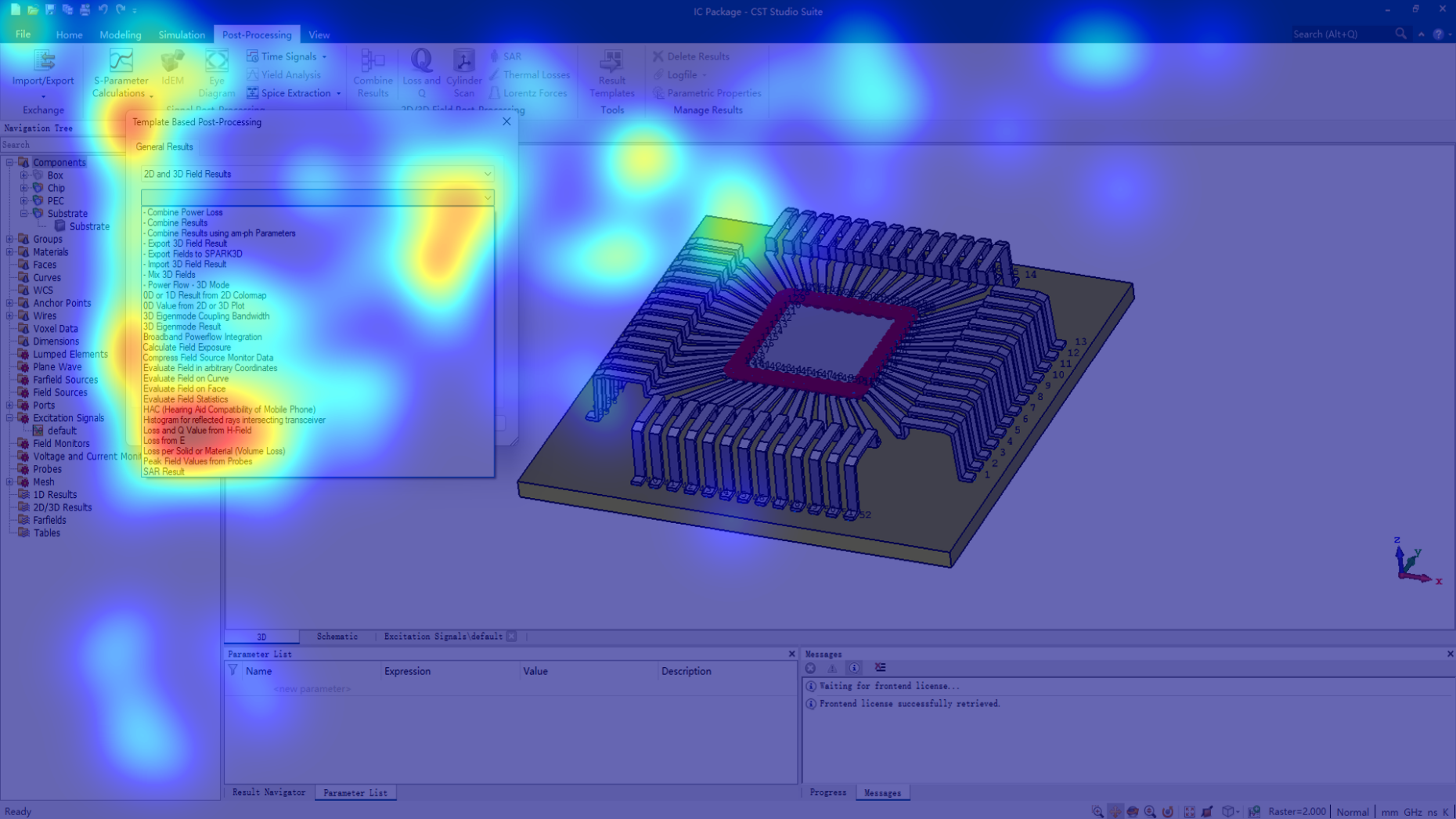}
\centering
\end{minipage}%
}%
\subfigure[HFSS Magnetical]{
\begin{minipage}[t]{0.23\linewidth}
\includegraphics[width=\linewidth]{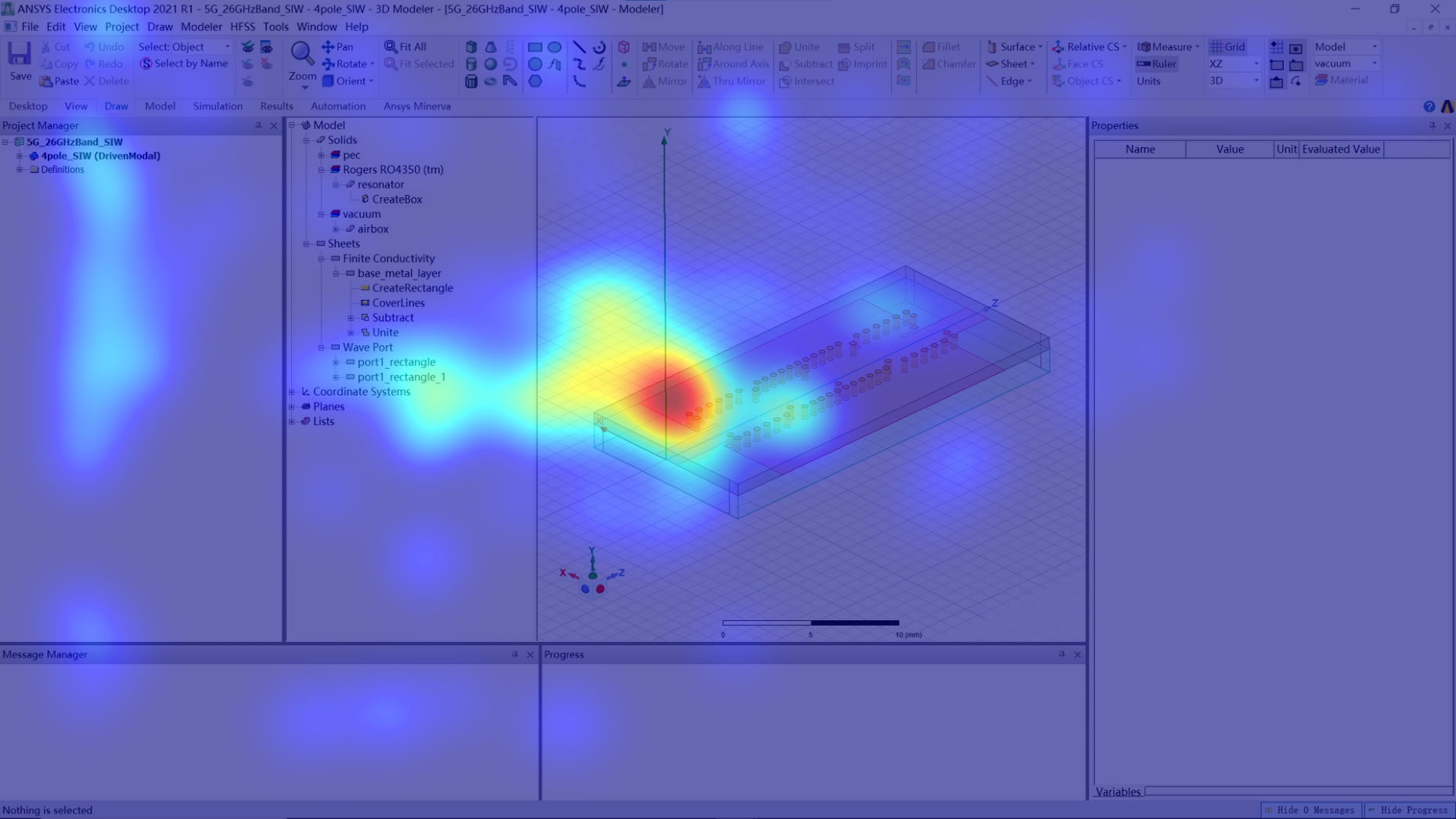}
\centering
\end{minipage}%
}%
\vspace{-2.5mm}

\caption{For each Field+Software combination, the Region of Interest (ROI) according to user clicks in the interface. ROIs for the same software show similar distribution, and are concentrated in the options bar.}
\vspace{-4mm}
\label{fig:click}
\end{figure*}

\begin{figure*}[t]
\subfigcapskip=-6pt 
\subfigbottomskip=2pt 
\centering
\subfigure[COMSOL Acoustic]{
\begin{minipage}[t]{0.23\linewidth}
\includegraphics[width=\linewidth]{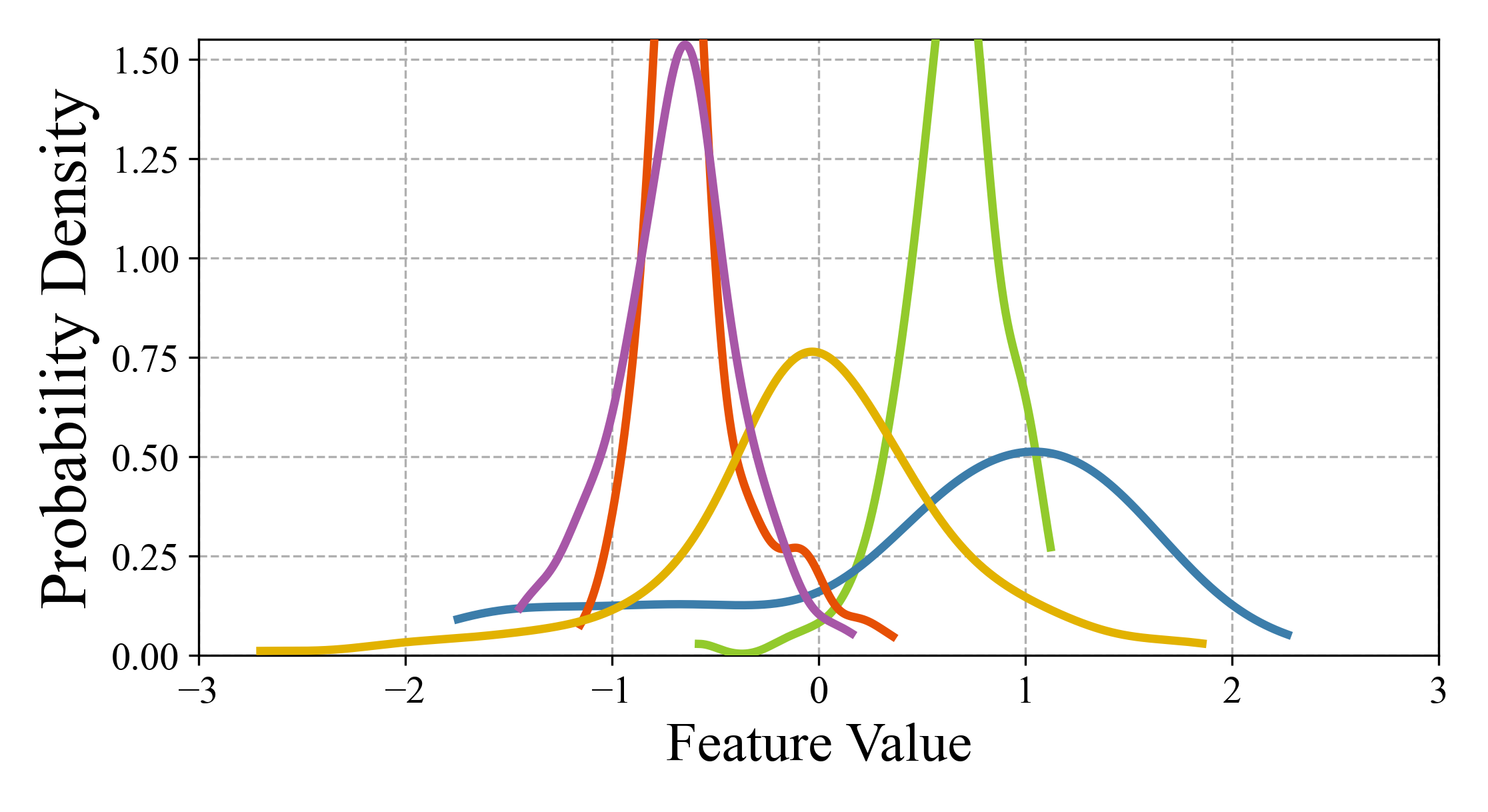}
\centering
\end{minipage}%
}%
\subfigure[COMSOL Optical]{
\begin{minipage}[t]{0.23\linewidth}
\includegraphics[width=\linewidth]{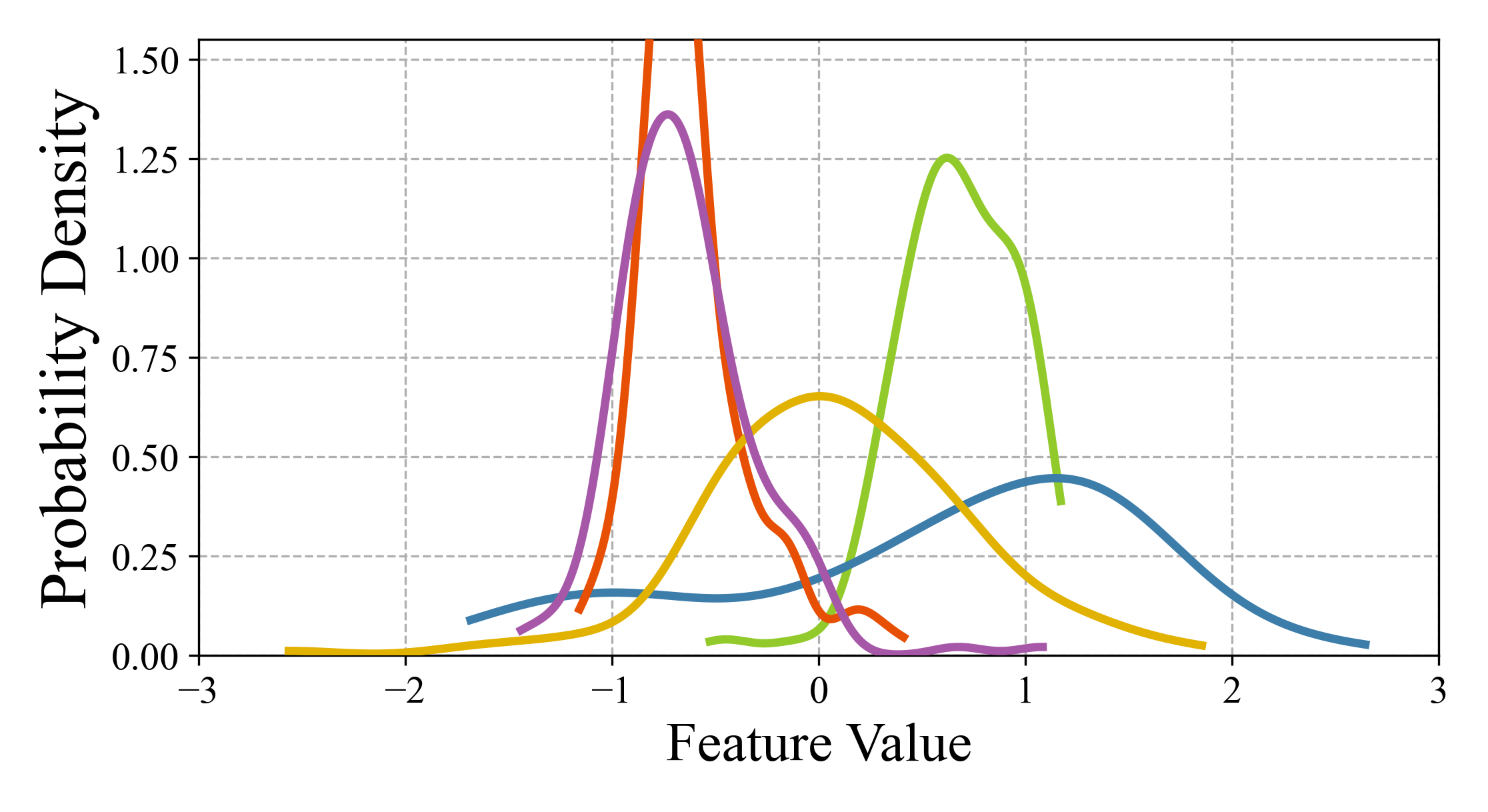}
\centering
\end{minipage}%
}%
\subfigure[COMSOL Mechanical]{
\begin{minipage}[t]{0.23\linewidth}
\includegraphics[width=\linewidth]{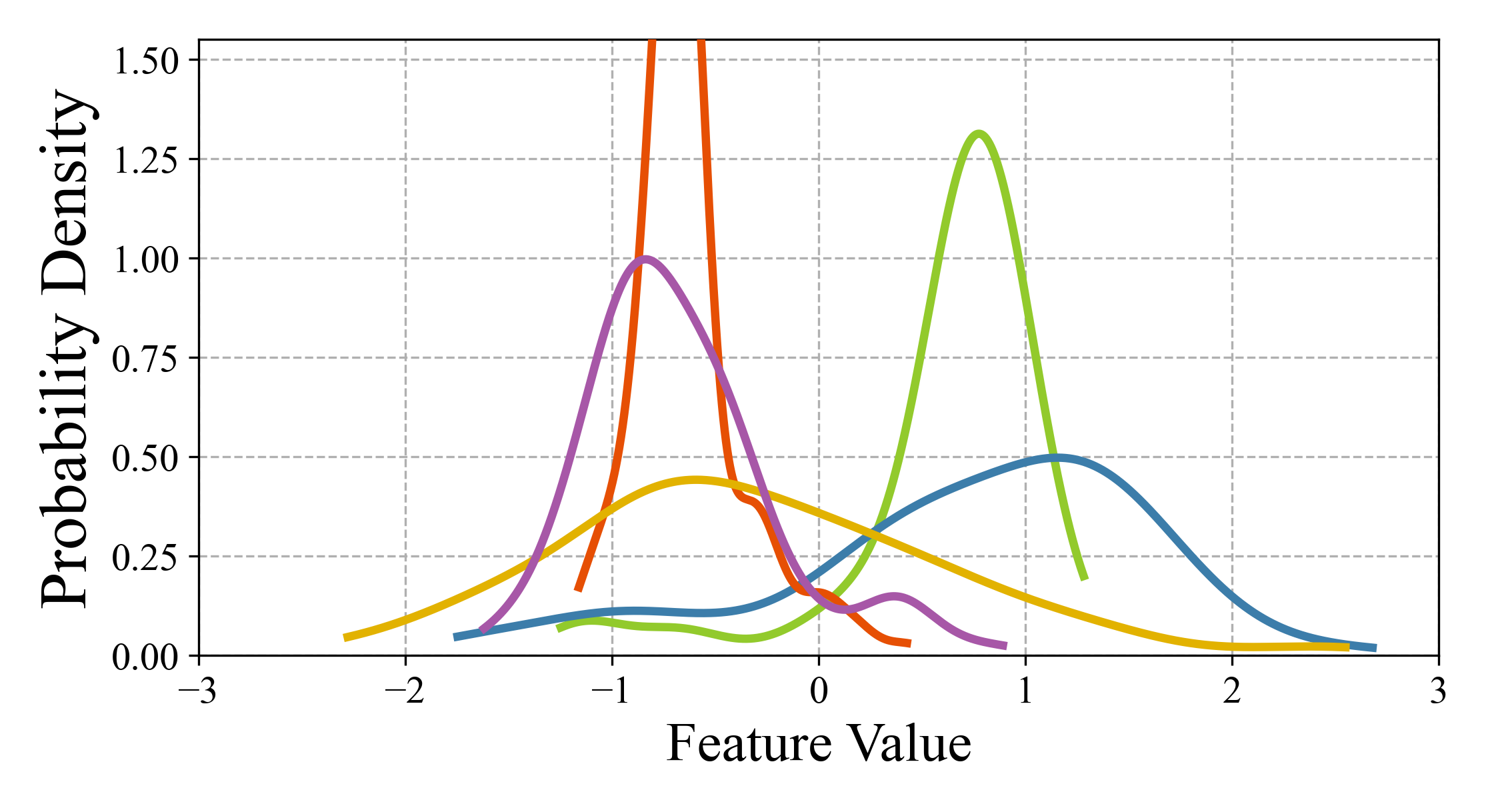}
\centering
\end{minipage}%
}%
\subfigure[COMSOL Thermal]{
\begin{minipage}[t]{0.23\linewidth}
\includegraphics[width=\linewidth]{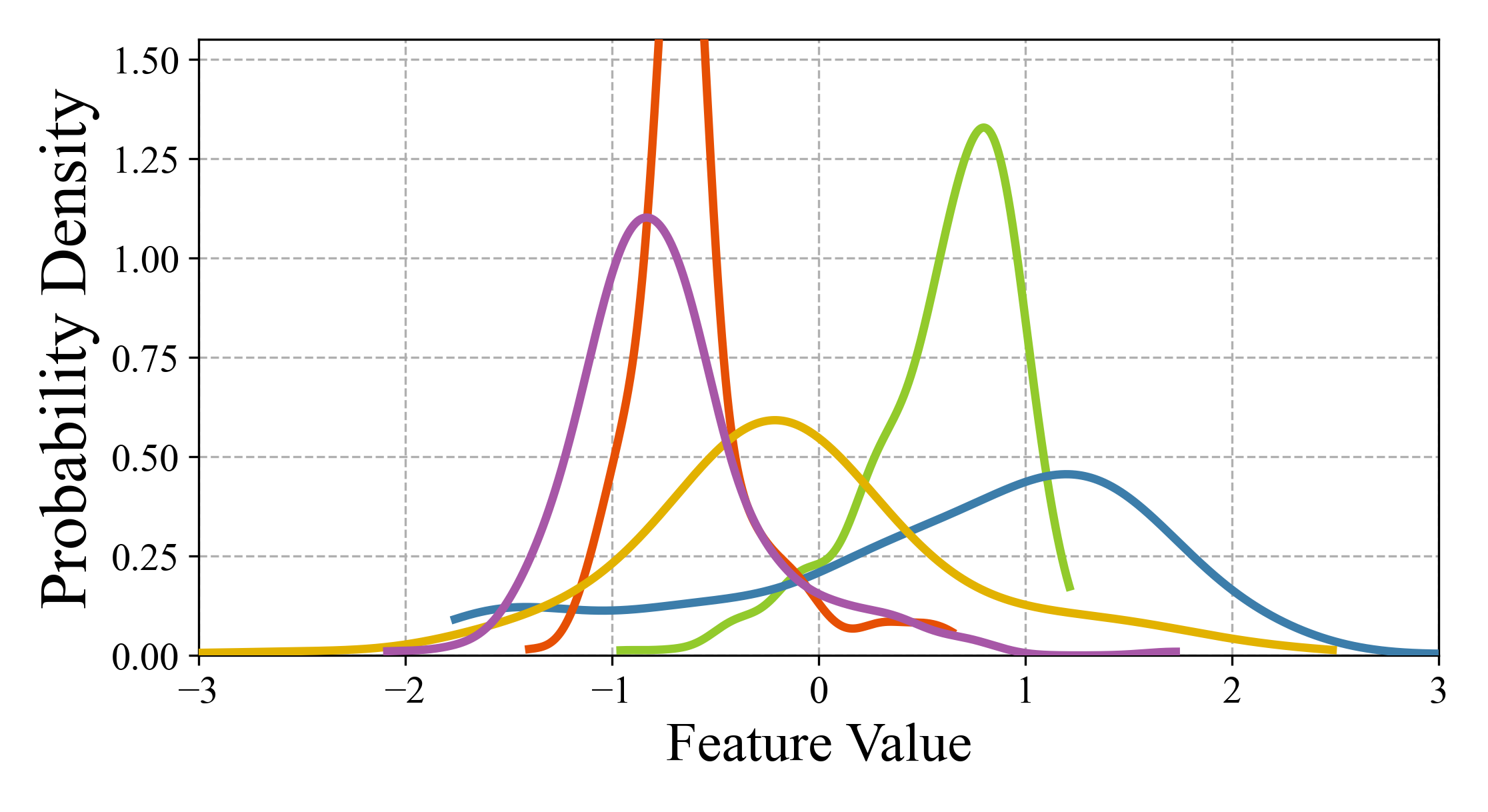}
\centering
\end{minipage}%
}%
\vspace{-2mm}

\subfigure[Flotherm Thermal]{
\begin{minipage}[t]{0.23\linewidth}
\includegraphics[width=\linewidth]{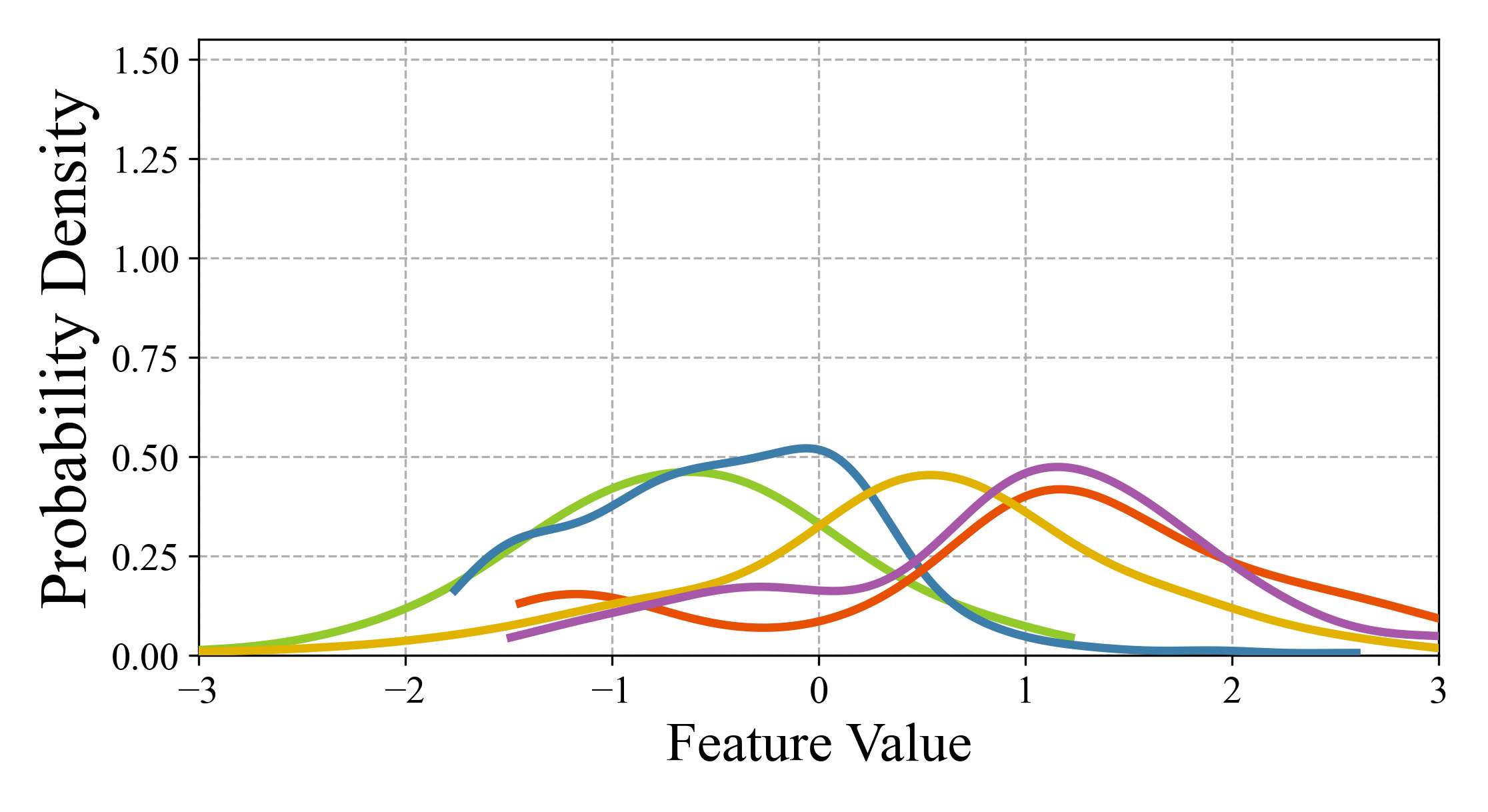}
\centering
\end{minipage}%
}%
\subfigure[ICEPAK Thermal]{
\begin{minipage}[t]{0.23\linewidth}
\includegraphics[width=\linewidth]{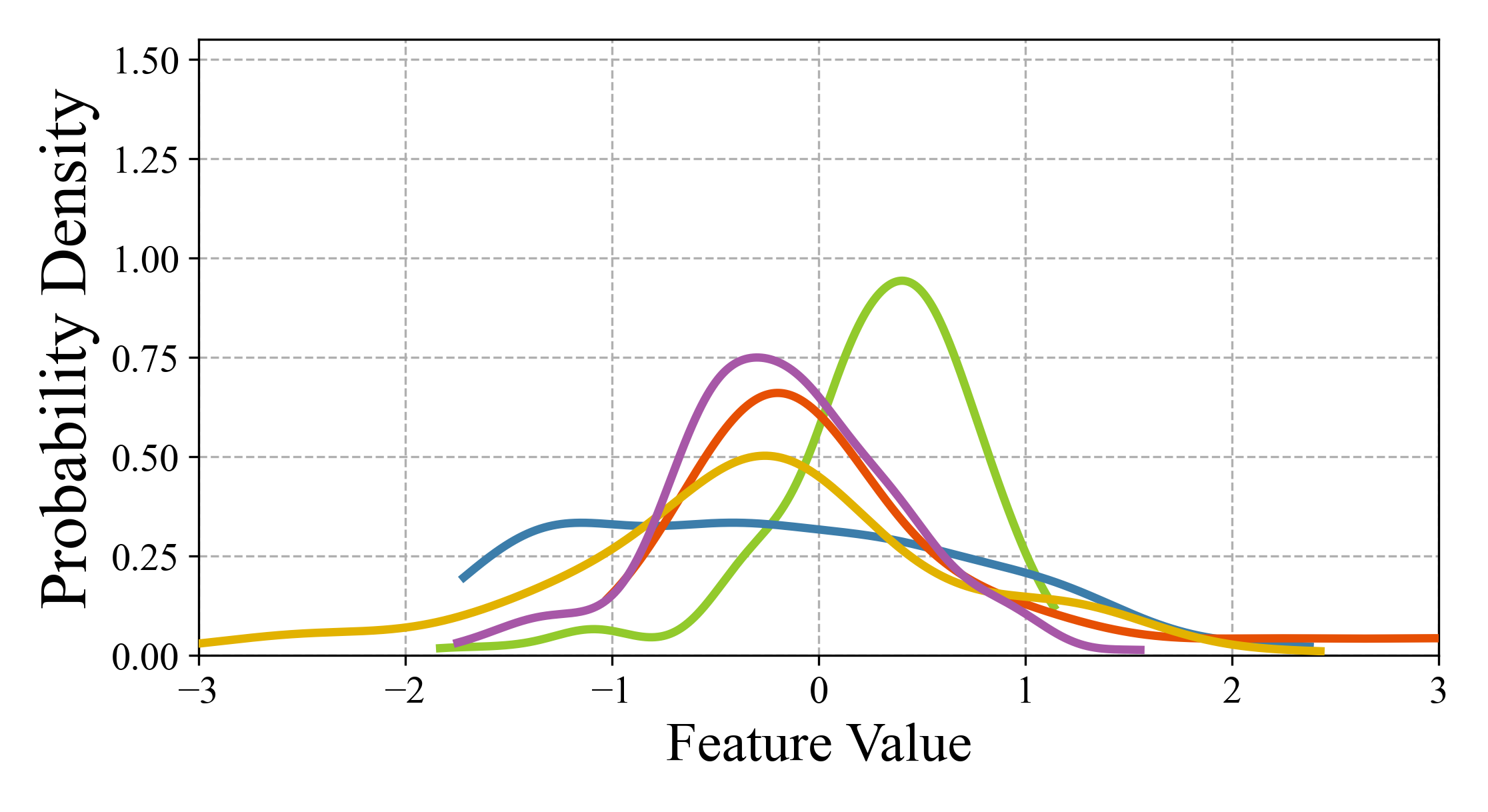}
\centering
\end{minipage}%
}%
\subfigure[CST Magnetical]{
\begin{minipage}[t]{0.23\linewidth}
\includegraphics[width=\linewidth]{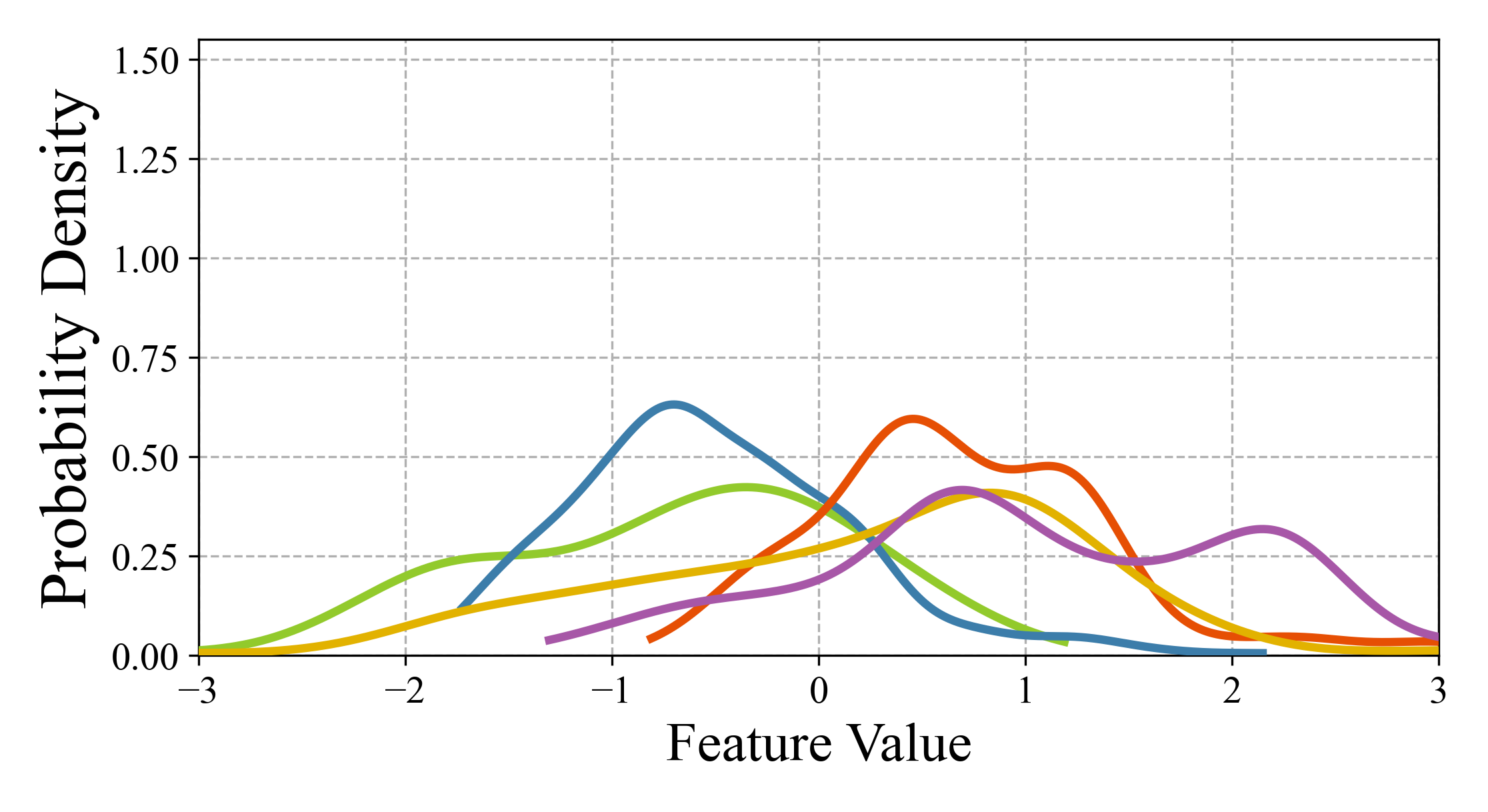}
\centering
\end{minipage}%
}%
\subfigure[HFSS Magnetical]{
\begin{minipage}[t]{0.23\linewidth}
\includegraphics[width=\linewidth]{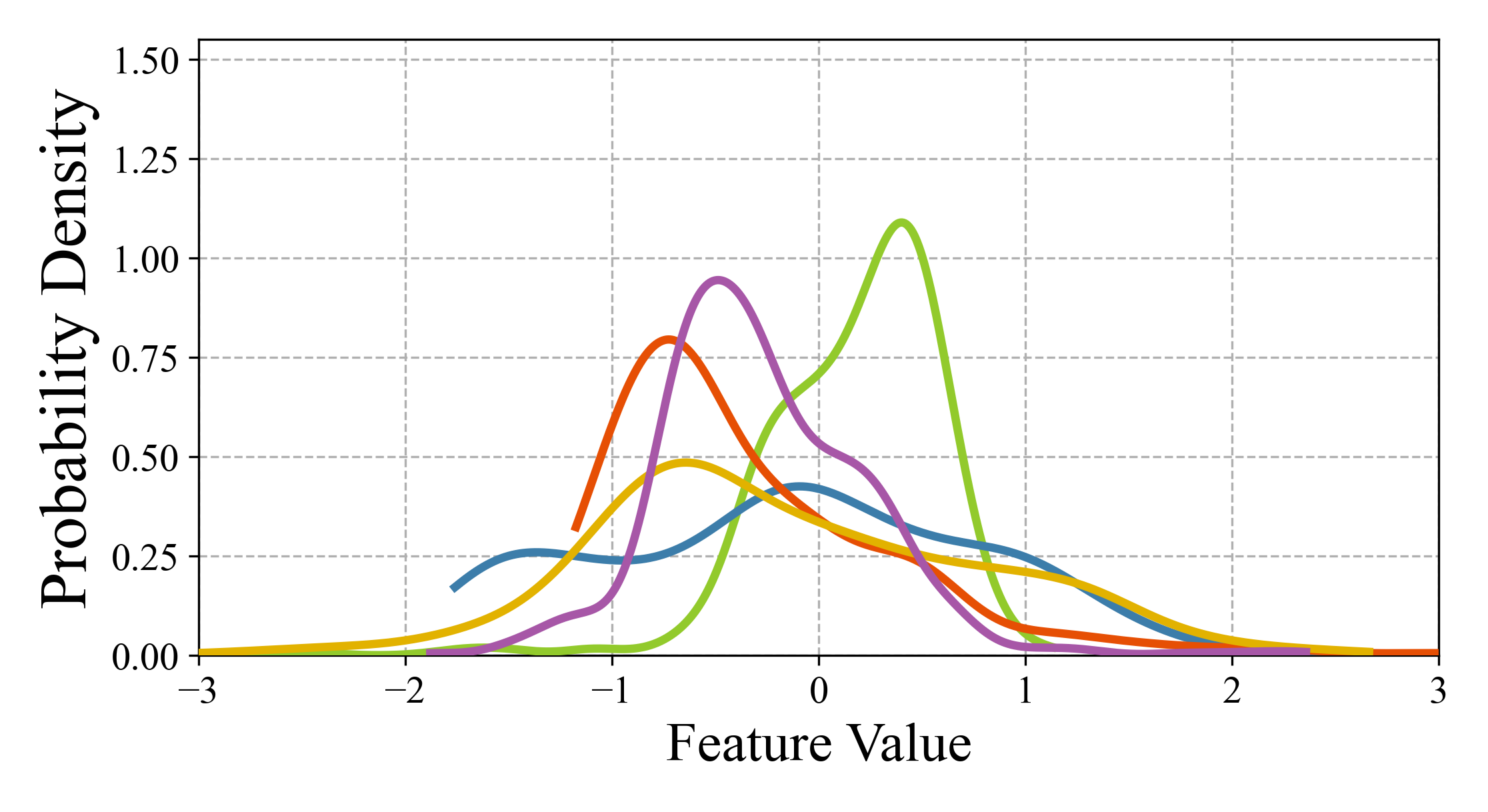}
\centering
\end{minipage}%
}%
\vspace{-2mm}

\caption{Low-level feature distribution of each Field+Software combination. Different colors denote \GroupA{Luminance}, \GroupB{Contrast}, \GroupC{Chrominance}, \GroupD{Blur}, and \GroupE{Spatial Information}, which are decisive factors for ROI.}
\label{fig:Feature}
\vspace{-3mm}

\end{figure*}

\subsection{Real-world Validation}
\label{sec:dataset-d}
Owing to the nature of EDA tasks, the ultimate verdict on a GUI agent is delivered not by the OS simulation, but by the fabricated component. We therefore selected 10 representative samples, including 5 multiphysics fields under each of Action scores (1 or 0), and ran silicon tape-outs to test whether the agent execution mark in Section \ref{sec:dataset-c} predicts Real-world success. Masks were prepared strictly according to the operation sequences generated by the agentic AI. Wafer-level measurements (e.g., wavelength, insertion loss, extinction ratio) show that every design with Action score 1 in CAD, satisfies the scientist Demand formulated in Section \ref{sec:dataset-a}, whereas all samples scored 0 deviate from the specified threshold by $>$ 10 \%, directly confirming that the virtual Action score is a reliable leading indicator of physical viability and closing the loop from on-screen decision to silicon validation.

\section{Data Analysis}

\subsection{Difficulty Annotation}
\label{sec:human}

Beyond the original 5 senior EDA engineers, to better analyze human performance in EDA tasks, we expand our annotation team to (i) assign a difficulty label to every sample and (ii) establish human baselines against which agentic models can be compared. 5 Electrical-Related Ph.D. students and 5 computer-literate undergraduates are involved to perform the Action task using the Stage-2 interface in Figure \ref{fig:interface}; a majority of 3/5 correct clicks inside the Ground-Truth bounding box defines the difficulty level. If the correct position is clicked by the majority of undergraduates, the sample is labeled as Easy level and their aggregated inference is archived as Human (average level); if only the Ph.D. majority succeeds, the sample is labeled as Normal level and recorded as Human (expert level); otherwise the sample is marked as Hard level. \footnote{Ph.D. majors: 2$\times$electronic engineering, 2$\times$computer science, and 1$\times$communication; None samples in GUI-EDA is solved by undergraduates but missed by Ph.D. students, ruling out Easy/Normal inversion.} Consequently, every GUI-EDA sample now carries 4 orthogonal tags: Field, Software, Resolution and Difficulty — whose interrelations are dissected in this Section to pinpoint the principal challenges and future directions for using GUI agents in EDA tasks.

\begin{figure*}[t]
\centering

\begin{minipage}[t]{0.95\linewidth}
\includegraphics[width=\linewidth]{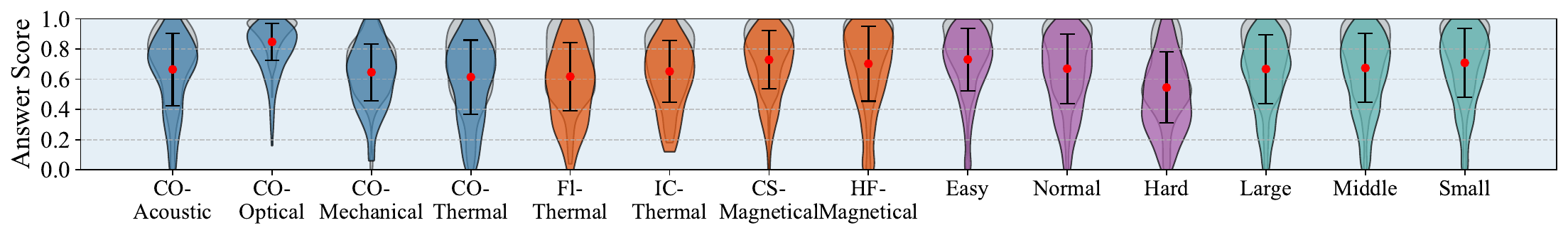}
\centering
\end{minipage}%
\vspace{-1.5mm}

\begin{minipage}[t]{0.95\linewidth}
\includegraphics[width=\linewidth]{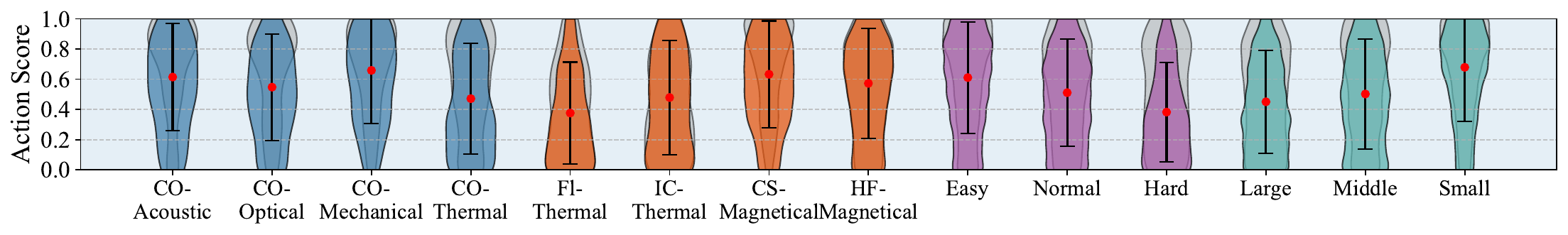}
\centering
\end{minipage}%

\vspace{-2mm}

\caption{Answer/Action Score distribution under different Field, Software, Resolution, and Difficulty, based on the performance of six advanced GUI Agents. Colored/Gray denote Precise/Recall for Answer (Top), and Horizontal/Vertical for Action (Below).}
\label{fig:Violin}
\vspace{-4mm}
\end{figure*}

\begin{figure*}[t]
    \centering
    
    \begin{minipage}[t]{0.33\linewidth}
    \includegraphics[width=\linewidth]{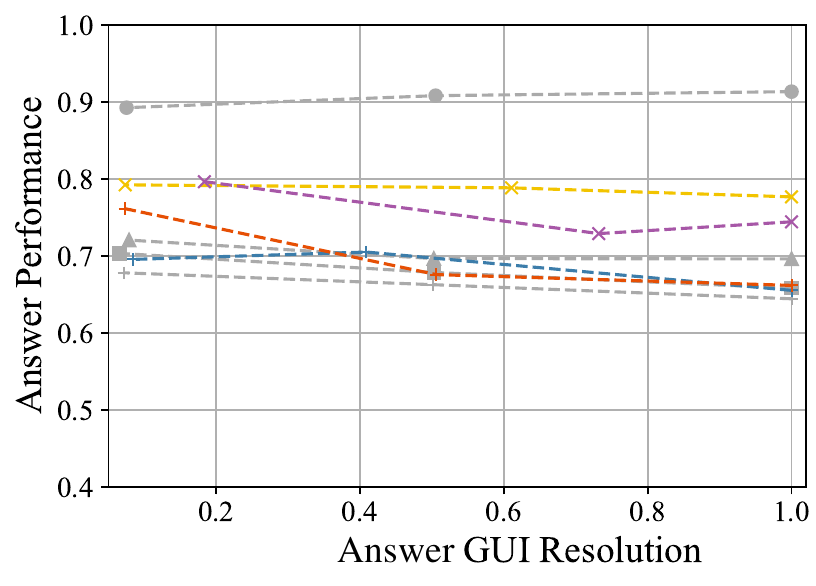}
    \centering
    \end{minipage}%
    \hspace{3mm}
    \begin{minipage}[t]{0.508\linewidth}
    \includegraphics[width=\linewidth]{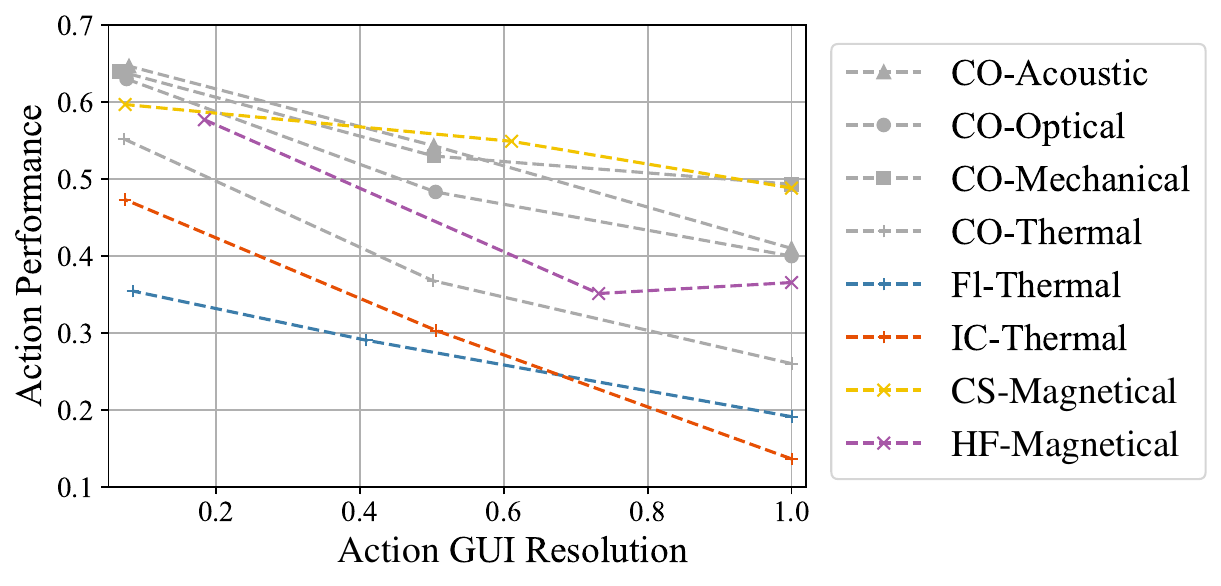}
    \centering
    \end{minipage}%
    \vspace{-2mm}
    \caption{The impact of different Resolution on Answer Score (Left) and Action Score (Right), based on the average scores of six advanced GUI Agents. The horizontal axis represents the ratio of pixels to the original $3840\times2160$ size. Resolution has no effect on Answer, smaller Resolution has a certain impact on better Action.}
    \label{fig:curve}
    \vspace{-3mm}
\end{figure*}


\subsection{Field \& Software Analysis}

Across the 8 Field-Software combination in GUI-EDA, we analyze both the human click distribution—i.e. the Region-of-interest (ROI), and the low-level image statistics of the interface, to discuss the correlation between Field and Software.

Heat-maps in Figure \ref{fig:click} reveal a universal toolbar bias: for every combination, most clicks fall within the top or left ribbon. Jensen–Shannon divergence among the spatial distributions ranges from 0.02 to 0.05, far below the random-expectation null ($p < 0.001$). Contours for the same Software (e.g. COMSOL) are almost identical across its 4 physics Fields, whereas different tools assigned to the same Field (e.g. CO/Fl/IC-Thermal) diverge markedly. Thus button location in the Software, not Fields content, chiefly governs the ROI.

Figure \ref{fig:Feature} shows that this preference is driven by low-level salience: Luminance, Chrominance, and Spatial Information detail all peak in the toolbar, while Contrast and Blur vary little for each combination. 
For software interfaces with brighter, sharper, and more structured contrast, the distribution of the three low-level attributes is more extreme, as shown in Figures \ref{fig:Feature} (a-d), where the Action pixel is more likely to be contained in the toolbar than the flat region, as shown in Figures \ref{fig:click} (a-d).

In conclusion, the GUI Action in EDA tasks exhibits a toolbar congregation that depend on CAD Software type more than multiphysics Field, corroborating the Action–Software correlation posited in Figure \ref{fig:corr}. Moreover, the more extreme low-level attributes are, the denser the solution space is compressed into the toolbar, furnishing a usable prior for future CAD-targeted GUI Agents.





\subsection{Resolution \& Difficulty Analysis}

After obtaining the GT for each sample in GUI-EDA, we perform inference using MLLM and GUI Agent according to the steps in Section \ref{sec:dataset-c}. The model pool is listed in Section \ref{sec:exp-settings}. For Answer and Action, we select the six best-performing models, taking their average performance to characterize the current Agents ability on the task. Following the description in Section \ref{sec:dataset-c}, the quantitative indicators of this `performance' are ($\romannumeral1$) Answer Score: the average of Precision and Recall; ($\romannumeral2$) Action Score: both Horizontal and Vertical coordinates are within the GT range. the Colored/Gray areas represent the Precision/Recall in Answer and the Accuracy of the Vertical/Horizontal coordinates in Answer respectively.

Figure \ref{fig:Violin} shows although the score histograms of four Fields in COMSOL and the other four Software differ in shape, their average are almost identical, and no agent pushes the Action score beyond 0.6, confirming that the bottleneck is a generic handicap rather than a single-Software flaw. For Answer, Precision/Recall-oriented distributions are overlapped, whereas for Action the Vertical (Color) region sit consistently below the Horizontal (Grey) ones, indicating left-right is easier than up-down search. Since buttons are packed more densely along the vertical axis and are themselves horizontally elongated rectangles, so a vertical miss is more probable, revealing `precise vertical pointing' as a cross-platform interaction bottleneck.
Difficulty and Resolution exert qualitatively different pressures. Answer scores decay monotonically from Easy to Hard, whereas Action scores collapse as the semantic-level burden turns into a pixel-level error. Shrinking GUI size enlarges the button-pixel ratio, leaving Answer unchanged but markedly improving Action, evidencing `larger screen → lower click tolerance' as another key constraint.


\begin{table}[t]
\centering
    \renewcommand\arraystretch{1.25}
    \renewcommand\tabcolsep{6pt}
    \belowrulesep=0pt\aboverulesep=0pt
    \caption{Answer\&Action correlation in 3 resolution size.}
    \label{tab:resolution-corr}
    \vspace{-5pt}
    \resizebox{\linewidth}{!}{
    \begin{tabular}{l|llllll}
    \toprule
Resolution    & \multicolumn{2}{c}{Large}       & \multicolumn{2}{c}{Middle}      & \multicolumn{2}{c}{Small}       \\ \cdashline{1-7}
Task          & \multicolumn{1}{c}{SRCC} & \multicolumn{1}{c}{PLCC} & \multicolumn{1}{c}{SRCC} & \multicolumn{1}{c}{PLCC} & \multicolumn{1}{c}{SRCC} & \multicolumn{1}{c}{PLCC} \\ \midrule
CO-Acoustic   & 0.2987   & 0.3041   & 0.3877   & 0.4371   & 0.3371   & 0.4791   \\
CO-Optical    & 0.3174   & 0.2403   & 0.3099   & 0.2435   & 0.1920   & 0.1483   \\
CO-Mechanical & 0.3411   & 0.3699   & 0.2786   & 0.3397   & 0.0275   & 0.3627   \\
CO-Thermal    & 0.4934   & 0.4757   & 0.3823   & 0.3483   & 0.3352   & 0.3414   \\
Fl-Thermal    & 0.1445   & 0.1018   & 0.2070   & 0.1801   & 0.2159   & 0.2068   \\
IC-Thermal    & 0.2764   & 0.2447   & 0.5220   & 0.5047   & 0.5387   & 0.5366   \\
CS-Magnetical & 0.2616   & 0.2212   & 0.2589   & 0.2468   & 0.2986   & 0.3169   \\
HF-Magnetical & 0.2325   & 0.2859   & 0.3747   & 0.3734   & 0.3766   & 0.4141   \\
All           & 0.3564   & 0.3383   & 0.3654   & 0.3354   & 0.3217   & 0.3118  \\ \bottomrule
\end{tabular}
    }
    \vspace{-2mm}
\end{table}

\begin{table}[t]
\centering
    \renewcommand\arraystretch{1.25}
    \renewcommand\tabcolsep{4.5pt}
    \belowrulesep=0pt\aboverulesep=0pt
    \caption{Action score increases as GUI become smaller.}
    \label{tab:resolution-grow}
    \vspace{-5pt}
    \resizebox{\linewidth}{!}{
    \begin{tabular}{l|lll}
    \toprule
Resolution    & \multicolumn{1}{c}{Large}  & \multicolumn{1}{c}{Middle}    & \multicolumn{1}{c}{Small}    \\ \midrule
CO-Acoustic   & 0.4100 & 0.5433$_{\rm(+0.13, 32.51\%)}$  & 0.6467$_{\rm(+0.10, 19.03\%)}$ \\
CO-Optical    & 0.4000 & 0.4833$_{\rm(+0.08, 20.83\%)}$  & 0.6300$_{\rm(+0.15, 30.35\%)}$ \\
CO-Mechanical & 0.4933 & 0.5300$_{\rm(+0.04, 7.44\%)}$   & 0.6400$_{\rm(+0.11, 20.75\%)}$ \\
CO-Thermal    & 0.2604 & 0.3681$_{\rm(+0.11, 41.36\%)}$  & 0.5521$_{\rm(+0.18, 49.99\%)}$ \\
Fl-Thermal    & 0.1915 & 0.2908$_{\rm(+0.10, 51.85\%)}$  & 0.3546$_{\rm(+0.06, 21.94\%)}$ \\
IC-Thermal    & 0.1367 & 0.3033$_{\rm(+0.17, 121.8\%)}$ & 0.4733$_{\rm(+0.17, 56.05\%)}$ \\
CS-Magnetical & 0.4884 & 0.5491$_{\rm(+0.06, 12.43\%)}$  & 0.5969$_{\rm(+0.05, 8.71\%)}$  \\
HF-Magnetical & 0.3514 & 0.3657$_{\rm(+0.01, 4.07\%)}$   & 0.5771$_{\rm(+0.21, 57.81\%)}$ \\
All           & 0.3432 & 0.4274$_{\rm(+0.08, 24.53\%)}$  & 0.5588$_{\rm(+0.13, 30.74\%)}$ \\ \bottomrule
\end{tabular}
    }
    \vspace{-3mm}
\end{table}

Figure \ref{fig:curve} further quantifies how Resolution influences the Answer and Action Score above: we plot the mean scores of the six best-performing models against the fraction of the native $3840\times2160$ canvas that each crop occupies. Answer remains flat across Small ($0.1\times$), Middle ($0.5\times$), and Large ($1\times$) scales, whereas Action declines monotonically in seven of the eight field-and-software pairs (all except HF-Magnetical). The uniform drop from Small-Middle-Large confirms that Action, rather than Answer, is the Resolution-sensitive term and therefore merits further analysis.

We therefore distill two observations: ($\romannumeral1$) Answer and Action pursue distinct objectives and exhibit markedly different score patterns; ($\romannumeral2$) Action is the more fragile task, simultaneously sensitive to both Difficulty and Resolution, while Answer responds only to Difficulty. Guided by these findings we next examine ($\romannumeral1$) the Answer–Action correlation computed by SRCC and PLCC) and ($\romannumeral2$) the Action Score gain across three Resolution size and three Difficulty levels.

Regarding Resolution, Table \ref{tab:resolution-corr} shows as resolution decreases from Large to Small, the Answer-Action correlation coefficients for the eight Field+Software combinations generally fluctuate about 0.35, without consistent upward or downward trend. CO-Thermal correlation drops from 0.49 to 0.34, while IC-Thermal correlation initially rises and then falls, showing fluctuations. Table \ref{tab:resolution-grow}, however, shows a monotonically scaling process increases the Action score by 20\%-50\%, with IC-Thermal increase the most reaching 120\%, and Fl-Thermal also experiencing a sustained benefit. The juxtaposition of these two tables demonstrates that spatial scale compression can improve the Action Score by optimizing the ROI, but the statistical correlation between Answer and Action remains loose, failing to simultaneously improve Answer semantic comprehension. Resolution optimization only has a unidirectional effect on the pixel-level.

\begin{table}[t]
\centering
    \renewcommand\arraystretch{1.25}
    \renewcommand\tabcolsep{6pt}
    \belowrulesep=0pt\aboverulesep=0pt
    \caption{Answer\&Action correlation in 3 difficulty level.}
    \label{tab:difficulty-corr}
    \vspace{-5pt}
    \resizebox{\linewidth}{!}{
    \begin{tabular}{l|llllll}
    \toprule
Resolution    & \multicolumn{2}{c}{Hard}       & \multicolumn{2}{c}{Normal}      & \multicolumn{2}{c}{Easy}       \\ \cdashline{1-7}
Task          & \multicolumn{1}{c}{SRCC} & \multicolumn{1}{c}{PLCC} & \multicolumn{1}{c}{SRCC} & \multicolumn{1}{c}{PLCC} & \multicolumn{1}{c}{SRCC} & \multicolumn{1}{c}{PLCC} \\ \midrule
CO-Acoustic & 0.4340  & 0.5092  & 0.0433  & 0.0544  & 0.1051  & 0.0994 \\
CO-Optical & 0.2757  & 0.3702  & 0.0384  & 0.1021  & 0.0994  & 0.0870 \\ 
CO-Mechanical & 0.2555  & 0.2451  & 0.4365  & 0.4575  & 0.2000  & 0.1652 \\ 
CO-Thermal & 0.2881  & 0.3109  & 0.6076  & 0.6103  & 0.2193  & 0.1830 \\ 
Fl-Thermal & 0.3341  & 0.3348  & 0.2626  & 0.2882  & 0.4009  & 0.3561 \\ 
IC-Thermal & 0.2411  & 0.2674  & 0.6183  & 0.6014  & 0.0386  & 0.0116  \\
CS-Magnetical & 0.0767  & 0.1515  & 0.2332  & 0.2625  & 0.2787  & 0.2179 \\ 
HF-Magnetical & 0.3663  & 0.3783  & 0.3370  & 0.3427  & 0.3279  & 0.3835  \\
All & 0.4021  & 0.4103  & 0.3220  & 0.3457  & 0.3526  & 0.3445  \\ \bottomrule
\end{tabular}
    }
    \vspace{-2mm}
\end{table}

\begin{table}[t]
\centering
    \renewcommand\arraystretch{1.25}
    \renewcommand\tabcolsep{4.5pt}
    \belowrulesep=0pt\aboverulesep=0pt
    \caption{Action score increases as question become easier.}
    \label{tab:difficulty-grow}
    \vspace{-5pt}
    \resizebox{\linewidth}{!}{
    \begin{tabular}{l|lll}
    \toprule
Resolution    & \multicolumn{1}{c}{Hard}  & \multicolumn{1}{c}{Normal}    & \multicolumn{1}{c}{Easy}    \\ \midrule
CO-Acoustic & 0.3048  & 0.3968$_{\rm(+0.09,30.21\%)}$ & 0.6148$_{\rm(+0.22,54.94\%)}$ \\
CO-Optical & 0.3810  & 0.4233$_{\rm(+0.04,11.11\%)}$ & 0.4934$_{\rm(+0.07,16.56\%)}$ \\
CO-Mechanical & 0.3333  & 0.4381$_{\rm(+0.11,31.43\%)}$ & 0.6241$_{\rm(+0.19,42.45\%)}$ \\
CO-Thermal & 0.2585  & 0.3768$_{\rm(+0.12,45.77\%)}$ & 0.4753$_{\rm(+0.10,26.13\%)}$ \\
Fl-Thermal & 0.3095  & 0.3166$_{\rm(+0.01,2.29\%)}$ & 0.3312$_{\rm(+0.02,4.62\%)}$ \\
IC-Thermal & 0.1143  & 0.3598$_{\rm(+0.25,214.8\%)}$ & 0.4991$_{\rm(+0.14,38.73\%)}$ \\
CS-Magnetical & 0.2619  & 0.5098$_{\rm(+0.25,94.65\%)}$ & 0.6421$_{\rm(+0.13,25.95\%)}$ \\
HF-Magnetical & 0.4776  & 0.4890$_{\rm(+0.01,2.39\%)}$ & 0.5143$_{\rm(+0.03,5.17\%)}$ \\
All & 0.3051  & 0.4138$_{\rm(+0.10,35.62\%)}$ & 0.5242$_{\rm(+0.11,26.71\%)}$ \\ \bottomrule
\end{tabular}
    }
    \vspace{-3mm}
\end{table}

Regarding Difficulty, Table \ref{tab:difficulty-corr} shows as the difficulty of questions decreases from Hard to Easy, the Answer-Action correlation coefficient increases from 0.20 to 0.50. IC-Thermal is only 0.04 in the Hard setting, but rebounds to 0.24 in the Easy setting, indicating that reduced cognitive load can significantly reduce semantic ambiguity and thus enhance Answer and Action together. Table \ref{tab:difficulty-grow} also shows that Action scores in the Normal setting increase by an average of 35\% compared to the Hard setting, with IC-Thermal scores increasing by over 200\%. These two tables demonstrate that reducing difficulty not only improves Action Scores but also, by clarifying the target semantics, leads to a simultaneous increase in Answer Scores, truly achieving correct GUI Agent operation.

Based on the analysis above, changes in Resolution primarily affect pixel-level accuracy. Their effect is limited to pixel-level error correction, and their impact on action optimization is limited. Changes in Difficulty, on the other hand, determine semantic-level comprehension. By adjusting the degree of semantic ambiguity, they optimize semantic-level answers, then ultimately optimizing pixel-level actions. The current bottleneck of GUI Agents has shifted from spatial localization to semantic comprehension. Future performance improvements should prioritize reducing cognitive complexity rather than further compressing the interface scale.

\begin{figure*}[t]
\subfigcapskip=-4pt 
\subfigbottomskip=2pt 
\centering
\subfigure[COMSOL Acoustic]{
\begin{minipage}[t]{0.23\linewidth}
\includegraphics[width=\linewidth]{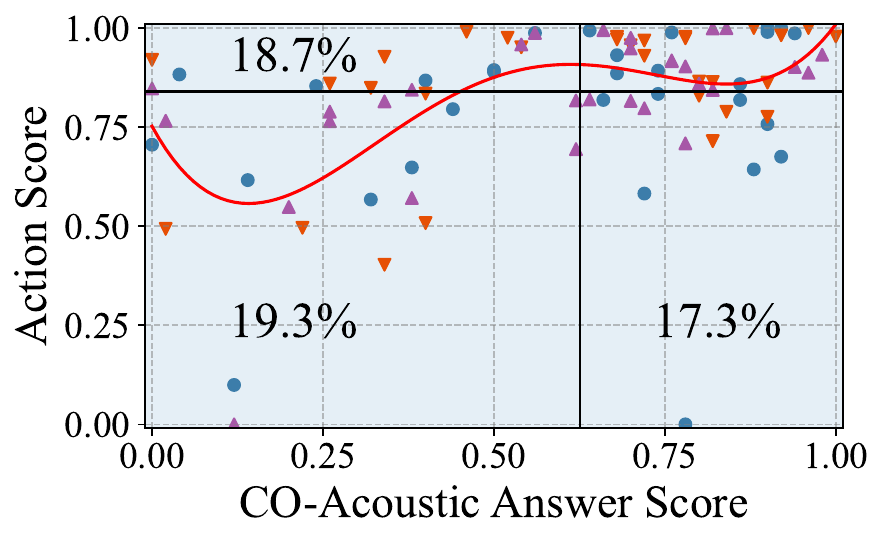}
\centering
\end{minipage}%
}%
\subfigure[COMSOL Thermal]{
\begin{minipage}[t]{0.23\linewidth}
\includegraphics[width=\linewidth]{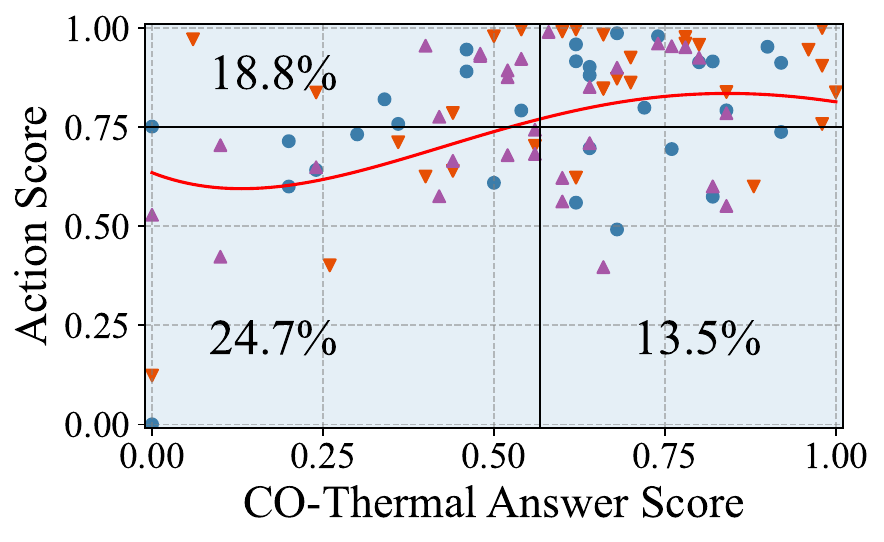}
\centering
\end{minipage}%
}%
\subfigure[COMSOL Mechanical]{
\begin{minipage}[t]{0.23\linewidth}
\includegraphics[width=\linewidth]{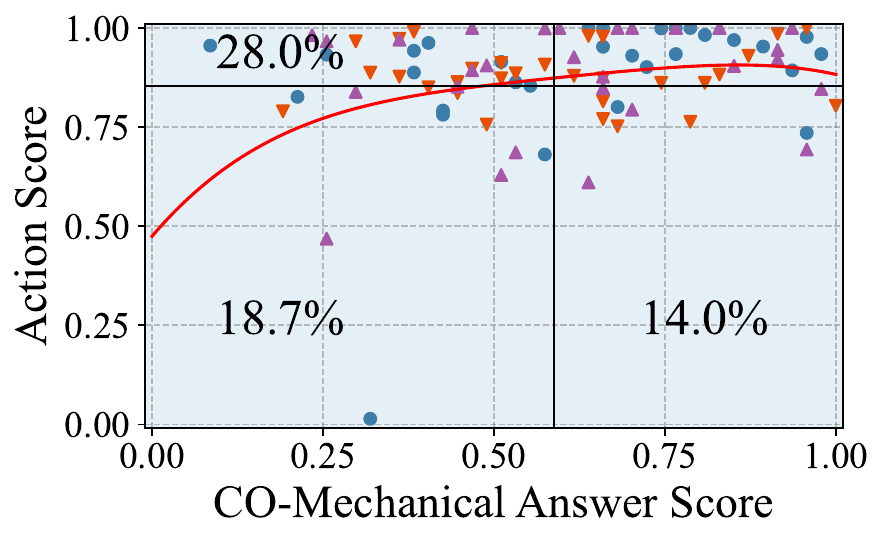}
\centering
\end{minipage}%
}%
\subfigure[COMSOL Optical]{
\begin{minipage}[t]{0.23\linewidth}
\includegraphics[width=\linewidth]{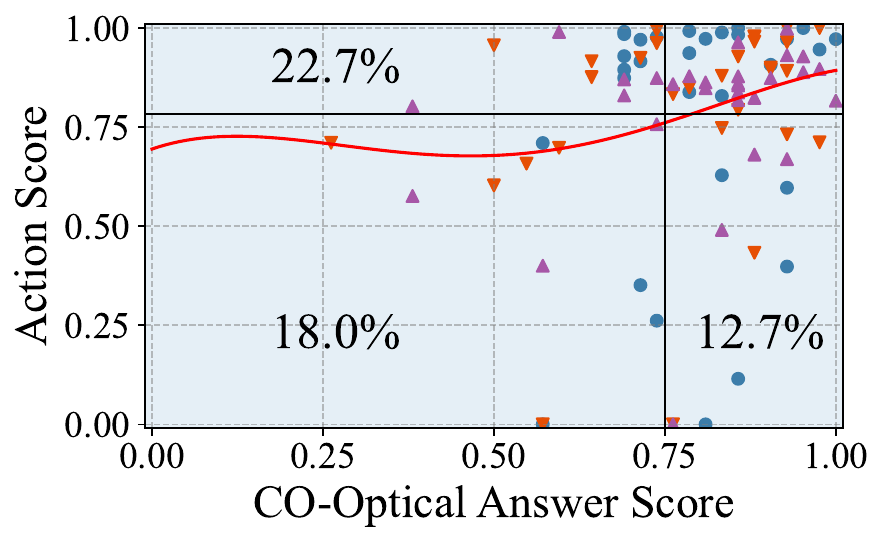}
\centering
\end{minipage}%
}%
\vspace{-2mm}

\subfigure[CST Magnetical]{
\begin{minipage}[t]{0.23\linewidth}
\includegraphics[width=\linewidth]{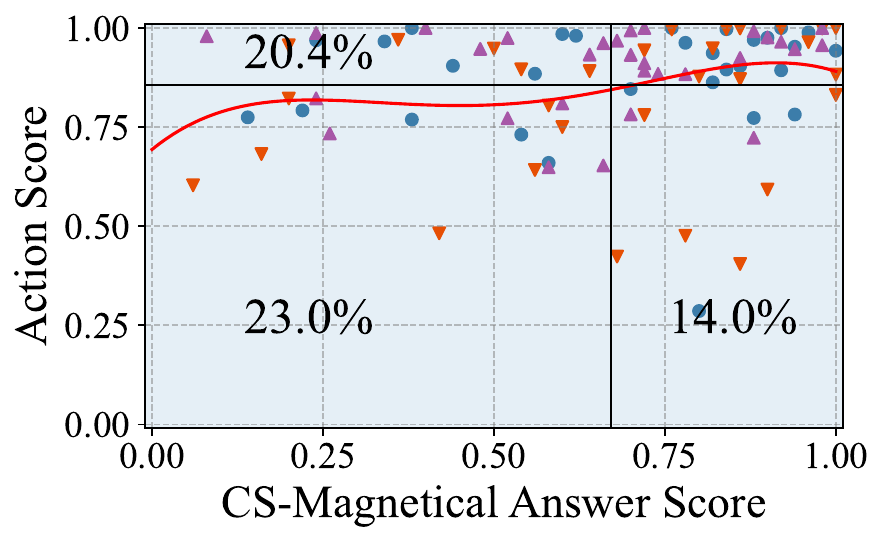}
\centering
\end{minipage}%
}%
\subfigure[Flotherm Thermal]{
\begin{minipage}[t]{0.23\linewidth}
\includegraphics[width=\linewidth]{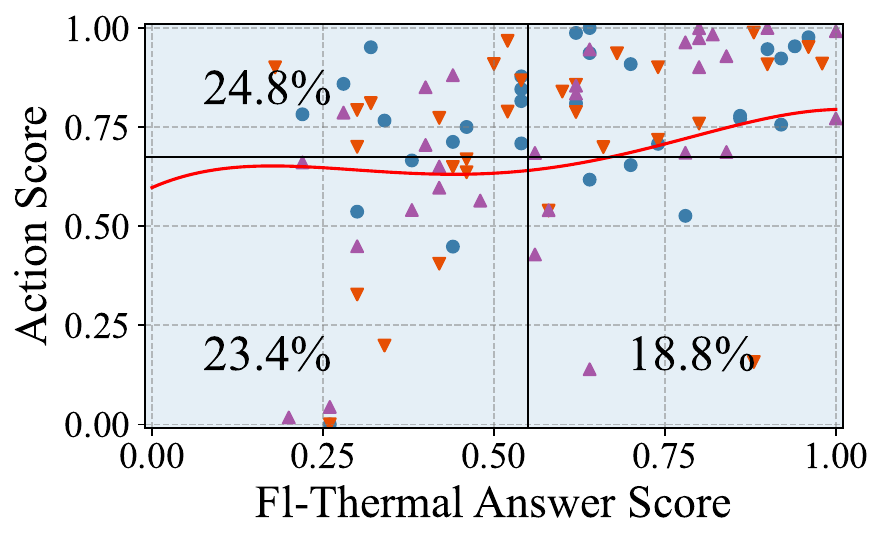}
\centering
\end{minipage}%
}%
\subfigure[HFSS Magnetical]{
\begin{minipage}[t]{0.23\linewidth}
\includegraphics[width=\linewidth]{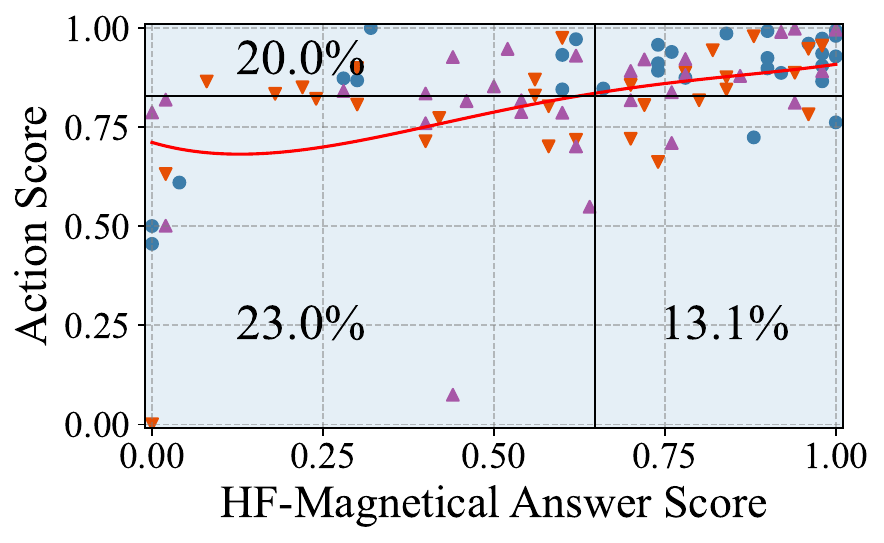}
\centering
\end{minipage}%
}%
\subfigure[ICEPAK Thermal]{
\begin{minipage}[t]{0.23\linewidth}
\includegraphics[width=\linewidth]{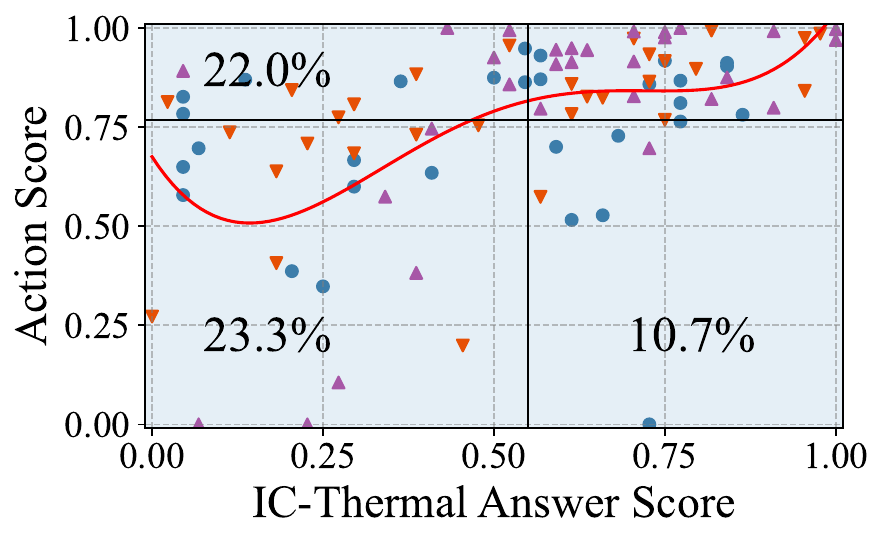}
\centering
\end{minipage}%
}%
\vspace{-2mm}

\caption{The Answer-Action gap in eight Field+Software Combination. The \GroupB{$\bullet$}/\GroupC{$\blacktriangledown$}/\GroupE{$\blacktriangle$} dots represent instance in GUI-EDA with Easy/Normal/Hard Difficulty, and the two straight lines represent the average Answer and Action Scores.
For all Combinations and Difficulty levels, there are numerous instance with low Answer Score but high Action Score, or vice versa.}
\label{fig:axis}
\vspace{-2mm}
\end{figure*}

\section{Proposed Method}

\subsection{Answer \& Action Collaboration}




Our data analysis reveals that for CAD software used in EDA tasks, it's more effective to maximize the capabilities of GUI agents at the semantic level rather than processing the GUI as an image at the pixel level. While there are approaches like GUI-Reflection\cite{relate:gui-reflection}, GUI-Actor\cite{relate:gui-actor}, and LearnAct\cite{relate:learnact} that use interface region cropping and multiple clicks to aid GUI agents in grounding, these approaches rely on pixel-level image processing and are unsuitable for EDA tasks. When a GUI agent is used in software like Word and Excel, the success or failure of Answer-Action is typically identical. However, for CAD software interfaces:

\begin{itemize}
\item If the Agent possesses EDA prior knowledge: This requires abundant specialized training data beyond common sense, which can only be provided by general MLLMs. However, their limited localization capability often results in a correct textual Answer, but the Action pixel is completely inconsistent with the description above. 
\item If the Agent lacks prior knowledge of EDA: A specialized GUI agent could be used. However, due to its lack of familiarity with the EDA knowledge and CAD interface, the Answer would be semantically incorrect, but the Action may barely click the adjacent button.
\end{itemize}

Therefore, combining the comprehension capabilities of MLLMs with the execution capabilities of GUI Agents is crucial for EDA tasks. This `cerebrum+cerebellum' paradigm has been proven effective in concurrent works \cite{relate:mmbenchgui,relate:cerebrum,relate:guir1}, simply linking GPT-4o and UI-TARS can improve the Office suite performance by 20\%. This paradigm has even greater potential for EDA. As shown in Figure \ref{fig:axis}, we calculated the average Answer/Action score for each example in GUI-EDA based on the responses of six advanced Agents and divided them into four phase limits based on the average values:

\begin{itemize}
    \item Phase 1: Answer $>$ Average, Action $>$ Average.
    \item Phase 2: Answer $<$ Average, Action $>$ Average.
    \item Phase 3: Answer $<$ Average, Action $<$ Average.
    \item Phase 4: Answer $>$ Average, Action $<$ Average.
\end{itemize}
where Phase 1 denotes current GUI Agent can successfully solve an EDA sample, and others indicate failure. However, only Phase 3 represents current Agent cannot solve this EDA problem, while the failed samples in Phase 2 and Phase 4 can be transformed into Phase 1 through the `cerebrum+cerebellum' cooperation mechanism. For all eight Field+Software combinations in Figure \ref{fig:axis}, the sum of proportions in Phase 2 and Phase 4 is greater than that of Phase 3, demonstrating the great potential of this mechanism. Most failure samples can be potentially solved by existing MLLM and GUI Agent without introducing new EDA knowledge. Therefore, a method that combines the advantages of such `cerebrum' and `cerebellum' is needed to solve EDA tasks.

\begin{figure*}[t]
    \centering
    \includegraphics[width=\linewidth]{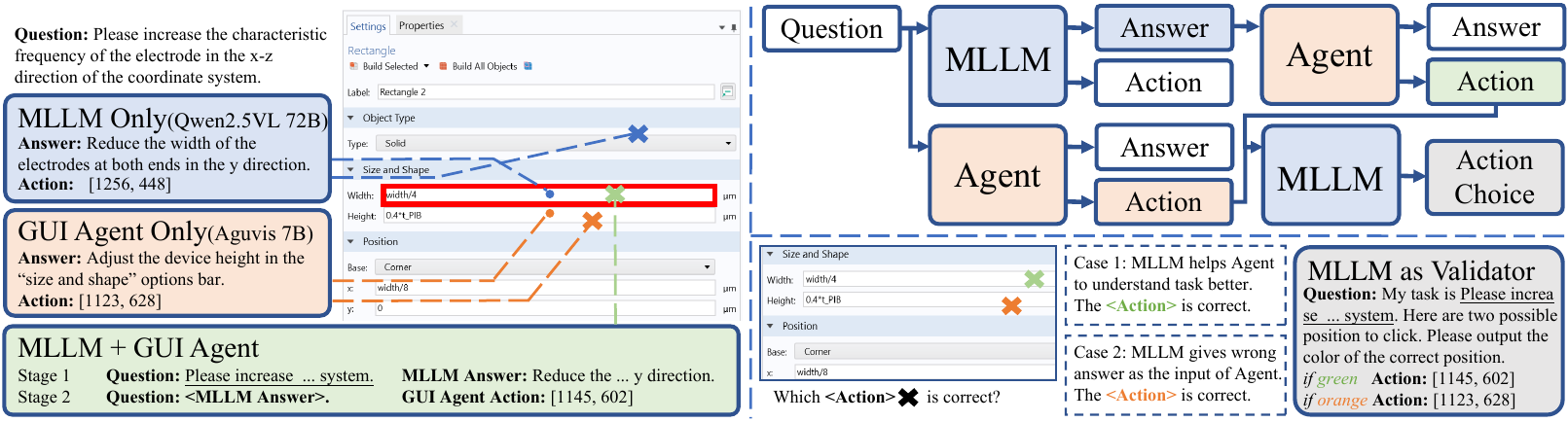}
    \caption{The method framework of EDAgent we proposed. EDAgent comprehensively considers the output of MLLM+GUI Agent and the Agent-only paradigm, leveraging MLLM comprehension while avoiding overthinking.}
    \label{fig:method}
    \vspace{-2mm}
\end{figure*}

\subsection{EDAgent Framework}


Click-error is divided into two parts, one is the comprehension error of the `cerebrum', and the other is the execution error of the `cerebellum'.
Thus we pack both comprehension--bias and execution--noise into a single random vector $\mathbf{e}_{\pi}$:
\begin{equation}
\mathbf{e}_{\pi}\triangleq\hat\mu_{\pi}-\mu
=\underbrace{\bigl(\mathbb{E}[\hat\mu_{\pi}\mid\mu]-\mu\bigr)}_{\textstyle\mathbf{a}_{\pi}}
+\underbrace{\bigl(\hat\mu_{\pi}-\mathbb{E}[\hat\mu_{\pi}\mid\mu]\bigr)}_{\textstyle\mathbf{b}_{\pi}}
\in\mathbb{R}^{2},
\end{equation}
where the $\mu$ is the correct click position and $\hat\mu_{\pi}$ is the actual click position under strategy $\pi$. Thus, the spatial error under strategy $\varepsilon_{\pi}$ can be the expected as squared miss-distance:
\begin{equation}
\label{equ:dir}
\varepsilon_{\pi}
\triangleq\mathbb{E}\|\mathbf{a}_{\pi}+\mathbf{b}_{\pi}\|^{2}
=\mathbb{E}\|\mathbf{a}_{\pi}\|^{2}+\mathbb{E}\|\mathbf{b}_{\pi}\|^{2},
\end{equation}
where
$\mathbf{a}_{\pi}$ is the comprehension-error,
$\mathbf{b}_{\pi}$ is the zero-mean execution-noise. 
Since $\mathbf{b}_{\pi}$ has equal probability of drifting in all directions, the cross term vanishes because $\mathbb{E}[\mathbf{b}_{\pi}]=\mathbf{0}$.
Hence, for MLLM and GUI-Agent strategy we have $\varepsilon_{M},\varepsilon_{G}$:
\begin{equation}
\varepsilon_{\mathrm{M}}=\mathbb{E}\|\mathbf{a}_{\rm M}\|^{2}+\mathbb{E}\|\mathbf{b}_{\rm M}\|^{2},
\qquad
\varepsilon_{\mathrm{G}}=\mathbb{E}\|\mathbf{a}_{\rm G}\|^{2}+\mathbb{E}\|\mathbf{b}_{\rm G}\|^{2}.
\end{equation}
where $\rm M, G$ denote MLLM and GUI-Agent strategies.
Under the $\rm G$ strategy, its comprehension is inferior to the $\rm M$ strategy, resulting in $a_G>a_M$. However, the actual click locations and descriptions of the MLLM are significantly different, while the GUI Agent ensures that `what you see is what you get', i.e., $b_G<<b_M$. Overall, $\varepsilon_{\mathrm{G}}<\varepsilon_{\mathrm{M}}$, making $\rm G$ the better strategy.

Therefore, a hybrid `MLLM talks, GUI-Agent clicks strategy can be applied to combine the strengths of $a_M$ and $b_G$. Because MLLM outputs abstract descriptions in natural language, while GUI-Agent relies on a pixel-level coordinate interface, the two have inherent differences in semantic granularity and representation space. When GUI-Agent maps textual prompts to specific regions, it can produce systematic biases due to over-interpretation or information loss with total error $\varepsilon_{\mathrm{M\!+\!G}}$:
\begin{equation}
\varepsilon_{\mathrm{M\!+\!G}}
=\mathbb{E}\|\mathbf{a}_{\mathrm{M}}+\mathbf{b}_{\mathrm{G}}+\mathbf{c}_{\mathrm{M}}\|^{2},
\end{equation}
where $\mathbf{c}_{\mathrm{M}}$ is the additional error introduced by an additional comprehension-execution mismatch.
Due to the homogeneity of the $\mathbf{b}_{\mathrm{G}}$ distribution, the related terms can be eliminated similar to (\ref{equ:dir}), thereby determining whether the hybrid strategy has positively optimized the GUI Agent through $\Delta$:

\begin{equation}
\label{equ:delta}
\Delta=\varepsilon_{\mathrm{M\!+\!G}}-\varepsilon_{\mathrm{\!G}}=\mathbb{E}\|\mathbf{a}_{\mathrm{M}}+\mathbf{c}_{\mathrm{M}}\|^{2}-\mathbb{E}\|\mathbf{a}_{\rm G}\|^{2},
\end{equation}
where $\mathbf{b}_{\mathrm{G}}(\mathbf{a}_{\mathrm{M}}+\mathbf{c}_{\mathrm{M}})$ is eliminated. The router needs only compare comprehension-side biases:
$\pi^{*}=\mathrm{M\!+\!G}$ if
$\|\mathbf{a}_{\mathrm{M}}+\mathbf{c}_{\mathrm{M}}\|^{2}
<\|\mathbf{a}_{\mathrm{G}}\|^{2}$,
and $\pi^{*}=\mathrm{G}$ otherwise.
yielding the single-threshold rule for applied strategy $\pi^{*}(\Delta)$:
\begin{equation}
\pi^{*}(\Delta)=
\mathrm{M\!+\!G}\cdot\mathbf{1}\bigl(s(\Delta)\ge\tau\bigr),
\end{equation}
where $s(\cdot)$ denotes the self-assessed confidence score. Thus minimize the overall expected error
$\mathbb{E}\|\mathbf{e}_{\pi^{*}(\Delta)}\|^{2}$. As shown in Figure \ref{fig:method}, our EDAgent first use MLLM for semantic comprehension.
Given the user instruction $Q$ and the interface $I$, the MLLM outputs a natural-language description $Ans$:
\begin{equation}
    Ans={\rm F_M}(Q,I)\in\Sigma^{*},
\end{equation}
where $\rm F_{M}(\cdot)$ is the MLLM and $\Sigma^{*}$ denotes the space of natural-language strings.
Conditioned on the textual description, the GUI Agent regresses a normalized click location $(x_0,y_0)$ for strategy $\rm G$ and $(x_1,y_1)$ for strategy $\rm M+G$:
\begin{equation}
\begin{split}
    (x_0,y_0)&={\rm F_G}(Ans,I)\in[0,1]^{2},\\
    (x_1,y_1)&={\rm F_G}(Q,I)\in[0,1]^{2},
\end{split}
\end{equation}
with $\rm F_{G}(\cdot)$ the GUI Agent localization network whose output is already scaled to the image relative coordinates.
The normalized coordinates are mapped back to the original resolution for actual screen interaction as $(\hat{x},\hat{y})$:
\begin{equation}
    (\hat{x},\hat{y})=\bigl(\lfloor xW+0.5\rfloor,\;\lfloor yH+0.5\rfloor\bigr),
\end{equation}
where $W$ and $H$ are the width and height of the screenshot in pixels.
As shown in (\ref{equ:delta}), among alll three variables of $\Delta$, two are determined by $\rm M$, and the remaining $\rm a_G$ represents comprehension. Therefore, $\rm F_M$ is qualified as the confidence score $s(\Delta)$ to select the strategy.
The same MLLM verifies whether the executed click satisfies the original instruction, yielding the final action choice $Act$:
\begin{equation}
\begin{split}
    s_i&=\sigma\bigl( {\rm F_M}\bigl(Q,\;I\odot\delta(\hat{x_i},\hat{y_i})\bigr)_{\rm \{Yes,No\}}\bigr),i \in \{0,1\}\\
    Act&=(s_0>s_1)\cdot(\hat{x_0},\hat{y_0}) + (s_0\leq s_1)\cdot(\hat{x_1},\hat{y_1}) \in\mathbb{Z}^{2},
\end{split}
\end{equation}
with $\delta(\cdot,\cdot)$ a single-impulse mask centered on the clicked pixel. We catch the output probability of `Yes' or `No' text logit from $\rm F_M$, and normalize probability of logits with the sigmoid function $\sigma(\cdot)$, to represent the confidence of the strategy. Thus, leverage MLLM comprehension while avoiding overthinking using MLLM-as-Validator, our EDAgent can choose the better pixel to click under both strategies.

\section{Experiment}
\subsection{Settings}
\label{sec:exp-settings}

GUI-EDA uses both general MLLM and specialized GUI Agent for testing, with 27 strong candidate models in total. We exclude old model (before July 2023) such as VisualGLM \cite{old:visualglm} and InstructBLIP \cite{old:instructblip}, to ensure all chosen MLLMs show excellent performance in past comprehension benchmarks \cite{bench:mmbench,bench:abench,bench:cmcbench,bench:rbench,bench:info,bench:AIBench,bench:large}, and all chosen GUI Agents demonstrate high accuracy in past non-EDA \cite{relate:gui-reflection,relate:osworld,relate:mmbenchgui} tasks.
Specifically, the MLLM open-source candidates have a size from 7B to 70B, and the closed-source candidates call the latest API interface (as of July 2025). GUI Agent is all open source candidates, with a size about 7B. All models are tested as zero-shot, including:

\begin{itemize}
    \item General MLLM: Claude3.7-API \cite{mllm:claude}, Gemini2.5Pro-API \cite{mllm:gemini}, GPT4o-API \cite{mllm:gpt4o}, InternVL2-40B \cite{mllm:internvl2}, InternVL2.5-78B \cite{mllm:internvl25}, InternVL2.5-38B \cite{mllm:internvl25}, Janus-7B \cite{mllm:janus}, Llama3.2-90B \cite{mllm:llama3}, LLaVANext-7B \cite{mllm:llavao}, LLaVAo-72B \cite{mllm:llavao}, MPlugOWL3-7B \cite{mllm:mplugowl3}, Nvlm-70B \cite{mllm:nvlm}, Ovis2-34B \cite{mllm:ovis2}, Phi3.5-7B \cite{mllm:phi35}, Qwen2VL-72B \cite{mllm:qwen2vl}, Qwen2VL-7B \cite{mllm:qwen2vl}, and Qwen2.5VL-72B \cite{mllm:qwen25vl};
    \item Specialized GUI Agent: AriaUI-18B \cite{gui:aria}, Aguvis-7B \cite{gui:aguvis}, CogAgent-9B \cite{gui:cogagent}, OSAtlas-7B \cite{gui:osatlas}, OSAtlasPro-7B \cite{gui:osatlas}, OSGenesis-AC-7B \cite{gui:osgenesis}, SeeClick-7B \cite{gui:seeclick}, ShowUI-2B \cite{gui:showui}, UITARS-7B \cite{gui:uitars}, and the EDAgent proposed in this work.
\end{itemize}

\begin{table*}[t]
\centering
    \renewcommand\arraystretch{1.25}
    \renewcommand\tabcolsep{4pt}
    \belowrulesep=0pt\aboverulesep=0pt
    \caption{The comprehension ability measured by Answer Score, listed by eight Field+Software combination with Original (Ori.) and Dynamic (Dyn.) Resolution. MLLMs ranked higher than GUI Agents. [Keys: \CLB{Best}; \CLA{Second Best}; {\faDesktop} GUI Agent.]
    }
    \label{tab:answer}
    \vspace{-5pt}
    \resizebox{\linewidth}{!}{

\begin{tabular}{l|rr:rr:rr:rr:rr:rr:rr:rr|r}
\toprule
software        & \multicolumn{8}{c:}{COMSOL}               & \multicolumn{2}{c:}{Flotherm}     & \multicolumn{2}{c:}{ICEPAK}       & \multicolumn{2}{c:}{CST}          & \multicolumn{2}{c|}{HFSS}         & \multicolumn{1}{c}{}    \\ \cdashline{1-17}
field           & \multicolumn{2}{c:}{Acoustic}     & \multicolumn{2}{c:}{Optical}      & \multicolumn{2}{c:}{Mechanical}   & \multicolumn{6}{c:}{Thermal}            & \multicolumn{4}{c|}{Magnetical}            & \multicolumn{1}{c}{}    \\ \cdashline{1-17}
group           & \multicolumn{1}{c}{Ori.} & \multicolumn{1}{c:}{Dyn.} & \multicolumn{1}{c}{Ori.} & \multicolumn{1}{c:}{Dyn.} & \multicolumn{1}{c}{Ori.} & \multicolumn{1}{c:}{Dyn.} & \multicolumn{1}{c}{Ori.} & \multicolumn{1}{c:}{Dyn.} & \multicolumn{1}{c}{Ori.} & \multicolumn{1}{c:}{Dyn.} & \multicolumn{1}{c}{Ori.} & \multicolumn{1}{c:}{Dyn.} & \multicolumn{1}{c}{Ori.} & \multicolumn{1}{c:}{Dyn.} & \multicolumn{1}{c}{Ori.} & \multicolumn{1}{c|}{Dyn.} & \multicolumn{1}{c}{\multirow{-3}{*}{Avg.}} \\ \midrule
Qwen2.5VL-72B    & \CLB{0.77}  & \CLB{0.76}  & 0.92  & 0.91  & \CLB{0.75}  & \CLB{0.74}  & \CLA{0.66}  & \CLA{0.68}  & \CLA{0.70}  & 0.73  & 0.70  & \CLA{0.74}  & \CLA{0.80}  & \CLB{0.81}  & \CLA{0.80}  & \CLA{0.79}  & \CLB{0.766}  \\
Qwen2VL-72B     & \CLA{0.71}  & 0.71  & 0.71  & 0.74  & \CLA{0.71}  & \CLA{0.71}  & 0.64  & 0.67  & \CLB{0.71}  & \CLB{0.75}  & \CLB{0.91}  & \CLB{0.89}  & 0.80  & \CLA{0.80}  & \CLB{0.80}  & \CLB{0.80}  & \CLA{0.755}  \\
InternVL2.5-78B  & 0.67  & 0.69  & 0.91  & \CLA{0.92}  & 0.63  & 0.67  & \CLB{0.67}  & \CLB{0.70}  & 0.66  & 0.70  & \CLA{0.71}  & 0.74  & \CLB{0.80}  & 0.80  & 0.76  & 0.77  & 0.738  \\
Ovis2-34B       & 0.69  & 0.69  & \CLB{0.96}  & \CLB{0.94}  & 0.61  & 0.66  & 0.63  & 0.67  & 0.69  & \CLA{0.73}  & 0.60  & 0.67  & 0.76  & 0.80  & 0.74  & 0.76  & 0.724  \\
Qwen2VL-7B      & 0.71  & \CLA{0.71}  & \CLA{0.93}  & 0.90  & 0.68  & 0.70  & 0.62  & 0.63  & 0.62  & 0.64  & 0.66  & 0.69  & 0.78  & 0.78  & 0.74  & 0.75  & 0.722  \\
InternVL2.5-38B  & 0.65  & 0.68  & 0.84  & 0.85  & 0.61  & 0.63  & 0.64  & 0.63  & 0.61  & 0.64  & 0.64  & 0.66  & 0.74  & 0.74  & 0.69  & 0.71  & 0.684  \\
LLaVA-o-72B      & 0.59  & 0.65  & 0.92  & 0.87  & 0.69  & 0.67  & 0.60  & 0.61  & 0.57  & 0.60  & 0.62  & 0.65  & 0.70  & 0.74  & 0.69  & 0.72  & 0.681  \\
InternVL2-40B   & 0.62  & 0.66  & 0.85  & 0.86  & 0.60  & 0.65  & 0.57  & 0.60  & 0.63  & 0.62  & 0.59  & 0.62  & 0.61  & 0.69  & 0.61  & 0.64  & 0.652  \\
Nvlm-70B        & 0.60  & 0.65  & 0.81  & 0.81  & 0.64  & 0.64  & 0.58  & 0.60  & 0.56  & 0.58  & 0.55  & 0.61  & 0.67  & 0.73  & 0.61  & 0.68  & 0.645  \\
Gemini-API      & 0.61  & 0.62  & 0.81  & 0.81  & 0.59  & 0.63  & 0.58  & 0.58  & 0.57  & 0.59  & 0.50  & 0.56  & 0.70  & 0.75  & 0.69  & 0.68  & 0.642  \\
GPT4o-API       & 0.55  & 0.62  & 0.81  & 0.80  & 0.57  & 0.60  & 0.55  & 0.60  & 0.60  & 0.61  & 0.60  & 0.65  & 0.61  & 0.71  & 0.62  & 0.67  & 0.634  \\
Llama3-90B      & 0.59  & 0.59  & 0.81  & 0.84  & 0.61  & 0.64  & 0.53  & 0.57  & 0.59  & 0.60  & 0.54  & 0.58  & 0.67  & 0.69  & 0.62  & 0.64  & 0.632  \\
{\faDesktop} UITARS-7B       & 0.64  & 0.64  & 0.80  & 0.78  & 0.56  & 0.59  & 0.58  & 0.58  & 0.59  & 0.60  & 0.52  & 0.58  & 0.68  & 0.72  & 0.60  & 0.61  & 0.629  \\
{\faDesktop} OSGenesis-AC-7B & 0.51  & 0.59  & 0.73  & 0.75  & 0.57  & 0.63  & 0.55  & 0.57  & 0.52  & 0.55  & 0.50  & 0.58  & 0.69  & 0.73  & 0.61  & 0.65  & 0.608  \\
{\faDesktop} OSAtlas-7B      & 0.53  & 0.56  & 0.67  & 0.66  & 0.61  & 0.63  & 0.47  & 0.50  & 0.52  & 0.53  & 0.56  & 0.56  & 0.68  & 0.69  & 0.59  & 0.63  & 0.587  \\
Phi35-7B        & 0.50  & 0.55  & 0.83  & 0.84  & 0.52  & 0.52  & 0.48  & 0.51  & 0.59  & 0.59  & 0.48  & 0.50  & 0.58  & 0.60  & 0.57  & 0.58  & 0.578  \\
{\faDesktop} OSAtlasPro-7B   & 0.57  & 0.53  & 0.72  & 0.64  & 0.68  & 0.62  & 0.56  & 0.50  & 0.46  & 0.48  & 0.54  & 0.54  & 0.56  & 0.58  & 0.62  & 0.60  & 0.574  \\
{\faDesktop} CogAgent-9B     & 0.53  & 0.63  & 0.73  & 0.72  & 0.54  & 0.57  & 0.42  & 0.49  & 0.50  & 0.49  & 0.48  & 0.56  & 0.64  & 0.68  & 0.55  & 0.55  & 0.568  \\
Claude-API      & 0.45  & 0.59  & 0.73  & 0.77  & 0.47  & 0.60  & 0.48  & 0.56  & 0.53  & 0.58  & 0.30  & 0.50  & 0.47  & 0.64  & 0.54  & 0.62  & 0.551  \\
LLaVANext-7B    & 0.50  & 0.53  & 0.73  & 0.69  & 0.50  & 0.55  & 0.44  & 0.46  & 0.55  & 0.55  & 0.38  & 0.41  & 0.54  & 0.58  & 0.54  & 0.56  & 0.531  \\
MPlugOwl3-7B    & 0.45  & 0.48  & 0.84  & 0.73  & 0.55  & 0.52  & 0.42  & 0.42  & 0.51  & 0.45  & 0.40  & 0.41  & 0.53  & 0.51  & 0.54  & 0.53  & 0.518  \\
Janus-7B        & 0.27  & 0.32  & 0.55  & 0.58  & 0.29  & 0.34  & 0.32  & 0.33  & 0.32  & 0.34  & 0.29  & 0.26  & 0.41  & 0.40  & 0.29  & 0.31  & 0.350   \\ \bottomrule              
\end{tabular}
    }
    \vspace{-3mm}
\end{table*}

In addition to machine intelligence, we also introduced five human experts, five average users, and five random guesses to the Action task as described in Section \ref{sec:human} for paralleled comparison. Noted that for both human and machine intelligence, the GUI-EDA data is inferenced in a random order to prevent any prior knowledge of the Field/Software from instilling familiarity and leading to exaggerated performance.

For evaluation criteria, we adopt the widely-used LLM-as-a-Judge paradigm for Answer: GPT-4o \cite{mllm:gpt4o} compares the model textual output with the GT description in terms of precision and recall, assigns a score of 0/0.5/1 and average the result over five independent runs. For Action: we simply check whether the predicted (x, y) coordinates fall inside the GT bounding box and assign a binary 0/1.
For GUI Agents whose output format is fixed as coordinates, we exclude their Answer Scores.
To better visualize the four factors of GUI-EDA, we list each Field and Software vertically; Resolution Large is labeled `Original', while re-sampled Middle and Small are grouped as `Dynamic'; Difficulty is conveyed by horizontal rows: Human (average)—able to solve Easy items, and Human (expert)—able to solve both Easy and Normal items.

All GUI screenshots are captured on a standard 4K UHD monitor under Windows 10 environment to match real-world CAD usage. MLLM and GUI Agent inference are carried out on a server with 8$\times$NVIDIA A800 SXM4 80GB GPU. Silicon validation is performed in a 65-nm PDK process with on-chip LDO at 1.2 V and 27 °C, applying transient AC sweep to the fabricated EDA devices.


\subsection{Benchmark Result and Discussion}

For Comprehension ability, the Action Score in Table \ref{tab:answer} demonstrate a clear performance hierarchy: \textbf{Current model has the initial ability to understand EDA tasks}, and the average score of most candidates exceed 0.6. In general, 
\textbf{MLLMs uniformly surpass specialized GUI Agents}, with an average Answer Score margin of approximately 0.12. Within both families, parameter scaling matters: 70–78 B models (e.g., Qwen2.5-VL-72B, 0.766) outscore their 7–9 B counterparts (e.g., Qwen2-VL-7B, 0.722) by roughly 0.04–0.06, indicating that increased model capacity improves visual-linguistic comprehension. Replacing original images with dynamically down-sampled counterparts produces only marginal gains—grand mean rises, suggesting that \textbf{Resolution reduction has a positive but limited effect for Answer Score}.
Notably, the boost is inconsistent: Fl-Thermal improve by 0.03, CS-Magnetical remain flat, implying that architectural design rather than pixel count dominates comprehension performance. 

\begin{table*}[t]
\centering
    \renewcommand\arraystretch{1.25}
    \renewcommand\tabcolsep{4pt}
    \belowrulesep=0pt\aboverulesep=0pt
    \caption{The execution ability measured by Action Score, listed by eight Field+Software combination with Original (Ori.) and Dynamic (Dyn.) Resolution. GUI Agents ranked higher than MLLMs, where most GUI Agents can reach human average level, while our proposed EDAgent surpasses human expert for the first time. [Keys: \CLB{Best}; \CLA{Second Best}; {\faDesktop} GUI Agent]}
    \label{tab:action}
    \vspace{-5pt}
    \resizebox{\linewidth}{!}{
\begin{tabular}{l|rr:rr:rr:rr:rr:rr:rr:rr|r}
\toprule
software        & \multicolumn{8}{c:}{COMSOL}               & \multicolumn{2}{c:}{Flotherm}     & \multicolumn{2}{c:}{ICEPAK}       & \multicolumn{2}{c:}{CST}          & \multicolumn{2}{c|}{HFSS}         & \multicolumn{1}{c}{}    \\ \cdashline{1-17}
field           & \multicolumn{2}{c:}{Acoustic}     & \multicolumn{2}{c:}{Optical}      & \multicolumn{2}{c:}{Mechanical}   & \multicolumn{6}{c:}{Thermal}            & \multicolumn{4}{c|}{Magnetical}            & \multicolumn{1}{c}{}    \\ \cdashline{1-17}
group           & \multicolumn{1}{c}{Ori.} & \multicolumn{1}{c:}{Dyn.} & \multicolumn{1}{c}{Ori.} & \multicolumn{1}{c:}{Dyn.} & \multicolumn{1}{c}{Ori.} & \multicolumn{1}{c:}{Dyn.} & \multicolumn{1}{c}{Ori.} & \multicolumn{1}{c:}{Dyn.} & \multicolumn{1}{c}{Ori.} & \multicolumn{1}{c:}{Dyn.} & \multicolumn{1}{c}{Ori.} & \multicolumn{1}{c:}{Dyn.} & \multicolumn{1}{c}{Ori.} & \multicolumn{1}{c:}{Dyn.} & \multicolumn{1}{c}{Ori.} & \multicolumn{1}{c|}{Dyn.} & \multicolumn{1}{c}{\multirow{-3}{*}{Avg.}} \\ \midrule
{\faDesktop} EDAgent        & \CLB{0.66}  & \CLB{0.71}  & \CLB{0.72}  & \CLB{0.75}  & \CLB{0.74}  & \CLB{0.73}  & \CLB{0.47}  & \CLB{0.62}  & \CLB{0.37}  & \CLB{0.46}  & \CLB{0.30}  & \CLB{0.45}  & \CLB{0.73}  & \CLB{0.76}  & \CLB{0.51}  & \CLB{0.58}  & \CLB{0.598}   \\ \cdashline{1-18}
Human (expert)    &	0.52 &	0.55&	0.61&	0.62&	0.62&	0.64&	 0.46&	0.49&	0.24&	0.31&	0.46&	0.48&	0.32&	0.38&	0.28&	0.36&	0.459 \\ \cdashline{1-18}
{\faDesktop} Aguvis-7B       & 0.40  & \CLA{0.53}  & \CLA{0.56}  & \CLA{0.58}  & 0.48  & 0.51  & 0.24  & 0.36  & 0.17  & 0.26  & 0.10  & 0.27  & \CLA{0.49}  & \CLA{0.59}  & 0.33  & 0.44  & 0.394   \\
{\faDesktop} OSAtlasPro-7B   & 0.32  & 0.51  & 0.36  & 0.50  & \CLA{0.50}  & \CLA{0.59}  & 0.29  & \CLA{0.37}  & 0.18  & 0.23  & 0.10  & 0.29  & 0.36  & 0.40  & 0.23  & 0.32  & 0.347   \\
{\faDesktop} UITARS-7B       & \CLA{0.42}  & 0.43  & 0.30  & 0.37  & 0.42  & 0.45  & \CLA{0.29}  & 0.34  & 0.12  & 0.23  & \CLA{0.18}  & 0.30  & 0.47  & 0.47  & 0.37  & 0.38  & 0.346   \\
{\faDesktop} CogAgent-9B     & 0.32  & 0.52  & 0.22  & 0.45  & 0.38  & 0.55  & 0.18  & 0.35  & 0.12  & 0.21  & 0.14  & \CLA{0.33}  & 0.42  & 0.51  & 0.37  & \CLA{0.44}  & 0.345   \\
{\faDesktop} Aria-18B        & 0.34  & 0.49  & 0.24  & 0.38  & 0.44  & 0.51  & 0.09  & 0.31  & \CLA{0.19}  & \CLA{0.29}  & 0.00  & 0.19  & 0.47  & 0.54  & \CLA{0.38}  & 0.42  & 0.330   \\
{\faDesktop} OSAtlas-7B      & 0.20  & 0.25  & 0.14  & 0.20  & 0.28  & 0.31  & 0.14  & 0.19  & 0.09  & 0.14  & 0.12  & 0.21  & 0.38  & 0.36  & 0.17  & 0.23  & 0.214   \\ \cdashline{1-18}
Human (average)     & 0.20 &	0.27 &	0.10 &	0.18 &	0.14 &	0.21 &	0.22 &	0.26 &	0.10 &	0.13 &	0.13 &	0.17 &	0.12 &	0.17 &	0.11 &	0.14 &	0.166 \\ \cdashline{1-18}
{\faDesktop} Show-2B         & 0.16  & 0.21  & 0.08  & 0.13  & 0.16  & 0.19  & 0.06  & 0.14  & 0.09  & 0.13  & 0.00  & 0.11  & 0.35  & 0.35  & 0.08  & 0.17  & 0.151   \\
Qwen2VL-72B     & 0.06  & 0.27  & 0.08  & 0.23  & 0.14  & 0.29  & 0.13  & 0.25  & 0.05  & 0.14  & 0.04  & 0.15  & 0.05  & 0.26  & 0.07  & 0.19  & 0.149   \\ 
Qwen25VL-72B    & 0.06  & 0.19  & 0.04  & 0.20  & 0.06  & 0.21  & 0.08  & 0.18  & 0.07  & 0.18  & 0.00  & 0.19  & 0.16  & 0.25  & 0.05  & 0.17  & 0.131   \\
Gemini-api      & 0.06  & 0.12  & 0.06  & 0.09  & 0.04  & 0.15  & 0.06  & 0.15  & 0.07  & 0.13  & 0.00  & 0.10  & 0.09  & 0.24  & 0.04  & 0.08  & 0.093   \\
{\faDesktop} SeeClick-7B     & 0.02  & 0.14  & 0.00  & 0.14  & 0.00  & 0.19  & 0.01  & 0.13  & 0.01  & 0.12  & 0.02  & 0.14  & 0.06  & 0.24  & 0.03  & 0.15  & 0.087   \\
InternVL25-78B  & 0.02  & 0.16  & 0.04  & 0.15  & 0.04  & 0.16  & 0.03  & 0.17  & 0.02  & 0.08  & 0.00  & 0.05  & 0.05  & 0.14  & 0.03  & 0.06  & 0.075   \\
{\faDesktop} OSGenesis-AC-7B & 0.08  & 0.07  & 0.08  & 0.10  & 0.06  & 0.08  & 0.04  & 0.06  & 0.02  & 0.03  & 0.02  & 0.06  & 0.05  & 0.12  & 0.04  & 0.05  & 0.060   \\
LLama3-90B      & 0.06  & 0.10  & 0.02  & 0.05  & 0.06  & 0.12  & 0.01  & 0.06  & 0.02  & 0.08  & 0.00  & 0.06  & 0.06  & 0.14  & 0.03  & 0.06  & 0.058   \\
InternVL25-38B  & 0.04  & 0.09  & 0.04  & 0.10  & 0.00  & 0.05  & 0.03  & 0.08  & 0.02  & 0.06  & 0.02  & 0.06  & 0.03  & 0.09  & 0.04  & 0.06  & 0.051   \\
Ovis2-34B       & 0.00  & 0.11  & 0.02  & 0.07  & 0.02  & 0.09  & 0.04  & 0.08  & 0.04  & 0.06  & 0.00  & 0.06  & 0.02  & 0.11  & 0.03  & 0.06  & 0.050   \\
GPT4o-api       & 0.06  & 0.08  & 0.00  & 0.03  & 0.06  & 0.07  & 0.02  & 0.08  & 0.02  & 0.06  & 0.00  & 0.03  & 0.02  & 0.08  & 0.01  & 0.04  & 0.042   \\
Qwen2VL-7B      & 0.04  & 0.08  & 0.00  & 0.03  & 0.06  & 0.10  & 0.02  & 0.04  & 0.03  & 0.05  & 0.00  & 0.02  & 0.02  & 0.09  & 0.02  & 0.05  & 0.041   \\
LLaVA-o-72B     & 0.02  & 0.06  & 0.02  & 0.07  & 0.02  & 0.07  & 0.02  & 0.07  & 0.02  & 0.05  & 0.00  & 0.03  & 0.00  & 0.09  & 0.01  & 0.03  & 0.037   \\
Claude-api      & 0.04  & 0.04  & 0.00  & 0.03  & 0.00  & 0.07  & 0.01  & 0.03  & 0.00  & 0.04  & 0.04  & 0.03  & 0.02  & 0.10  & 0.01  & 0.04  & 0.031   \\
Nvlm-70B        & 0.02  & 0.05  & 0.02  & 0.05  & 0.00  & 0.03  & 0.00  & 0.06  & 0.03  & 0.06  & 0.00  & 0.02  & 0.02  & 0.11  & 0.01  & 0.03  & 0.031   \\
MPlugOwl3-7B    & 0.00  & 0.03  & 0.00  & 0.02  & 0.00  & 0.03  & 0.00  & 0.03  & 0.00  & 0.03  & 0.00  & 0.00  & 0.01  & 0.06  & 0.01  & 0.02  & 0.015   \\
LLaVA-Next-7B   & 0.00  & 0.00  & 0.00  & 0.03  & 0.02  & 0.03  & 0.00  & 0.03  & 0.02  & 0.02  & 0.00  & 0.01  & 0.02  & 0.03  & 0.01  & 0.01  & 0.014   \\
InternVL2-7B    & 0.02  & 0.01  & 0.00  & 0.00  & 0.00  & 0.01  & 0.01  & 0.03  & 0.00  & 0.01  & 0.00  & 0.02  & 0.00  & 0.03  & 0.00  & 0.01  & 0.010   \\ \cdashline{1-18}
Random Guess    & 0.00  & 0.02  & 0.00  & 0.02  & 0.00  & 0.02  & 0.00  & 0.02  & 0.00  & 0.01  & 0.00  & 0.02  & 0.00  & 0.02  & 0.00  & 0.01  & 0.009   \\ \cdashline{1-18}
Janus-7B        & 0.00  & 0.03  & 0.00  & 0.03  & 0.00  & 0.02  & 0.00  & 0.01  & 0.00  & 0.02  & 0.00  & 0.00  & 0.00  & 0.02  & 0.01  & 0.01  & 0.009   \\
Phi35-7B        & 0.00  & 0.02  & 0.00  & 0.01  & 0.00  & 0.01  & 0.00  & 0.01  & 0.00  & 0.01  & 0.00  & 0.00  & 0.00  & 0.03  & 0.01  & 0.02  & 0.008               \\   \bottomrule            
\end{tabular}
    }
    \vspace{-3mm}
\end{table*}

For Execution ability, the Action Score in Table \ref{tab:action} shows that \textbf{current models remain markedly weaker at acting than understanding}. The highest mean (0.598) falls well below the comprehension ceiling, and most large MLLMs hover near or under 0.15, revealing a pronounced grounding gap when clicks must replace answers.
Across the eight field-software pairs, difficulty rises from the spacious buttons of CO-Acoustic (0.45) to the densely packed numeric tables of IC-Thermal. \textbf{GUI Agents consistently outperform MLLMs in every column}: EDAgent, Aguvis-7B, and other Action-specialized architectures claim the top five rows, while the best 72B MLLM manages barely a third of their score. The agents explicit action-token alignment and widget-locality priors evidently outweigh the raw reasoning capacity of billion-parameter MLLMs once physical interaction is required.
\textbf{Dynamic resolution, by contrast, delivers a clear and reproducible boost to Action Score}. Individual gains are largest where icons or cells become frustratingly small: Aguvis adds +0.17 on IC-Thermal, and Aria-18B even climbs from 0.00 to 0.19 on CS-Magnetical after the same resize. Enlarged effective widget footprints ease pixel-level regression, confirming that resolution reduction distinctly helps execution without reordering the GUI-first hierarchy.

Therefore, in a context where EDA execution is far more difficult than comprehension,
our proposed EDAgent establishes a new state-of-the-art in execution, \textbf{attaining an aggregate Action Score of 0.598 and, for the first time, surpassing the Human expert level} (0.459) by a statistically significant margin of 0.139. The gain is not uniform across domains: the largest advantage emerges in IC-Thermal, where EDAgent climbs from 0.47 (Original) to 0.62 (Dynamic) while the expert remains at 0.31, confirming the model efficacy in high-density, scroll-heavy tables that historically crippled both MLLMs and competing GUI agents. Substantial leads are also observed in CO-Acoustic (+0.14) and CS-Magnetical (+0.17), indicating robustness to disparate widget sizes and multi-tab workflows. Conversely, the smallest margin is recorded in CO-Optical (+0.13), a domain already saturated by human performance; here, further improvement is bounded by ceiling effects rather than architectural limitations. Overall, EDAgent widget-aware action vocabulary and resolution-agnostic policy yield super-human execution accuracy, with residual headroom in uncommon EDA toolkits such as ICEPAK.

\begin{figure*}[t]
    \centering
    \subfigure[Difficulty (Hard)]{
    \begin{minipage}[t]{0.29\linewidth}
    \includegraphics[width=\linewidth]{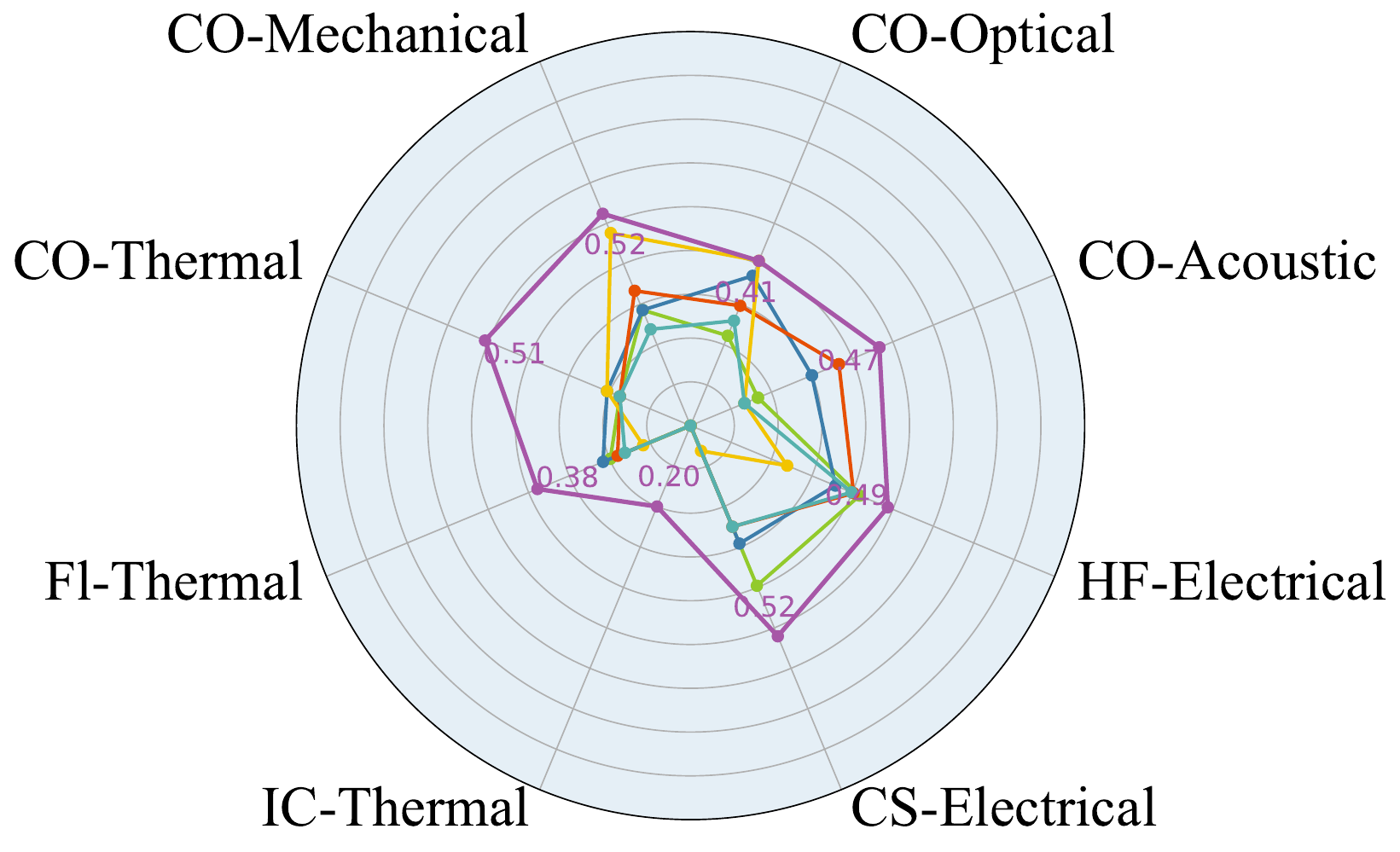}
    \centering
    \end{minipage}
    }%
    \subfigure[Difficulty (Normal)]{
    \begin{minipage}[t]{0.29\linewidth}
    \includegraphics[width=\linewidth]{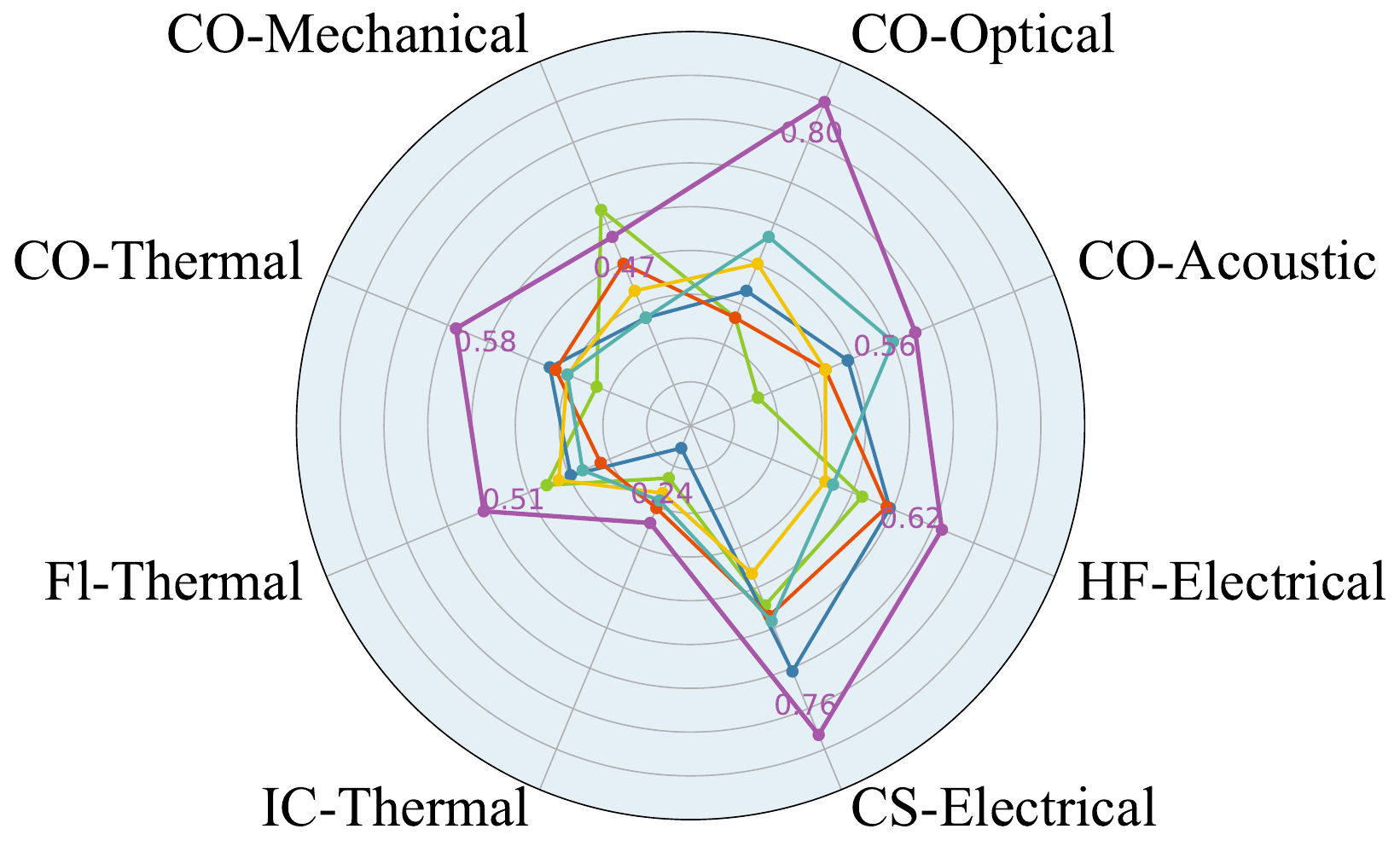}
    \centering
    \end{minipage}%
    }%
    \subfigure[Difficulty (Easy)]{
    \begin{minipage}[t]{0.4\linewidth}
    \includegraphics[width=\linewidth]{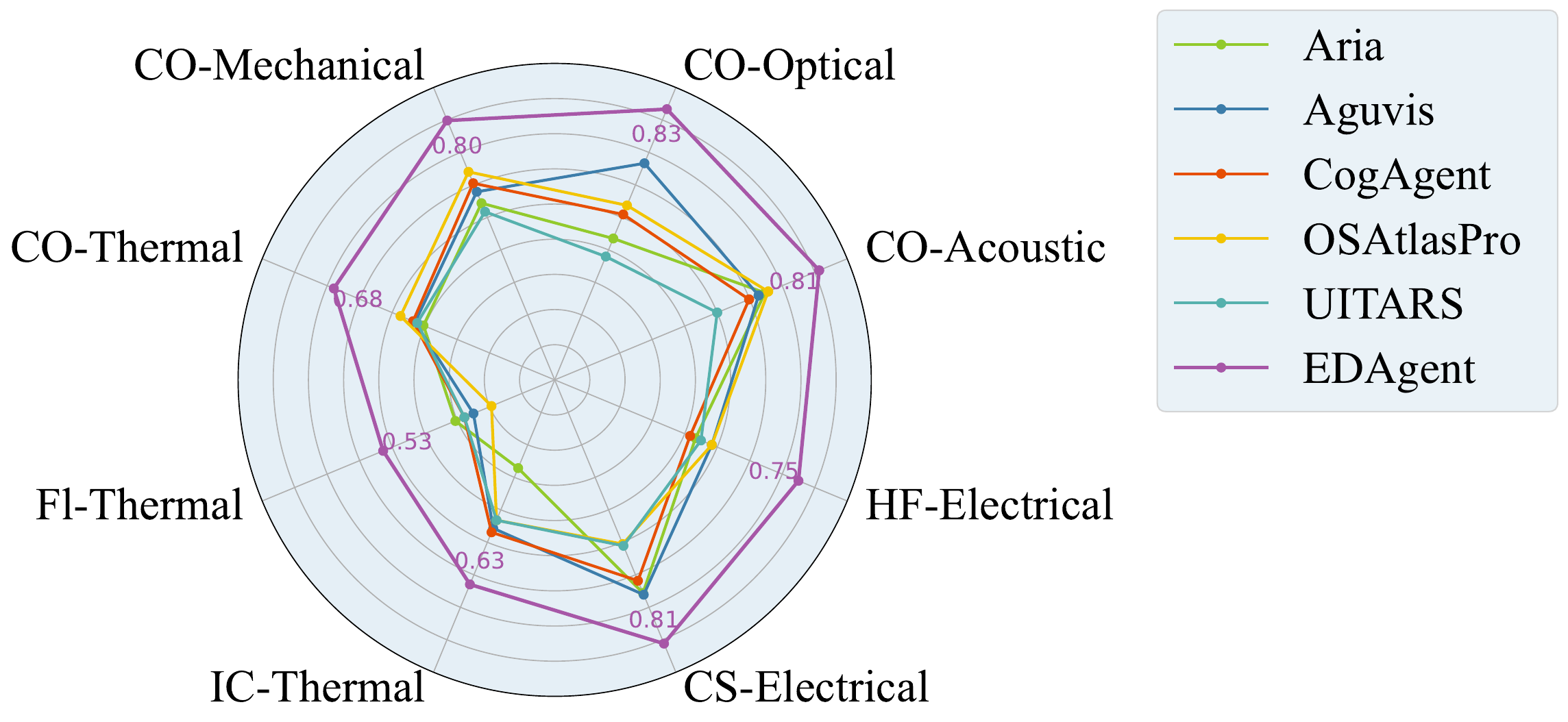}
    \centering
    \end{minipage}%
    }%

    \subfigure[Resolution (Large)]{
    \begin{minipage}[t]{0.29\linewidth}
    \includegraphics[width=\linewidth]{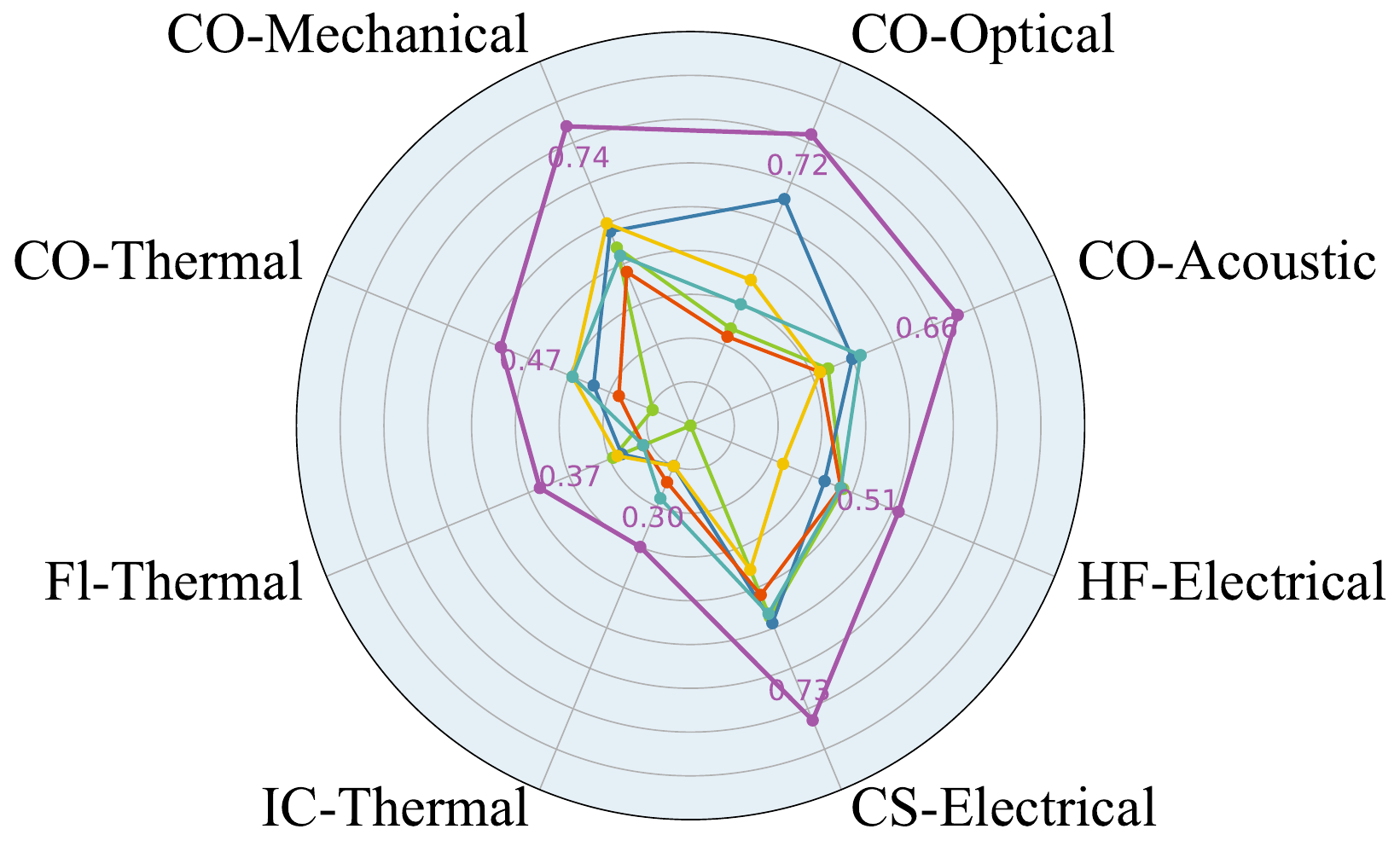}
    \centering
    \end{minipage}
    }%
    \subfigure[Resolution (Medium)]{
    \begin{minipage}[t]{0.29\linewidth}
    \includegraphics[width=\linewidth]{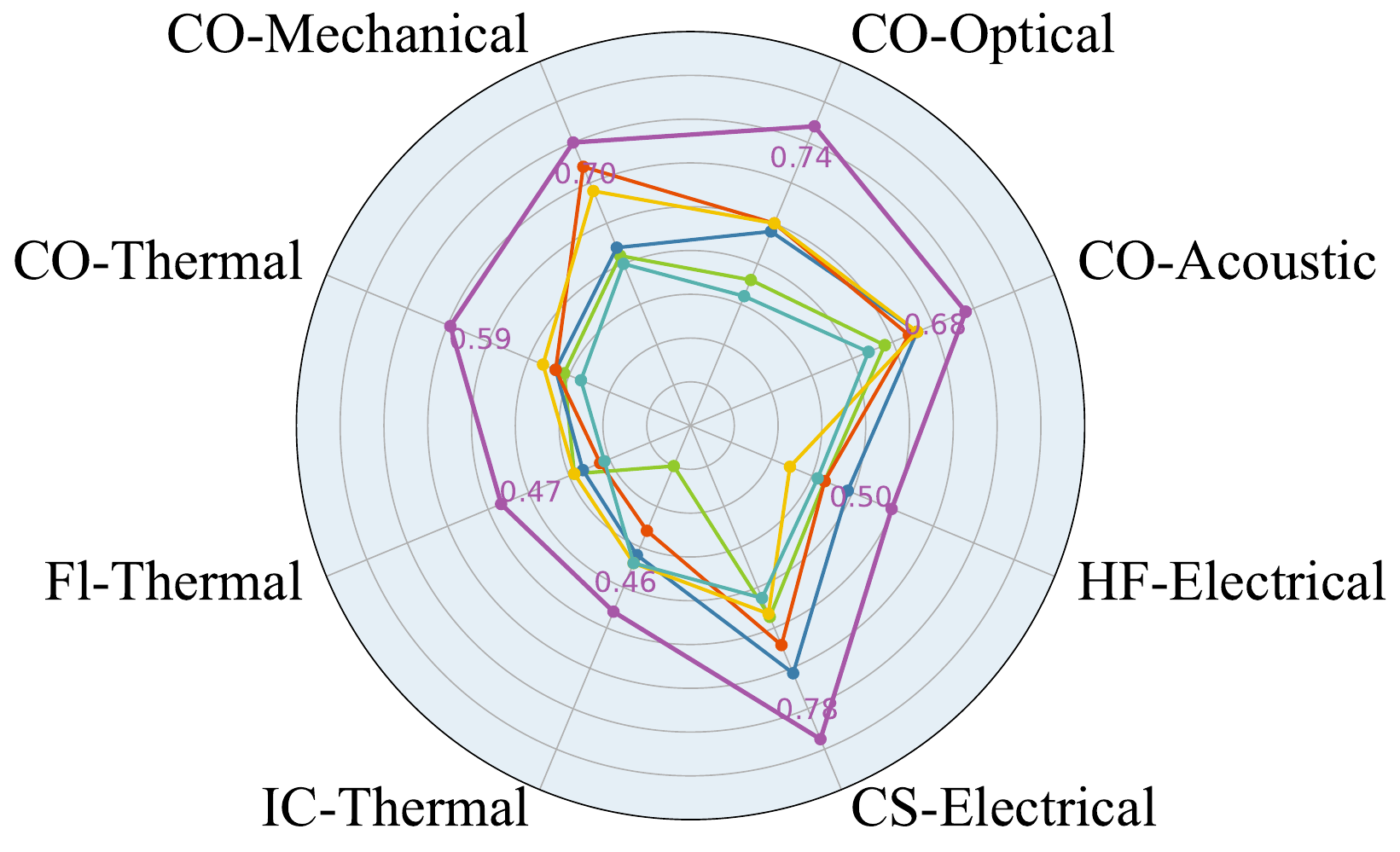}
    \centering
    \end{minipage}%
    }%
    \subfigure[Resolution (Small)]{
    \begin{minipage}[t]{0.4\linewidth}
    \includegraphics[width=\linewidth]{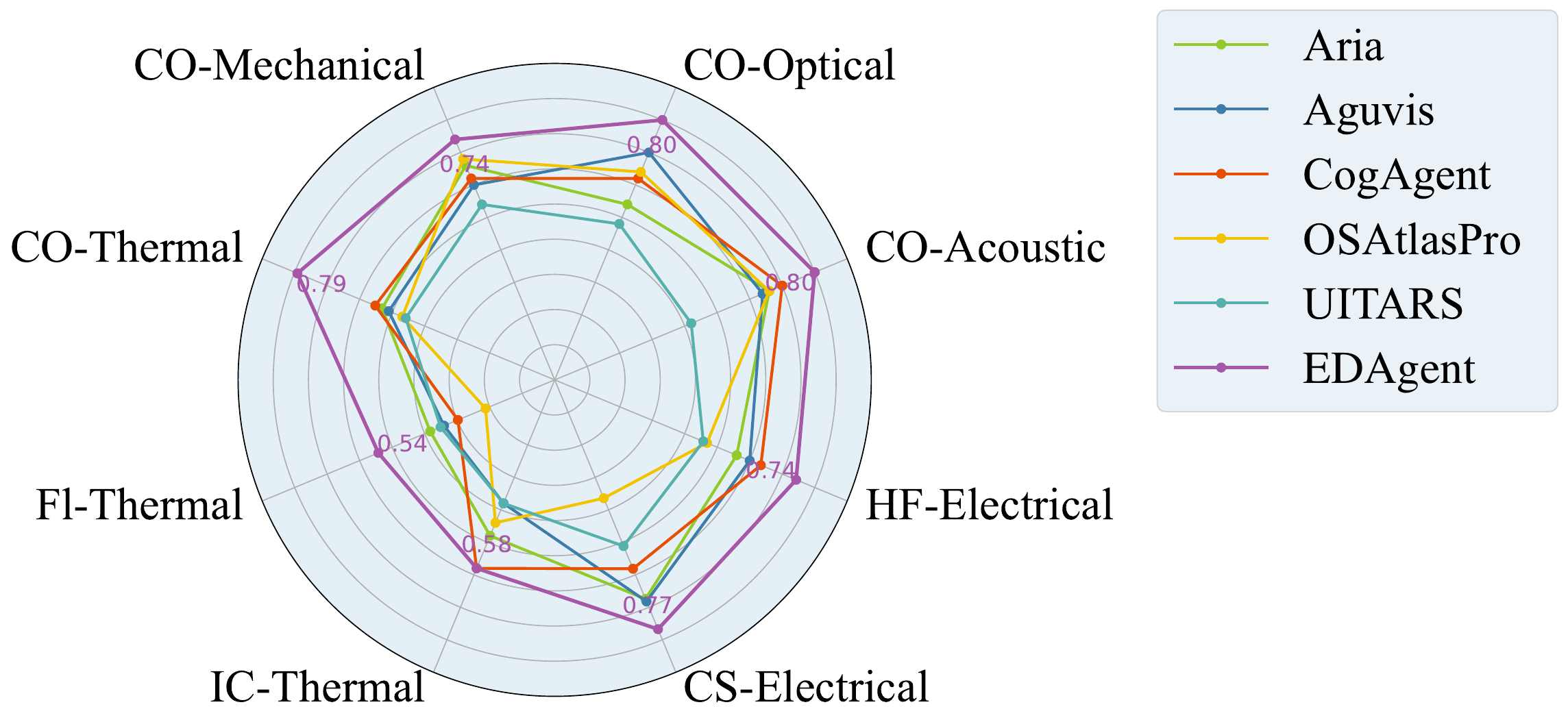}
    \centering
    \end{minipage}%
    }%
    \caption{The Action Score radar map for the eight Field+Software combination. Compared with the other five GUI Agents, our EDAgent exhibits significant advantages in EDA tasks with three Resolutions and Difficulty levels. EDAgent leads on every Subfigure, with the most significant lead on the challenging (a) and (d). (Radar axis range 0-0.9, each gird denotes 0.1)}
    \label{fig:radar}
    \vspace{-2mm}
\end{figure*}


\subsection{Proposed EDAgent Analysis}

This section dissects why the proposed EDAgent outperforms prior models on EDA tasks, and distills design cues for future EDA-purpose GUI Agents. We first benchmark the top-ranked models from Table \ref{tab:action} across the three Resolution–Difficulty level grid, then combine qualitative case with quantitative study to pinpoint exactly where our EDAgent gains its edge against other general-propose GUI Agents.

Figure \ref{fig:radar} presents a six-panel radar sweep of Action Score across the eight Field+Software combinations. Three upper columns stratify question Difficulty (Hard → Easy), while lower columns stratify inference Resolution (Large → Small); five GUI Agents selected for comparison (Aria, Aguvis, CogAgent, OSAtlasPro, UITARS) consistently populate the mid-lobes, leaving the EDAgent trace even overshoot the peripheral 0.9 ring.
At a glance, EDAgent encloses the largest area in every single hexagon, but the margin is not uniform; it widens precisely where competitors collapse. In the Difficulty (Hard, a) cell, the nearest rival peaks at 0.1 (CO-Thermal), whereas EDAgent achieves 0.51; when all models cannot solve the problem (IC-Thermal), EDAgent still maintains 0.20, yielding a statistically significant advantage. The same pattern repeats for HF-Electrical under Resolution (Small, d): EDAgent scores 0.74 versus 0.50 for the second-best model in CO-Mechanical, and 0.73 versus 0.50 in CS-Electrical, confirming challenging question and full-resolution interface domains are the primary source of superiority.
Conversely, the smallest advantage appears in (c) and (f). For such easy tasks, the existing baseline is already saturated; Though exceed 0.80, EDAgent only shows +0.05 superiority above Aguvis and OSAtlasPro, indicating spacious icons and linear workflows offer limited headroom for any policy.
Between extremes, the radar trace reveals three strengths of the proposed EDAgent:
\begin{itemize}
    \item Resolution elasticity: expanding interface from cropped Small size to original Large resolution degrades EDAgent mean by merely 10\%, while the pooled competitor mean drops 30\%, evidencing a invariance to pixel budget.
    \item Difficulty robustness: from Hard to Easy, the inter-lobe range is around 0.4 for all models. But due to the high baseline in Easy, EDAgent still preserves acceptable performance even under comprehensive CS-Electrical dialogs and dense FL-Thermal tables in Hard Subfigure.
    \item Phase continuity: Aside from the more challenging IC-Thermal, the polygons rendered by EDAgent are all convex, while other models may be concave in Fl-Thermal or HF-Electrical. Thus, EDAgent is well-balanced across multiple dimensions with no significant weaknesses.
\end{itemize}
Collectively, the radar map demonstrates that EDAgent not only \textbf{achieves the highest absolute Action Score in all $8\times6=48$ condition-phases}, but also preserves its edge where Difficulty and Resolution impact are jointly minimized, a reliability signature that prior GUI Agents failed to exhibit.

\begin{figure*}[t]
\centering

\subfigure[GT]{
\begin{minipage}[t]{0.24\linewidth}
\includegraphics[width=\linewidth]{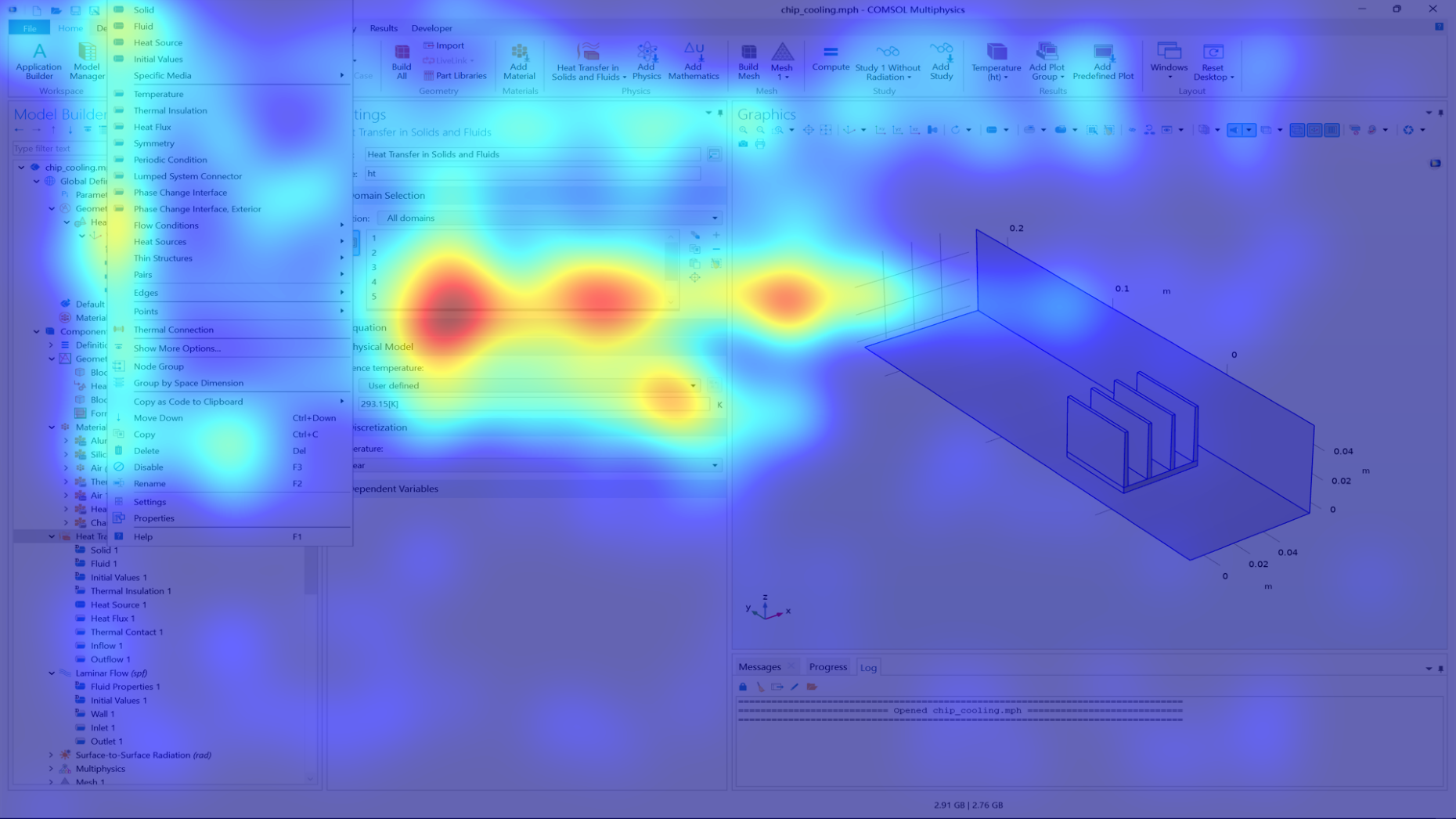}
\centering
\end{minipage}%
}%
\subfigure[Aguvis]{
\begin{minipage}[t]{0.24\linewidth}
\includegraphics[width=\linewidth]{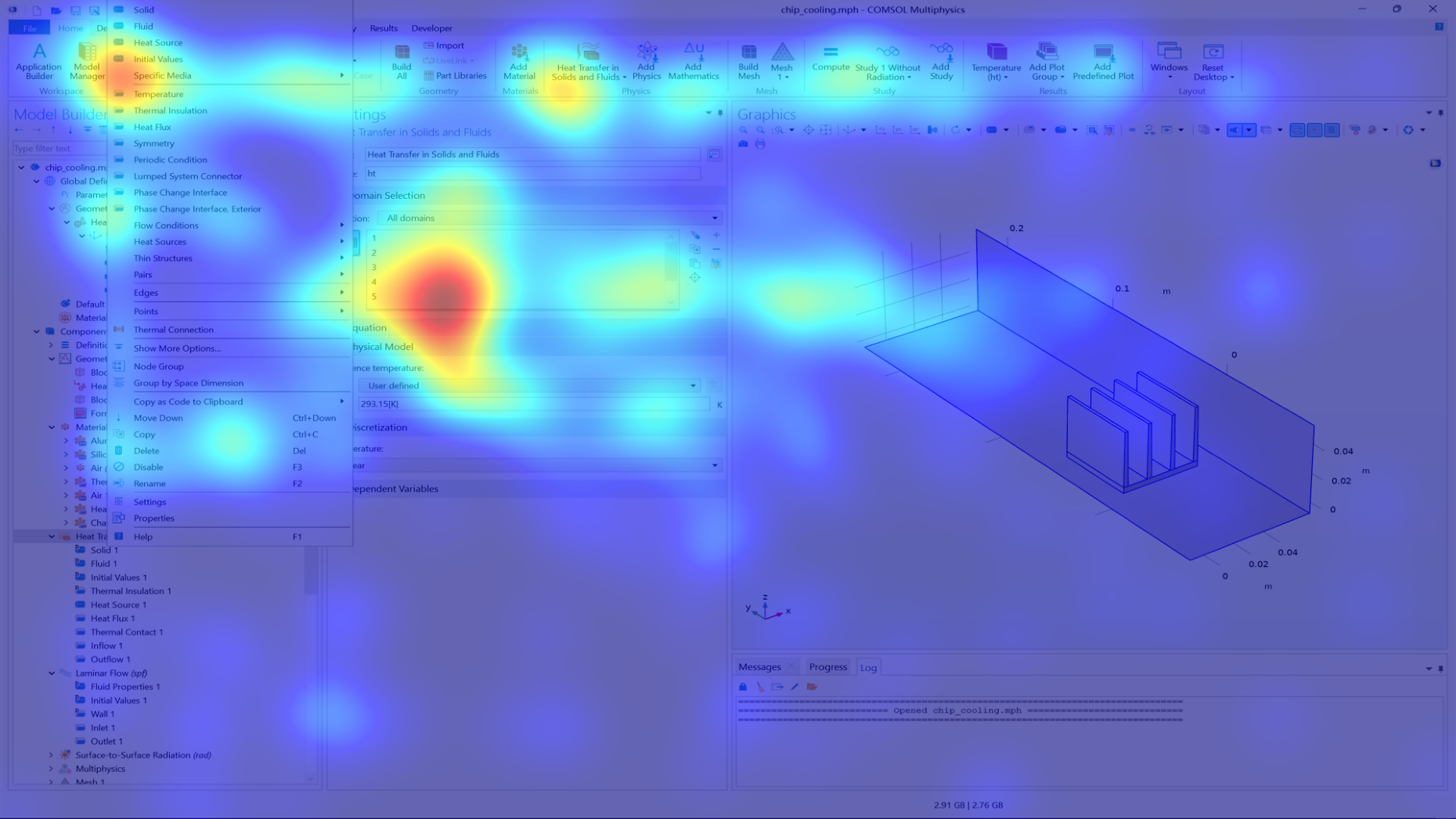}
\centering
\end{minipage}%
}%
\subfigure[Aria-UI]{
\begin{minipage}[t]{0.24\linewidth}
\includegraphics[width=\linewidth]{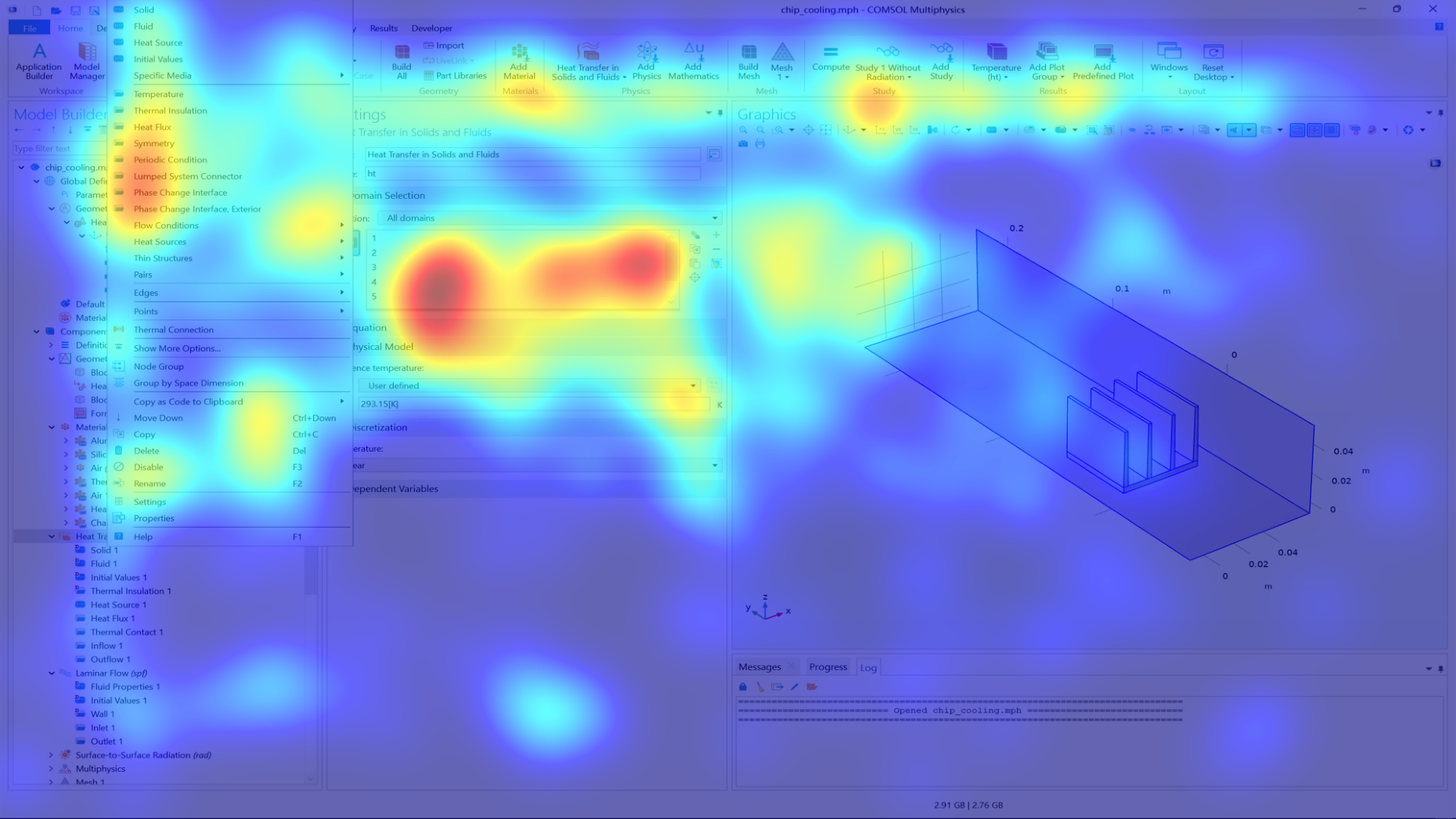}
\centering
\end{minipage}%
}%
\subfigure[CogAgent]{
\begin{minipage}[t]{0.24\linewidth}
\includegraphics[width=\linewidth]{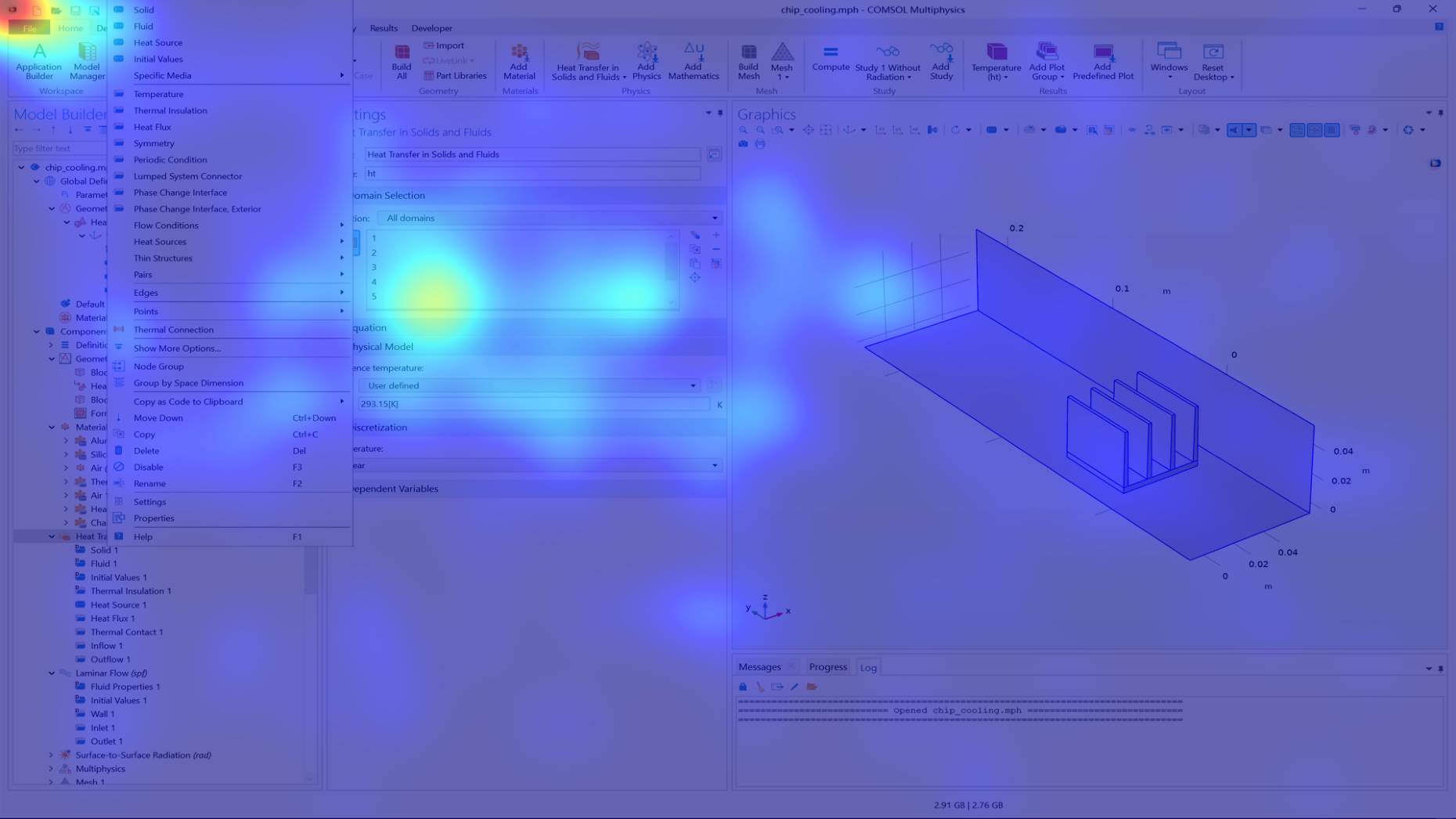}
\centering
\end{minipage}%
}
\vspace{-2mm}

\subfigure[OSAtlas]{
\begin{minipage}[t]{0.24\linewidth}
\includegraphics[width=\linewidth]{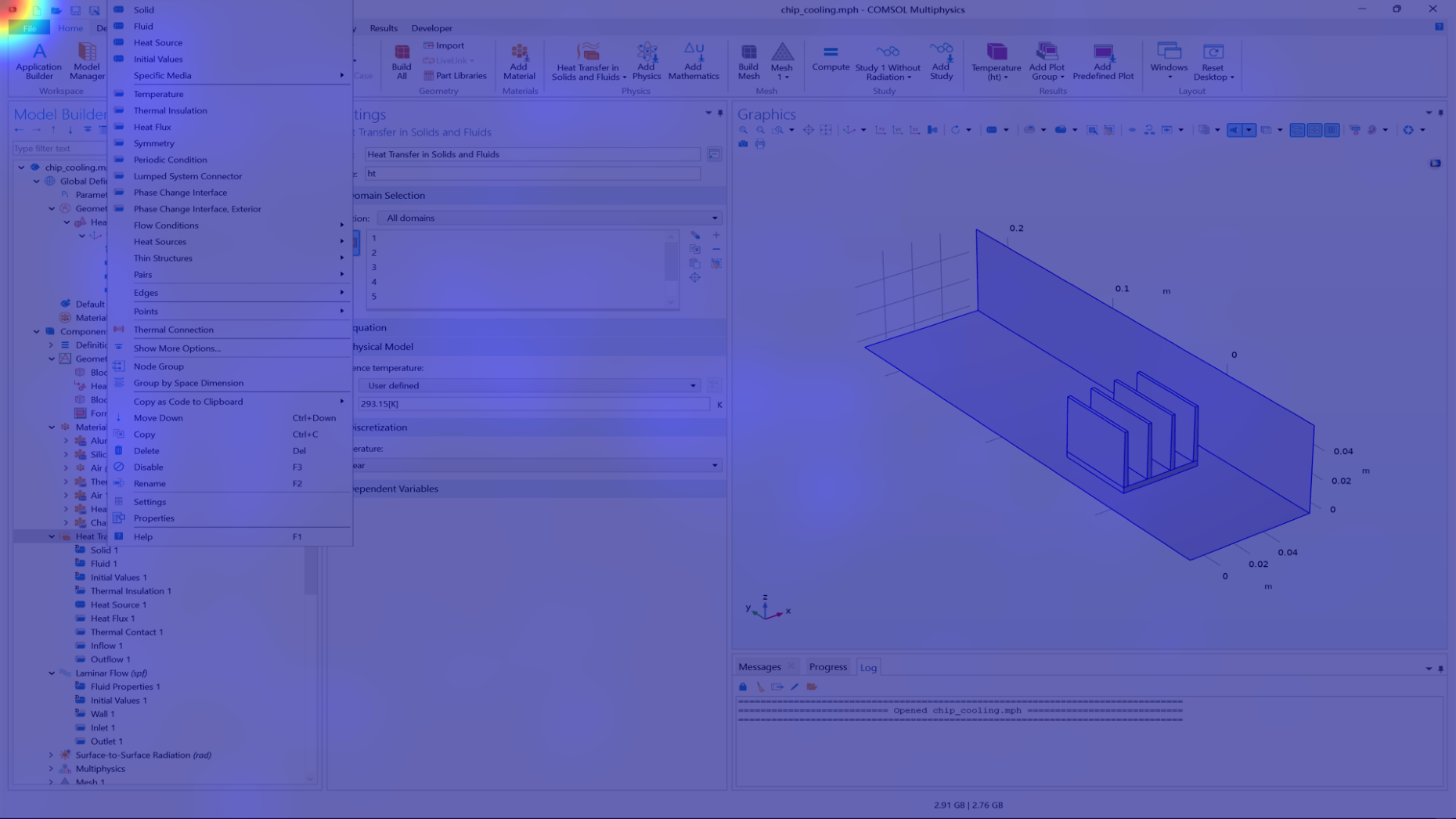}
\centering
\end{minipage}%
}%
\subfigure[OSAtlasPro]{
\begin{minipage}[t]{0.24\linewidth}
\includegraphics[width=\linewidth]{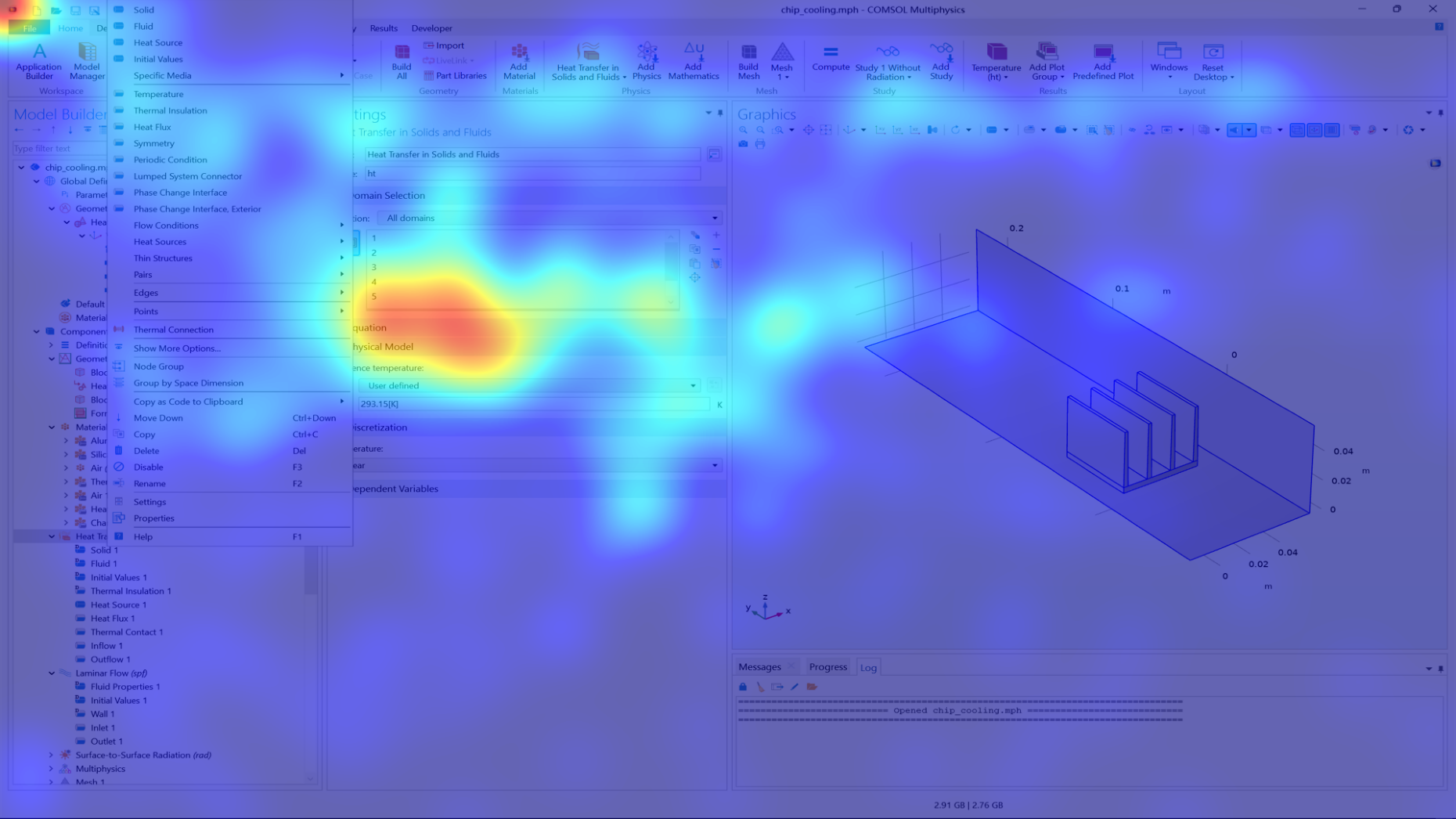}
\centering
\end{minipage}%
}%
\subfigure[UITARS]{
\begin{minipage}[t]{0.24\linewidth}
\includegraphics[width=\linewidth]{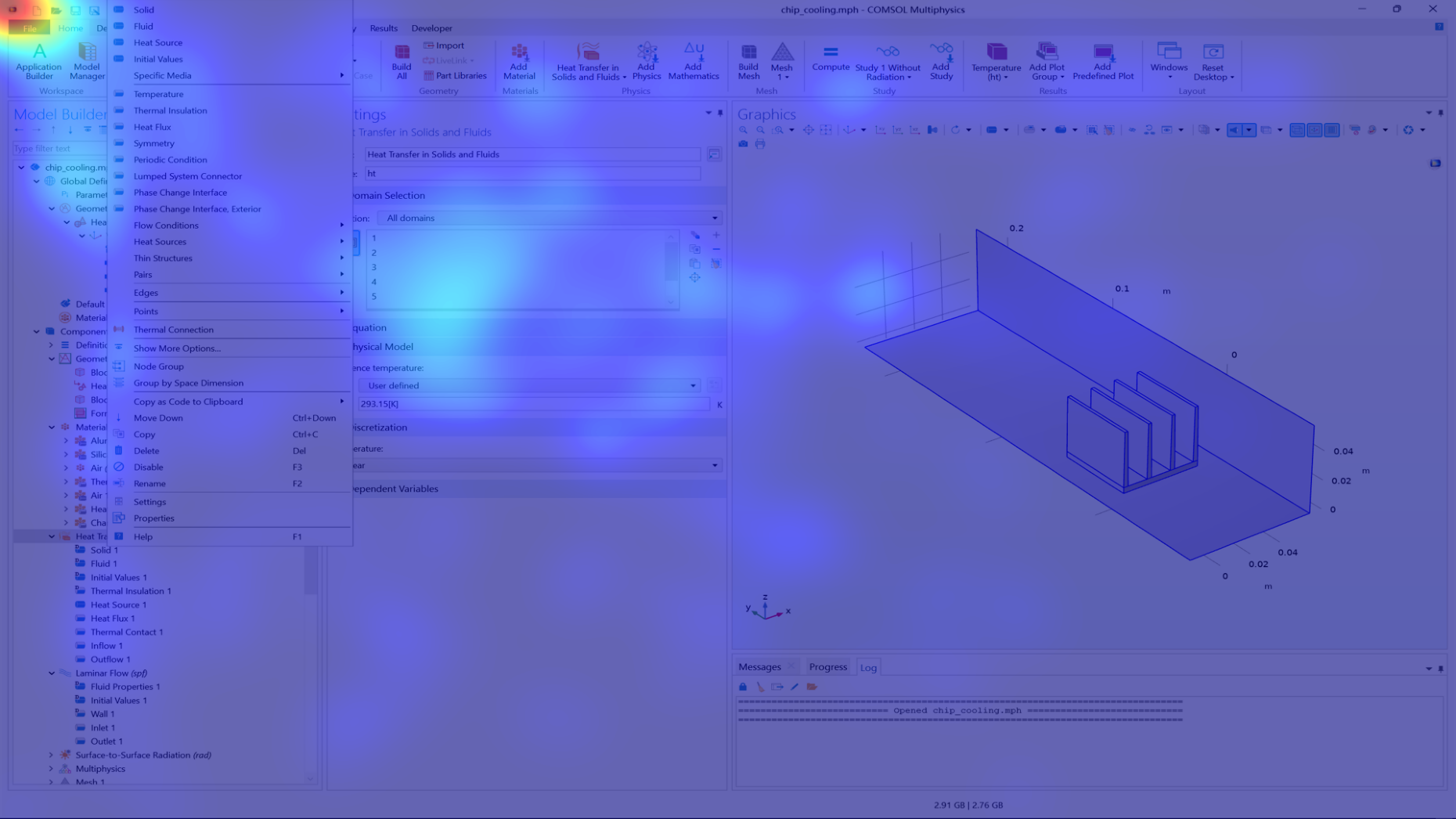}
\centering
\end{minipage}%
}%
\subfigure[EDAgent (proposed)]{
\begin{minipage}[t]{0.24\linewidth}
\includegraphics[width=\linewidth]{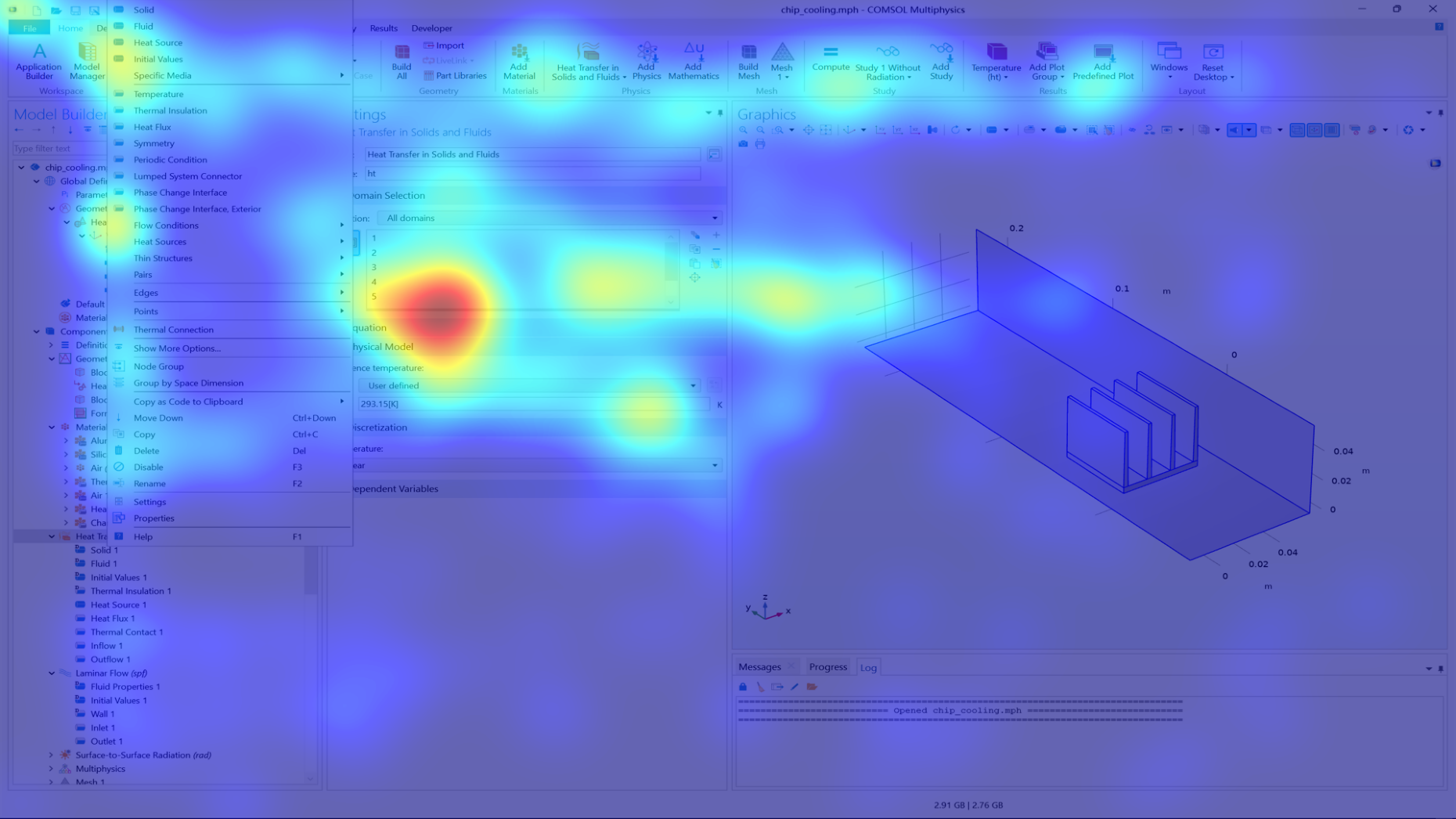}
\centering
\end{minipage}%
}%
\vspace{-2mm}
\caption{ROI distribution of GUI-EDA click locations, including the engineer-operated GT, six advanced GUI agents, and the proposed EDAgent. The click locations of existing GUI Agents are either too scattered or too concentrated in the upper left corner. Only the EDAgent and GT have the most consistent distribution. (Zoom in for detail)}
\label{fig:Pref}
\vspace{-2mm}
\end{figure*}

For qualitative cases, Figure \ref{fig:Pref} compares the spatial distribution of clicks in GUI-EDA dataset across six existing GUI Agents and our EDAgent. GT annotations produced by CAD engineers form a single, compact mode slightly below the upper-left menu bar at (a), indicating that \textbf{effective workflows center on the menu-bar but still require occasional excursions to the canvas and property pane}, where existing GUI Agents exhibit two systematic distortions.
\begin{itemize}
    \item Hyper-concentration: OSAtlas (e) and UITARS (g) collapse almost all probability mass into the same 6\%-region of the interface, producing a Dirac-like spike that misses downstream clicks on ports, boundary-condition panels and the graphics viewport.
    \item Over-dispersion: Aria-UI (c) and OSAtlasPro (f) spread density across the full $3840\times2160$ canvas, yielding frequent mis-clicks on irrelevant tool icons.
\end{itemize}
EDAgent (h) recovers the GT profile with the lowest KL-divergence. Its MLLM comprehension head explicitly models `menu-bar', `canvas' and `dialog' components, so the predicted heat-map presents a primary mode coincident with GT and aligned to the property for menu-bar, exactly the balance required for reproducible, engineer-like interaction sequences.

\begin{figure*}[t]
\centering

\subfigure[CogAgent]{
\begin{minipage}[t]{0.24\linewidth}
\includegraphics[width=\linewidth]{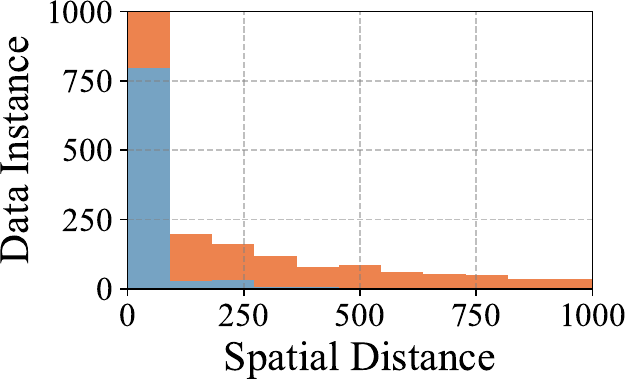}
\includegraphics[width=\linewidth]{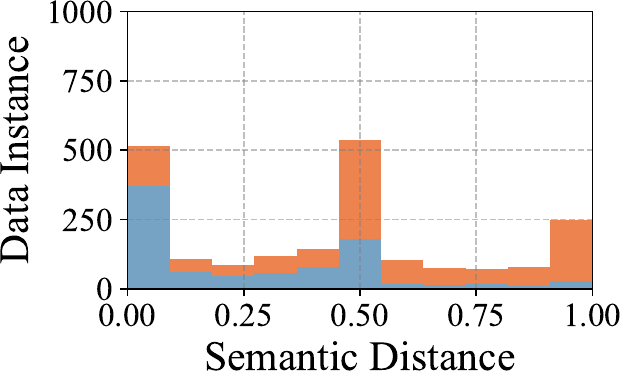}
\vspace{-2mm}
\centering
\end{minipage}%
}%
\subfigure[OSAtlasPro]{
\begin{minipage}[t]{0.24\linewidth}
\includegraphics[width=\linewidth]{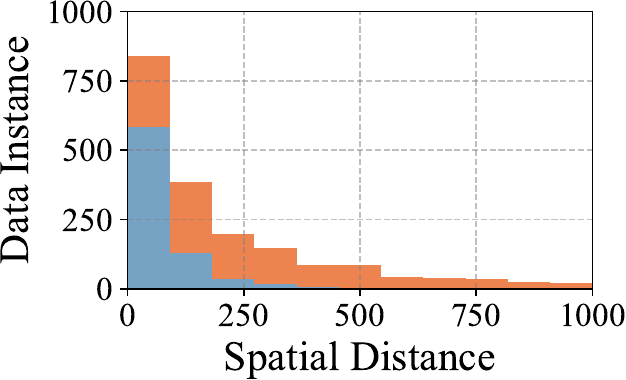}
\includegraphics[width=\linewidth]{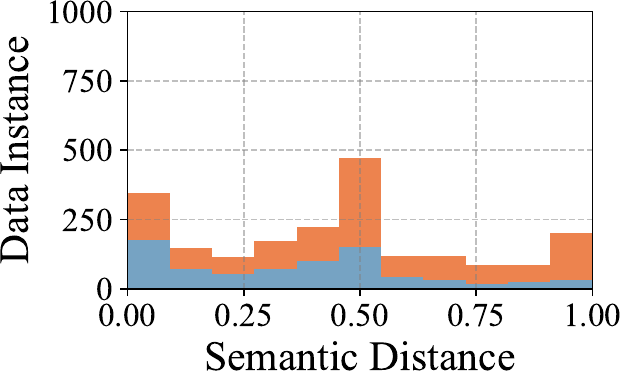}
\vspace{-2mm}
\centering
\end{minipage}%
}%
\subfigure[InternVL2.5]{
\begin{minipage}[t]{0.24\linewidth}
\includegraphics[width=\linewidth]{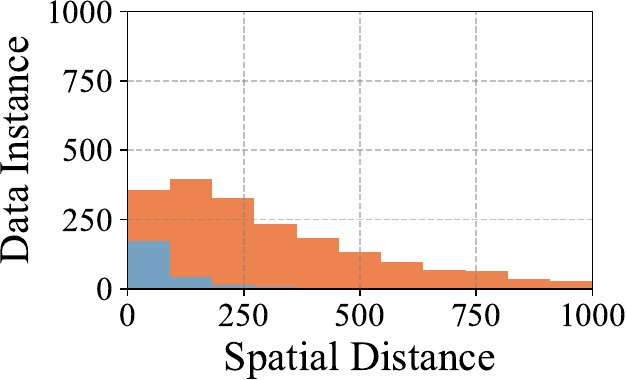}
\includegraphics[width=\linewidth]{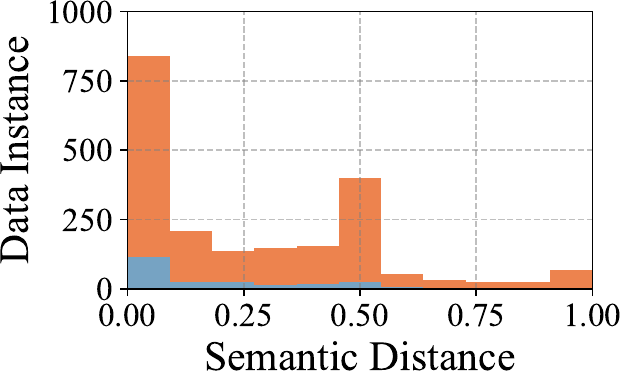}
\vspace{-2mm}
\centering
\end{minipage}%
}%
\subfigure[EDAgent (proposed)]{
\begin{minipage}[t]{0.24\linewidth}
\includegraphics[width=\linewidth]{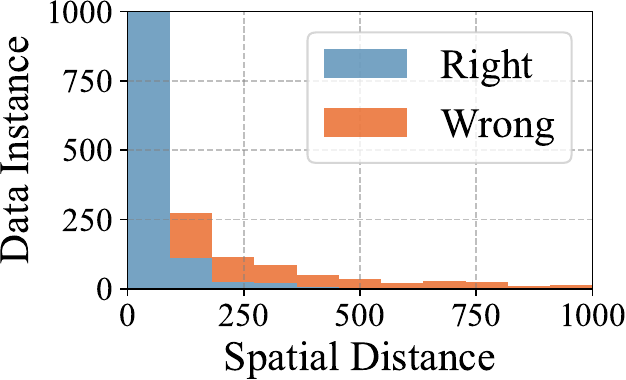}
\includegraphics[width=\linewidth]{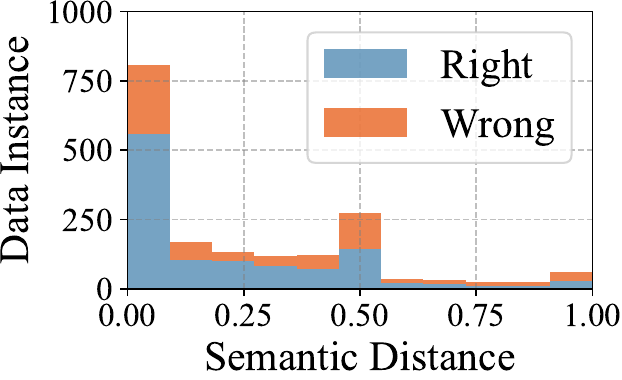}
\vspace{-2mm}
\centering
\end{minipage}%
}%
\vspace{-2mm}

\caption{The distribution of Right and Wrong examples in GUI-EDA, as inferred by four representative models. AGUI Agent (a, b) has many Wrong examples, but their spatial distance from GT location is within 100 pixels, indicating that a slight movement can correct them; MLLM (c) also shows many Wrong samples, but their semantic distance from GT description is less than 0.1, suggesting they are already understood by the model. Our EDAgent (d) perfectly corrects both types of errors.}
\label{fig:distance}
\vspace{-2mm}
\end{figure*}

For quantitative study, Figure \ref{fig:distance} disentangles the Right and Wrong Action sample of four representative policies, along two orthogonal axes: spatial displacement from the engineer click, and semantic divergence from the engineer description intent. The spatial distance comes directly from the pixel difference, while the semantic distance comes from the Answer Score in Section \ref{sec:exp-settings}, where:
\begin{itemize}
    \item GUI Agents (a, b) concentrate their wrong predictions within a 100-pixel radius of the ground-truth coordinate, producing a dominant first bin in the spatial histogram. This implies that the policy has recovered the approximate widget neighbourhood but lacks the sub-pixel refinement to deliver the final successful click.
    \item MLLM (c) exhibits the converse pathology. its spatial spread is wide, yet the semantic distance distribution is sharply peaked below 0.1, indicating that the underlying intent has been correctly parsed. Failures therefore stem from an inability to map an accurately understood goal onto an exact screen coordinate.
\end{itemize}
Therefore, we believe existing models are not fundamentally incapable for EDA tasks; rather, they consistently land in the immediate vicinity of the correct solution—median spatial offset$<$100 px, semantic distance$<$0.1—indicating that the requisite knowledge and perception are already present. \textbf{What is missing is not extra EDA training data, but an integrative mechanism that couples semantic comprehension with sub-pixel motor execution} (`cerebrum' and `cerebellum' function). 
Here, although the histogram of the proposed EDAgent (d) has no significant difference in spatial distance from (a) and semantic distance from (c), it eliminates the `understood-but-misclicked' and `clicked-nearly-but-missed' regimes simultaneously.
Therefore, rather than endowing models with novel capabilities, CAD software operation can be achieved by fully exploiting the existing capacities of MLLMs and GUI Agents. This principle will serve as the guiding doctrine for future endeavors in employing GUI Agents for EDA tasks.

\begin{table*}[t]
\centering
    \renewcommand\arraystretch{1.25}
    \renewcommand\tabcolsep{4pt}
    \belowrulesep=0pt\aboverulesep=0pt
    \caption{Ablation study for three core modules in EDAgent, namely MLLM for comprehension, GUI Agent for execution, and MLLM as a validator. EDAgent has the most suitable comprehension and execution engine for EDA tasks, and the Valid mechanism further improves its performance. [Keys: \CLB{Best}; \CLA{Second Best}]}
    \label{tab:ablation}
    \vspace{-5pt}
    \resizebox{\linewidth}{!}{
    \begin{tabular}{l|l|c|rr:rr:rr:rr:rr:rr:rr:rr}
    \toprule
\multicolumn{1}{c|}{\multirow{3}{*}{MLLM}} & \multicolumn{1}{c|}{\multirow{3}{*}{GUI Agent}} & \multicolumn{1}{c|}{\multirow{3}{*}{Valid}} & \multicolumn{8}{c:}{COMSOL}               & \multicolumn{2}{c:}{Flotherm}     & \multicolumn{2}{c:}{ICEPAK}       & \multicolumn{2}{c:}{CST}          & \multicolumn{2}{c}{HFSS}         \\ \cdashline{4-19}
\multicolumn{1}{c|}{}  & \multicolumn{1}{c|}{}        & \multicolumn{1}{c|}{}      & \multicolumn{2}{c:}{Acoustic}     & \multicolumn{2}{c:}{Optical}      & \multicolumn{2}{c:}{Mechanical}   & \multicolumn{6}{c:}{Thermal}            & \multicolumn{4}{c}{Magnetical}            \\ \cdashline{4-19}
\multicolumn{1}{c|}{}  & \multicolumn{1}{c|}{}        & \multicolumn{1}{c|}{}      & \multicolumn{1}{c}{Ori.} & \multicolumn{1}{c:}{Dyn.} & \multicolumn{1}{c}{Ori.} & \multicolumn{1}{c:}{Dyn.} & \multicolumn{1}{c}{Ori.} & \multicolumn{1}{c:}{Dyn.} & \multicolumn{1}{c}{Ori.} & \multicolumn{1}{c:}{Dyn.} & \multicolumn{1}{c}{Ori.} & \multicolumn{1}{c:}{Dyn.} & \multicolumn{1}{c}{Ori.} & \multicolumn{1}{c:}{Dyn.} & \multicolumn{1}{c}{Ori.} & \multicolumn{1}{c:}{Dyn.} & \multicolumn{1}{c}{Ori.} & \multicolumn{1}{c}{Dyn.} \\ \midrule

Qwen2.5VL              & Aguvis   & {\faCheckSquare}     & \CLB{0.66}  & \CLA{0.71}  & \CLB{0.72}  & \CLB{0.75}  & \CLB{0.74}  & \CLA{0.73}  & \CLA{0.47}  & \CLB{0.62}  & \CLB{0.37}  & \CLB{0.46}  & 0.30  & \CLA{0.45}  & \CLB{0.73}  & \CLB{0.76}  & 0.51  & 0.58  \\
Qwen2.5VL              & Aguvis   &        & 0.64  & 0.71  & \CLA{0.62}  & 0.69  & 0.66  & 0.71  & \CLB{0.48}  & \CLA{0.61}  & 0.28  & 0.44  & \CLB{0.33}  & 0.41  & \CLA{0.71}  & 0.73  & 0.47  & 0.53  \\
Qwen2.5VL              &          &        & 0.06  & 0.19  & 0.04  & 0.20  & 0.06  & 0.21  & 0.08  & 0.18  & 0.07  & 0.18  & 0.00  & 0.19  & 0.16  & 0.25  & 0.05  & 0.17  \\
   & Aguvis   &        & 0.40  & 0.53  & 0.56  & 0.58  & 0.48  & 0.51  & 0.24  & 0.36  & 0.17  & 0.26  & 0.10  & 0.27  & 0.49  & 0.59  & 0.33  & 0.44  \\ \cdashline{1-19}
InternVL2.5            & Aguvis   & {\faCheckSquare}     & 0.60  & 0.68  & 0.60  & \CLA{0.73}  & 0.66  & \CLB{0.74}  & 0.42  & 0.58  & 0.26  & 0.41  & 0.28  & 0.42  & 0.70  & \CLA{0.76}  & 0.46  & 0.56  \\
InternVL2.5            & Aguvis   &        & 0.48  & 0.57  & 0.48  & 0.63  & 0.60  & 0.68  & 0.42  & 0.53  & 0.24  & 0.39  & 0.20  & 0.36  & 0.65  & 0.71  & 0.41  & 0.50  \\
Ovis2                 & Aguvis   & {\faCheckSquare}     & \CLA{0.64}  & \CLB{0.73}  & 0.60  & 0.70  & \CLA{0.70}  & 0.72  & 0.39  & 0.57  & 0.22  & 0.35  & 0.20  & 0.40  & 0.64  & 0.73  & \CLB{0.60}  & \CLB{0.61}  \\
Ovis2                 & Aguvis   &        & 0.52  & 0.65  & 0.46  & 0.61  & 0.60  & 0.63  & 0.36  & 0.51  & 0.18  & 0.37  & 0.18  & 0.32  & 0.61  & 0.67  & \CLA{0.56}  & \CLA{0.59}  \\
Qwen2.5VL              & CogAgent    & {\faCheckSquare}     & 0.52  & 0.69  & 0.42  & 0.63  & 0.72  & 0.77  & 0.42  & 0.60  & 0.32  & \CLA{0.45}  & \CLA{0.32}  & \CLB{0.48}  & 0.67  & 0.72  & 0.31  & 0.42  \\
Qwen2.5VL              & CogAgent    &        & 0.56  & 0.69  & 0.44  & 0.62  & 0.68  & 0.73  & 0.43  & 0.59  & 0.22  & 0.42  & 0.24  & 0.39  & 0.67  & 0.70  & 0.26  & 0.37  \\
Qwen2.5VL              & OSAtlasPro      & {\faCheckSquare}     & 0.56  & 0.68  & 0.48  & 0.64  & 0.64  & 0.71  & 0.45  & 0.56  & \CLA{0.33}  & 0.36  & 0.18  & 0.39  & 0.54  & 0.57  & 0.31  & 0.36  \\
Qwen2.5VL              & OSAtlasPro      &        & 0.58  & 0.68  & 0.44  & 0.56  & 0.58  & 0.65  & 0.38  & 0.51  & 0.31  & 0.32  & 0.18  & 0.38  & 0.53  & 0.54  & 0.26  & 0.39 \\ \bottomrule   
\end{tabular}
    }
    \vspace{-2.5mm}
\end{table*}

\subsection{Ablation Study}

Table \ref{tab:ablation} presents an ablation study that quantifies the individual contribution of the three core modules embedded in EDAgent: (i) an MLLM-based comprehension engine, (ii) a GUI Agent-based execution engine, and (iii) an MLLM-based validator that refines the final action sequence. Each row reports the Action Score achieved by a specific MLLM \& GUI Agent combination, under both Original (Ori.) and Dynamic (Dyn.) Resolutions, enabling a controlled assessment of module necessity and interchangeability.
Overall, compared to pipelines with disabled or replaced modules, \textbf{the integrated EDAgent model achieves best/second-best in 13/16 dimensions, demonstrating the contribution of each module.}

The upper-block experiments (rows above the horizontal rule) substantiate the indispensability of each EDAgent constituent. The fully-integrated configuration, Qwen2.5-VL for comprehension, Aguvis for execution, and the MLLM validator enabled—delivers the highest Action Score. Systematically ablating the validator while freezing the remaining pipeline reduces the performance of 14/16 dimensions, demonstrating that the validation stage eliminates low-confidence actions would otherwise register as false clicks. Eliminating the GUI Agent and allowing the MLLM to emit raw coordinates collapses performance to all dimensions, corroborating that high-level semantic understanding alone is insufficient for pixel-accurate manipulation. Conversely, retaining only the GUI agent (Aguvis-7B without MLLM guidance) yields 0.44, whose performance witness catastrophic decline in complex Fl-Thermal and IC-Thermal interface. Each module therefore contributes a statistically significant marginal gain, and their sequential arrangement produces a super-additive effect that no single component can replicate.

The lower-block ablations generalize these findings across alternative comprehension–execution couples. When InternVL2.5 is substituted for Qwen2.5-VL, activating the validator elevates the score from 0.37 to 0.42; an analogous uplift (+0.02 $\sim$ +0.06) is observed for Ovis2, CogAgent, and OSAtlasPro pairings, indicating that the benefit of validator is architecture-agnostic.
Specifically, for InternVL2.5+Aguvis, the validator significantly improves the CO-Acoustic and CO-Mechanical dimensions, which already have high baselines, with a gain of nearly 0.1. For Qwen2.5VL+CogAgent, the validator excels at optimizing the difficult Fl-Thermal and IC-Thermal tasks. In short, applying MLLM as validator has a significant positive effect on any MLLM+GUI Agent combination.
Among all evaluated combinations, the Qwen2.5-VL+Aguvis dyad consistently produces the highest absolute scores both with and without validation, justifying its selection as the default backbone in EDAgent. Collectively, the ablations verify that (i) the validator enhances every MLLM–GUI-agent dyad, and (ii) the chosen comprehension–execution pair realises the upper-bound of empirical performance, corroborating the optimality of the integrated EDAgent design.

\subsection{Silicon Tape-out Validation}

\begin{figure}[tb]
\centering
\subfigure[LRA Sample (Success) Platform: HFSS]{
\begin{minipage}[t]{\linewidth}
\includegraphics[width=0.62\linewidth,height=0.34\linewidth,keepaspectratio=false]{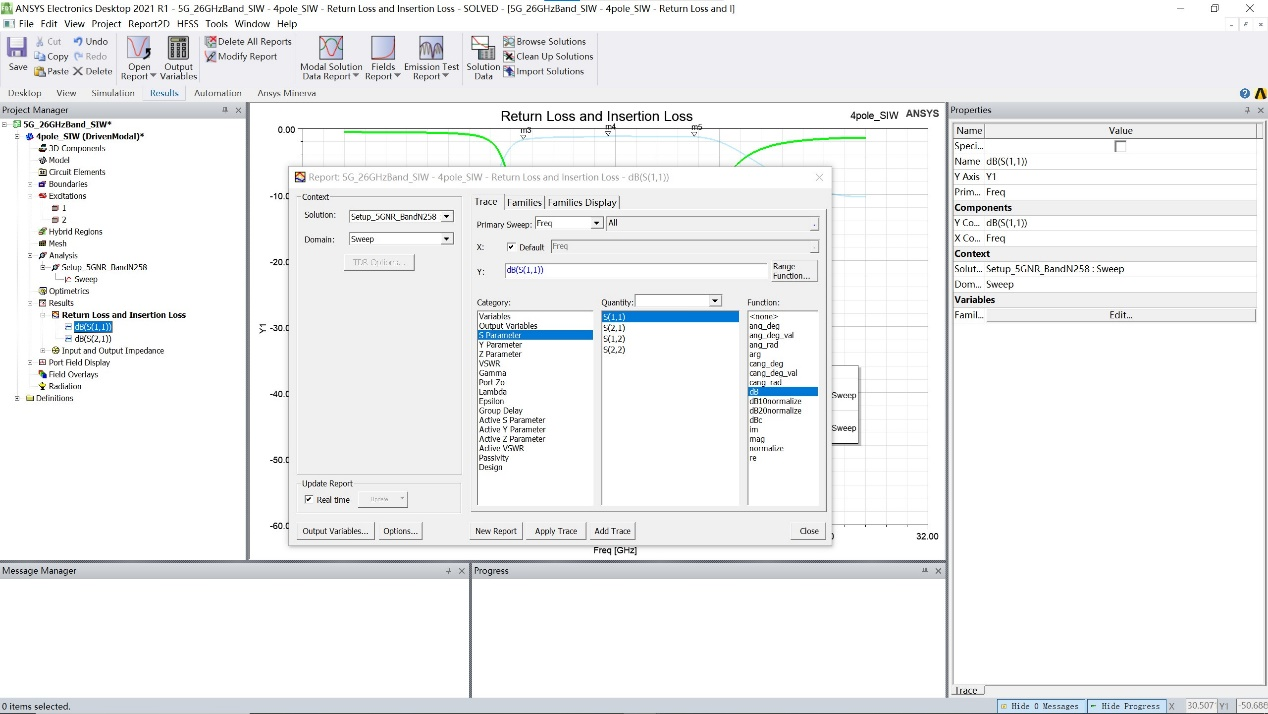}
\includegraphics[width=0.34\linewidth]{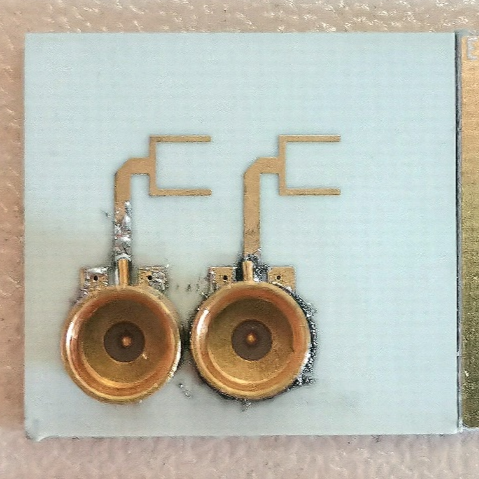}
\centering
\end{minipage}%
}%
\vspace{-2mm}
\subfigure[LTCC Sample (Success) Platform: HFSS]{
\begin{minipage}[t]{\linewidth}
\includegraphics[width=0.62\linewidth,height=0.34\linewidth,keepaspectratio=false]{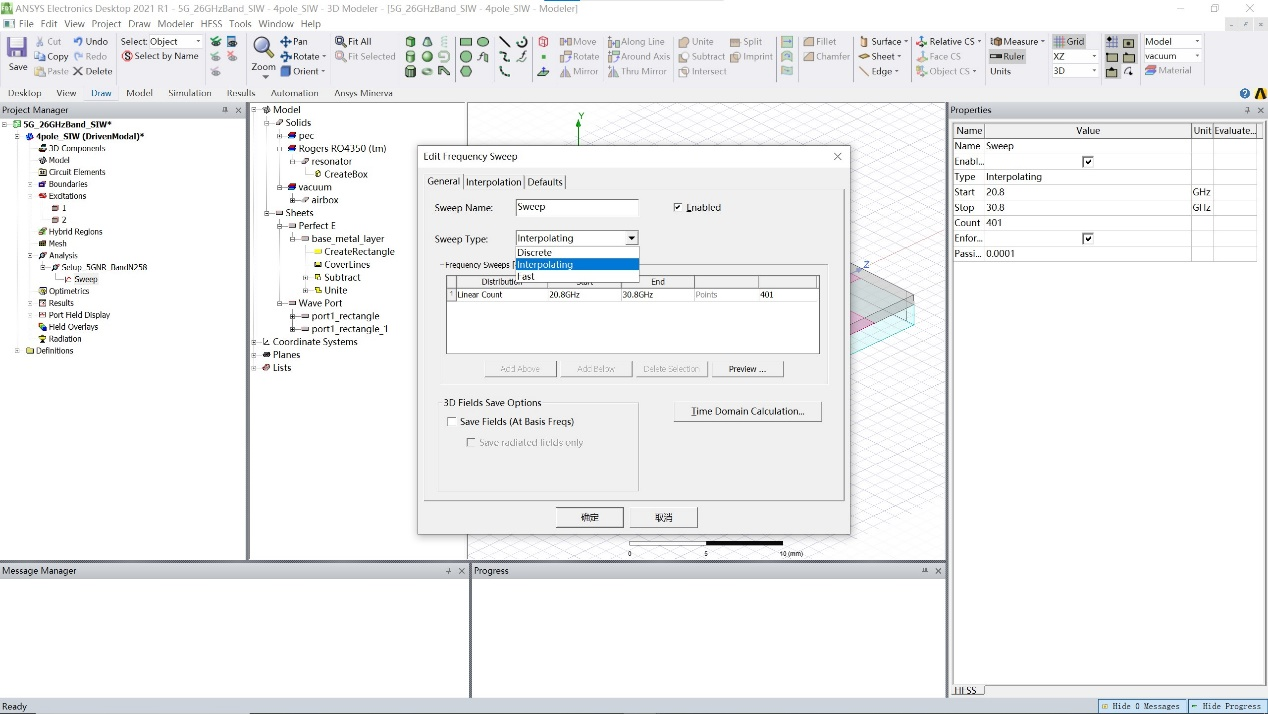}
\includegraphics[width=0.34\linewidth]{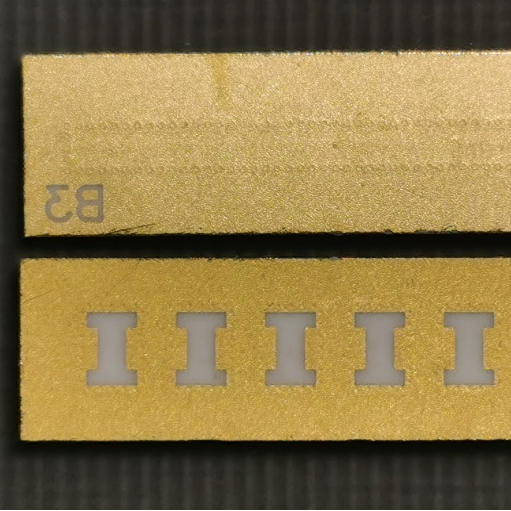}
\centering
\end{minipage}%
}%
\vspace{-2mm}
\subfigure[SIW Sample (Failed) Platform: CST]{
\begin{minipage}[t]{\linewidth}
\includegraphics[width=0.62\linewidth,height=0.34\linewidth,keepaspectratio=false]{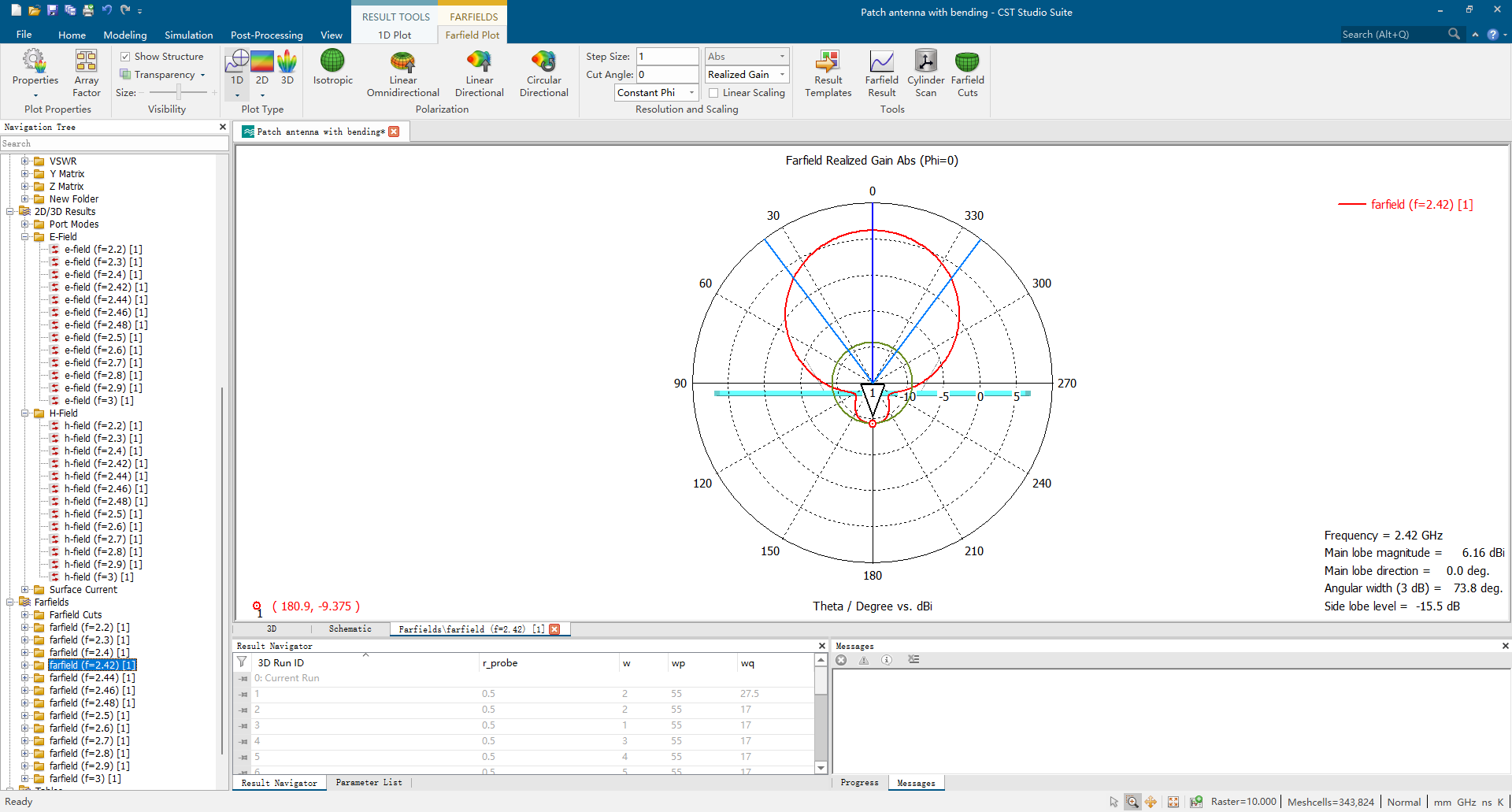}
\includegraphics[width=0.34\linewidth]{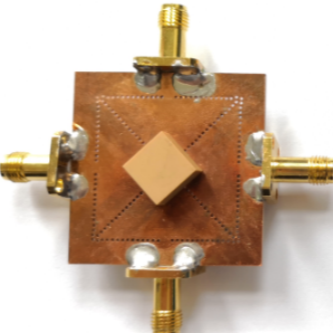}
\centering
\end{minipage}%
}%
\vspace{-2mm}
\subfigure[PA Sample (Failed) Platform: COMSOL]{
\begin{minipage}[t]{\linewidth}
\includegraphics[width=0.62\linewidth,height=0.34\linewidth,keepaspectratio=false]{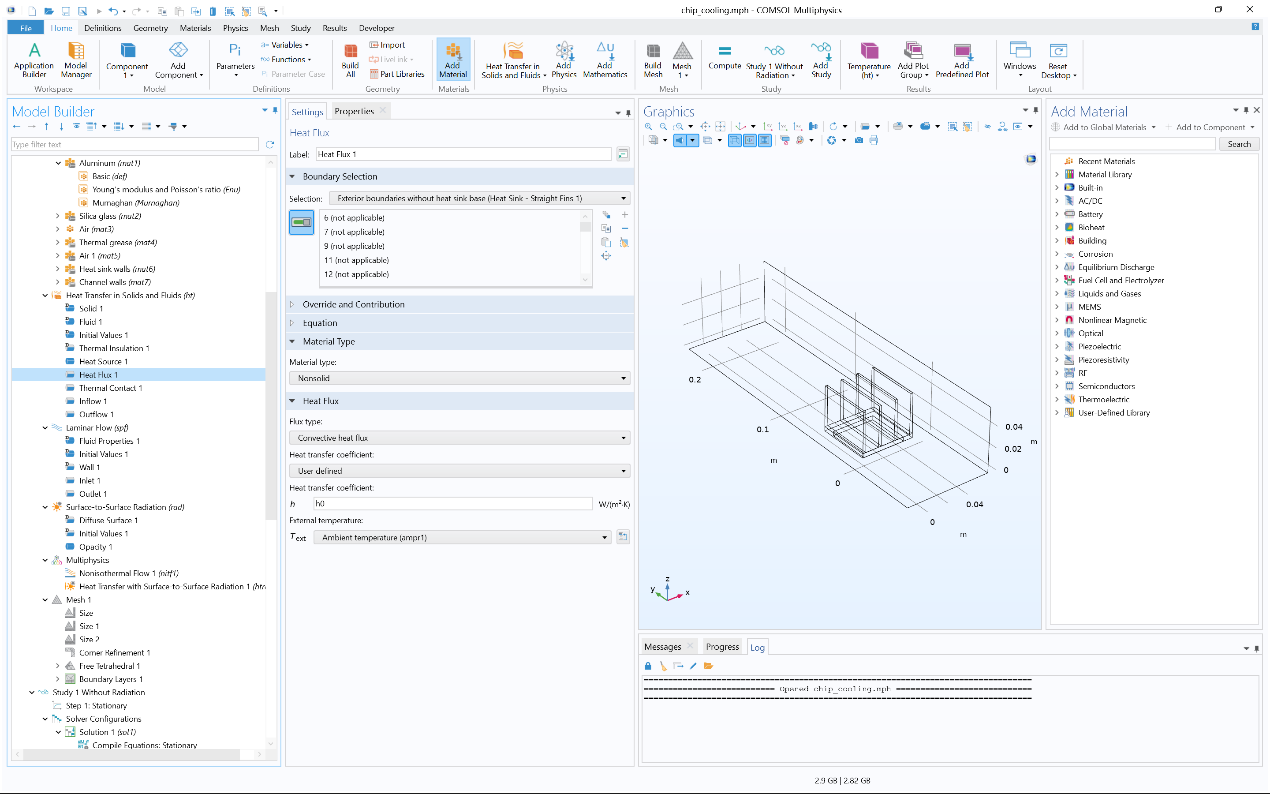}
\includegraphics[width=0.34\linewidth]{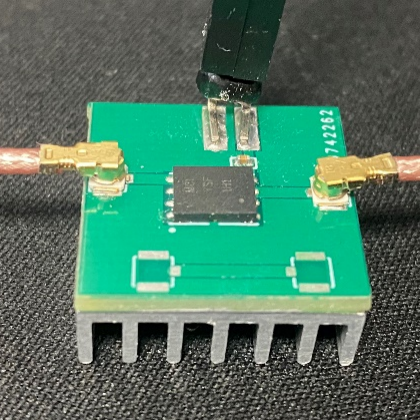}
\centering
\end{minipage}%
}%
\vspace{-2mm}
\caption{Real-world silicon validation based on the instruction in CAD software. Correct execution in the virtual environment can ensure successful completion of EDA tasks during Real-world manufacturing, and vice versa. (Zoom in for detail)}
\vspace{-3mm}
\label{fig:hardware}
\end{figure}

To rigorously assess the physical fidelity of GUI-driven EDA flows, we emulated the complete silicon manufacturing pipeline and identified a single Key Step whose pixel-level configuration critically determines die yield, as illustrated in Figure \ref{fig:interface}. This step is injected into the virtual CAD environment as an image prompt; the GUI Agent-generated Action is executed, while all preceding and subsequent steps are held at their golden values operated by engineers. 
Considering the five Fields and five Software in the proposed GUI-EDA dataset, we construct seven independent tape-out iterations to reveal perfect parity between virtual correctness and silicon functionality: every software-successful run produced a defect-free wafer, and every software mis-click was reproduced as an observable failure in the fabricated device.
Figure \ref{fig:hardware} exemplifies two positive and two negative cases to prove such simulation-to-real correspondence. 

\noindent
(a) Layered Resonator Antenna (LRA) Case
\begin{itemize}
    \item Scientist demand: A broadband solution for zero-gap LRA arrays—cut mutual coupling without extra structures and keep overlapping bandwidth $\leq5$ GHz.
    \item GUI Agent operation: In the Key-Step panel, the agent enlarged HFSS S-parameter monitors and reduced the $\Delta$ mismatch between stacked substrates from 0.8 to 0.15.
    \item Silicon result: $S_{21} < - 25$ dB over 25.8$\sim$30.3 GHz gives 4.5 GHz overlap, meeting spec. The fabricated die exhibits a symmetric radiation pattern and a return-loss better than $-10$ dB across the same interval, demonstrating the GUI Agent Action satisfy electromagnetic constraints.
\end{itemize}
(b) Low Temperature Co-fired Ceramics (LTCC) Case
\begin{itemize}
    \item Scientist demand: Miniaturize an LRA-in-LTCC cell so that both lateral footprint and thickness are drastically shrunk while retaining wide impedance bandwidth and high radiation efficiency at 60 GHz.
    \item GUI Agent operation: The Agent reduce the ceramic stack from eight to five layers in HFSS `Fast' solve option, thus optimize the feed-line length for 56–62 GHz coverage.
    \item Silicon result: Measured peak gain 10.6 dB (-0.7 dB vs. 11.32 dB), radiation efficiency 63\% at 60 GHz; 3 dB gain bandwidth 5.3 GHz. The GUI Agent-guided geometry satisfies miniaturization and bandwidth specs.
\end{itemize}
(c) Substrate Integrated Waveguide (SIW) Case
\begin{itemize}
    \item Scientist demand: Realize a dual-polarised filter-antenna fed by an isosceles-right-triangle SIW cavity, targeting $>$ 78 dB port isolation in X-/Y-pol within the pass-band.
    \item GUI Agent operation: In the Key Step panel the agent disabled the bottom slot-etch option in CST and set the cavity-side wall taper angle to shorten fabrication steps.
    \item Silicon result: Fabricated device exhibits port isolation of only 52 dB ($<$78 dB), cross-pol suppression 28 dB, and a 3 dB beam-width asymmetry $>10^\circ$ between X and Y-pol; missing slot-etch causes mode contamination and clear spec failure.
\end{itemize}
(d) Power Amplifier (PA) Case
\begin{itemize}
\item Scientist demand: Maintain PA gain variation $\leq$1 dB over 50 °C temperature sweep by properly setting the heatsink convection coefficient.
\item GUI Agent operation: The agent lock the convection coefficient at its room-temperature default (5 $\rm W/m^{2}$) for the Key Step and instead raised the ambient temperature node that mistakenly treating the package as isothermal.
\item Silicon result: Measured gain drops 2.3 dB at 50 °C gradient (spec $\leq$1 dB); thermal images reveal a 22 °C junction-hotspot, confirming inadequate heat removal and from the Agent leads to out-of-spec performance.
\end{itemize}
These results substantiate that \textbf{our Action Score in the virtual CAD environment is strongly correlated to silicon-level success} as a sufficient predictor, validating the use of dynamic GUI benchmark as a surrogate for costly physical tape-outs.

\begin{figure}[t]
    \centering
    \includegraphics[width=\linewidth]{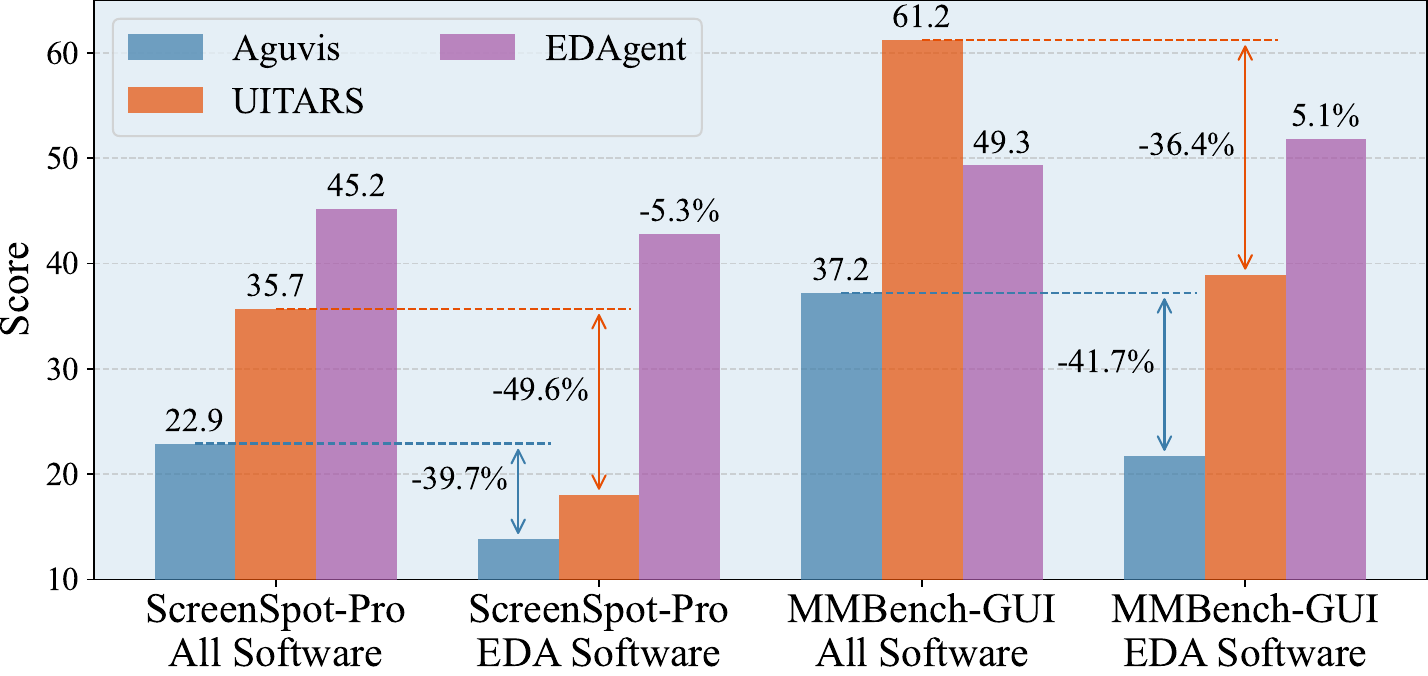}
    \caption{Cross-dataset validation for EDAgent and the other two advanced GUI Agents. EDAgent still performs well in Screenpot-Pro \cite{relate:screenspotpro} and MMBench-GUI \cite{relate:mmbenchgui}, especially shows no denegation in their CAD Software subsets.}
    \label{fig:cross}
    \vspace{-5mm}
\end{figure}

\subsection{Cross Dataset Validation}

Beyond the dataset we proposed, cross evaluation on ScreenSpot-Pro \cite{relate:screenspotpro} and MMBench-GUI \cite{relate:mmbenchgui} in Figure \ref{fig:cross} corroborates the transferability of EDAgent beyond the proprietary GUI-EDA suite. While the two comparative agents suffer pronounced performance degradation on CAD-centric subsets—Aguvis drops 5.3\% and UITARS declines 36.4\% relative to their full-set scores—EDAgent exhibits exceptional robustness: its ScreenSpot-Pro CAD score remains flat (+0.2\%) and its MMBench-GUI CAD result even improves by 1.9\%, yielding absolute gains of 6.1\% and 7.5\% over the strongest competitor, respectively. 
This divergence indicates the `cerebrum'+`cerebellum' synergy can improve the success rate of CAD software on various datasets, and the Validator ensures the `cerebrum' does not overthink, thereby avoiding the performance degradation of other software.
Consequently, EDAgent not only attains highest accuracy on in-domain EDA tasks but also preserves superiority on out-domain benchmarks, affirming its utility as a universal GUI Agent.

\section{Conclusion}

Motivated by the prohibitive cost of expert-level EDA labor, this study reframes professional EDA interaction as a high-impact automation target. We curated GUI-EDA, a large-scale, multi-resolution dataset spanning five fields and five CAD tools, engineer-annotated and silicon-validated. Systematic benchmarking of 26 advanced MLLMs and GUI Agents revealed a persistent performance ceiling, confirming that pixel-perfect, semantics-aware execution remains an open challenge.  

To bridge the gap, we propose EDAgent, fusing MLLM comprehension with GUI Agent execution under self-reflective validation. Action Score rises from 0.46 to 0.598, surpassing human experts for the first time while remaining robust across Resolutions, Difficulties, and out-of-domain benchmarks. Silicon runs show perfect cyber-physical parity: every virtual success yields a defect-free component, every mis-click a measurable failure. This closed-loop validation authenticates GUI-EDA as a low-cost evaluation surrogate for costly prototyping.

More broadly, our work extends GUI Agents from mundane office tasks to the economically vital arena of semiconductor design, offering a possible automation path that can save thousands of engineer-hours per product cycle. Ultimately, as Agents evolve from clicking spreadsheets to taping out chips, creativity, not repetitive pointing, will define the technological contribution of humanity.

\bibliographystyle{IEEEbib-jstsp}
\bibliography{cite}

@String(ICLR  = {Int. Conf. Learn. Represent.})

@String(ICLR  = {ICLR})

@String(ICLR = {Int. Conf. Learn. Represent.})

@article{relate:osworld,
  title={Osworld: Benchmarking multimodal agents for open-ended tasks in real computer environments},
  author={Xie, Tianbao and Zhang, Danyang and Chen, Jixuan and Li, Xiaochuan and Zhao, Siheng and Cao, Ruisheng and Hua, Toh J and Cheng, Zhoujun and Shin, Dongchan and Lei, Fangyu and others},
  journal={Advances in Neural Information Processing Systems},
  volume={37},
  pages={52040--52094},
  year={2024}
}

@misc{relate:mmbenchgui,
      title={MMBench-GUI: Hierarchical Multi-Platform Evaluation Framework for GUI Agents}, 
      author={Xuehui Wang and Zhenyu Wu and JingJing Xie and Zichen Ding and Bowen Yang and Zehao Li and Zhaoyang Liu and Qingyun Li and Xuan Dong and Zhe Chen and others},
      year={2025},
      eprint={2507.19478},
      archivePrefix={arXiv},
      primaryClass={cs.CV},
      url={https://arxiv.org/abs/2507.19478}, 
}

@misc{relate:screenspotpro,
      title={ScreenSpot-Pro: GUI Grounding for Professional High-Resolution Computer Use}, 
      author={Kaixin Li and Ziyang Meng and Hongzhan Lin and Ziyang Luo and Yuchen Tian and Jing Ma and Zhiyong Huang and Tat-Seng Chua},
      year={2025},
      eprint={2504.07981},
      archivePrefix={arXiv},
      primaryClass={cs.CV},
      url={https://arxiv.org/abs/2504.07981}, 
}

@article{relate:mind2web,
  title={Mind2web: Towards a generalist agent for the web},
  author={Deng, Xiang and Gu, Yu and Zheng, Boyuan and Chen, Shijie and Stevens, Sam and Wang, Boshi and Sun, Huan and Su, Yu},
  journal={Advances in Neural Information Processing Systems},
  volume={36},
  pages={28091--28114},
  year={2023}
}

@misc{relate:mind2weblive,
  title        = {Mind2Web-Live: Interactive Web Agent Evaluation at Scale},
  author       = {Christopher Rawles and Sarah Clinckemaillie and Yifan Chang and Jonathan Waltz and Gabrielle Lau and Marybeth Fair and Alice Li and William Bishop and Wei Li and Folawiyo Campbell-Ajala and Oriana Riva},
  eprint       = {arXiv preprint arXiv:2405.14573},
  year         = {2024}
}

@inproceedings{relate:Mind2Web-Multimodal,
  title={GPT-4V (ision) is a Generalist Web Agent, if Grounded},
  author={Zheng, Boyuan and Gou, Boyu and Kil, Jihyung and Sun, Huan and Su, Yu},
  booktitle={International Conference on Machine Learning},
  pages={61349--61385},
  year={2024},
  organization={PMLR}
}

@inproceedings{relate:Explorer-Web,
    title = "Explorer: Scaling Exploration-driven Web Trajectory Synthesis for Multimodal Web Agents",
    author = "Pahuja, Vardaan  and Lu, Yadong  and Rosset, Corby  and
      Gou, Boyu  and Mitra, Arindam and Whitehead, Spencer  and
      Su, Yu  and Awadallah, Ahmed Hassan",
    booktitle = "Findings of the Association for Computational Linguistics: ACL 2025",
    month = jul,
    year = "2025",
    address = "Vienna, Austria",   
    url = "https://aclanthology.org/2025.findings-acl.326/",
    doi = "10.18653/v1/2025.findings-acl.326",
    pages = "6300--6323",
    ISBN = "979-8-89176-256-5",
}

@misc{relate:Screenspot,
      title={SeeClick: Harnessing GUI Grounding for Advanced Visual GUI Agents}, 
      author={Kanzhi Cheng and Qiushi Sun and Yougang Chu and Fangzhi Xu and Yantao Li and Jianbing Zhang and Zhiyong Wu},
      year={2024},
      eprint={2401.10935},
      archivePrefix={arXiv},
      primaryClass={cs.HC}
}

@misc{relate:gui-reflection,
      title={GUI-Reflection: Empowering Multimodal GUI Models with Self-Reflection Behavior}, 
      author={Penghao Wu and Shengnan Ma and Bo Wang and Jiaheng Yu and Lewei Lu and Ziwei Liu},
      year={2025},
      eprint={2506.08012},
      archivePrefix={arXiv},
      primaryClass={cs.AI},
      url={https://arxiv.org/abs/2506.08012}, 
}

@misc{relate:tongui,
      title={TongUI: Building Generalized GUI Agents by Learning from Multimodal Web Tutorials}, 
      author={Bofei Zhang and Zirui Shang and Zhi Gao and Wang Zhang and Rui Xie and Xiaojian Ma and Tao Yuan and Xinxiao Wu and Song-Chun Zhu and Qing Li},
      year={2025},
      eprint={2504.12679},
      archivePrefix={arXiv},
      primaryClass={cs.CV},
      url={https://arxiv.org/abs/2504.12679}, 
}

@misc{relate:learnact,
      title={LearnAct: Few-Shot Mobile GUI Agent with a Unified Demonstration Benchmark}, 
      author={Guangyi Liu and Pengxiang Zhao and Liang Liu and Zhiming Chen and Yuxiang Chai and Shuai Ren and Hao Wang and Shibo He and Wenchao Meng},
      year={2025},
      eprint={2504.13805},
      archivePrefix={arXiv},
      primaryClass={cs.HC},
      url={https://arxiv.org/abs/2504.13805}, 
}

@misc{relate:gui-actor,
      title={GUI-Actor: Coordinate-Free Visual Grounding for GUI Agents}, 
      author={Qianhui Wu and Kanzhi Cheng and Rui Yang and Chaoyun Zhang and Jianwei Yang and Huiqiang Jiang and Jian Mu and Baolin Peng and Bo Qiao and Reuben Tan and others},
      year={2025},
      eprint={2506.03143},
      archivePrefix={arXiv},
      primaryClass={cs.CL},
      url={https://arxiv.org/abs/2506.03143}, 
}

@misc{relate:scienceboard,
      title={ScienceBoard: Evaluating Multimodal Autonomous Agents in Realistic Scientific Workflows}, 
      author={Qiushi Sun and Zhoumianze Liu and Chang Ma and Zichen Ding and Fangzhi Xu and Zhangyue Yin and Haiteng Zhao and Zhenyu Wu and Kanzhi Cheng and Zhaoyang Liu and others},
      year={2025},
      eprint={2505.19897},
      archivePrefix={arXiv},
      primaryClass={cs.AI},
      url={https://arxiv.org/abs/2505.19897}, 
}

@inproceedings{relate:guiworld,
    title={{GUI}-World: A Video Benchmark and Dataset for Multimodal {GUI}-oriented Understanding},
    author={Dongping Chen and Yue Huang and Siyuan Wu and Jingyu Tang and Huichi Zhou and Qihui Zhang and Zhigang He and Yilin Bai and Chujie Gao and Liuyi Chen and others},
    booktitle={The Thirteenth International Conference on Learning Representations},
    year={2025},
    url={https://openreview.net/forum?id=QarKTT5brZ}
}

@misc{relate:cerebrum,
      title={CODA: Coordinating the Cerebrum and Cerebellum for a Dual-Brain Computer Use Agent with Decoupled Reinforcement Learning}, 
      author={Zeyi Sun and Yuhang Cao and Jianze Liang and Qiushi Sun and Ziyu Liu and Zhixiong Zhang and Yuhang Zang and Xiaoyi Dong and Kai Chen and Dahua Lin and Jiaqi Wang},
      year={2025},
      eprint={2508.20096},
      archivePrefix={arXiv},
      primaryClass={cs.CV},
      url={https://arxiv.org/abs/2508.20096}, 
}

@misc{relate:computeruse,
      title={SEAgent: Self-Evolving Computer Use Agent with Autonomous Learning from Experience}, 
      author={Zeyi Sun and Ziyu Liu and Yuhang Zang and Yuhang Cao and Xiaoyi Dong and Tong Wu and Dahua Lin and Jiaqi Wang},
      year={2025},
      eprint={2508.04700},
      archivePrefix={arXiv},
      primaryClass={cs.AI},
      url={https://arxiv.org/abs/2508.04700}, 
}

@misc{relate:guir1,
      title={GUI-R1: A Generalist R1-Style Vision-Language Action Model For GUI Agents}, 
      author={Run Luo and Lu Wang and Wanwei He and Longze Chen and Jiaming Li and Xiaobo Xia},
      year={2025},
      eprint={2504.10458},
      archivePrefix={arXiv},
      primaryClass={cs.CV},
      url={https://arxiv.org/abs/2504.10458}, 
}

@article{intro:salary,
title={Rising between-workplace inequalities in high-income countries},
author = {Donald Tomaskovic-Devey  and Anthony Rainey  and Dustin Avent-Holt  and Nina Bandelj  and István Boza  and David Cort  and Olivier Godechot  and Gergely Hajdu  and Martin Hällsten  and Lasse Folke Henriksen  and others},
journal = {Proceedings of the National Academy of Sciences},
volume = {117},
number = {17},
pages = {9277-9283},
year = {2020},
doi = {10.1073/pnas.1918249117}}

@misc{intro:qwen,
  title={Qwen2-VL: Enhancing Vision-Language Model's Perception of the World at Any Resolution},
  author={Wang, Peng and Bai, Shuai and Tan, Sinan and Wang, Shijie and Fan, Zhihao and Bai, Jinze and Chen, Keqin and Liu, Xuejing and Wang, Jialin and Ge, Wenbin and others},
  eprint={arXiv preprint arXiv:2409.12191},
  year={2024}
}

@misc{intro:deepseek,
  title={Deepseek-vl: towards real-world vision-language understanding},
  author={Lu, Haoyu and Liu, Wen and Zhang, Bo and Wang, Bingxuan and Dong, Kai and Liu, Bo and Sun, Jingxiang and Ren, Tongzheng and Li, Zhuoshu and Sun, Yaofeng and others},
  eprint={arXiv preprint arXiv:2403.05525},
  year={2024}
}

@misc{intro:internlm,
      title={InternLM-XComposer2: Mastering Free-form Text-Image Composition and Comprehension in Vision-Language Large Model},
      author={Xiaoyi Dong and Pan Zhang and Yuhang Zang and Yuhang Cao and Bin Wang and Linke Ouyang and Xilin Wei and Songyang Zhang and Haodong Duan and Maosong Cao and others},
      eprint={arXiv preprint arXiv:2401.16420},
      year={2024}
}

@InProceedings{intro:internvl,
    author    = {Chen, Zhe and Wu, Jiannan and Wang, Wenhai and Su, Weijie and Chen, Guo and Xing, Sen and Zhong, Muyan and Zhang, Qinglong and Zhu, Xizhou and Lu, Lewei and others},
    title     = {InternVL: Scaling up Vision Foundation Models and Aligning for Generic Visual-Linguistic Tasks},
    booktitle = {IEEE/CVF Conference on Computer Vision and Pattern Recognition},
    month     = {June},
    year      = {2024},
    pages     = {24185-24198}
}

@misc{intro:dalle,
      title={Hierarchical Text-Conditional Image Generation with CLIP Latents}, 
      author={Aditya Ramesh and Prafulla Dhariwal and Alex Nichol and Casey Chu and Mark Chen},
      year={2022},
      eprint={2204.06125},
      archivePrefix={arXiv},
      primaryClass={cs.CV}
}

@misc{intro:xl,
      title={Text-Guided Synthesis of Artistic Images with Retrieval-Augmented Diffusion Models}, 
      author={Robin Rombach and Andreas Blattmann and Björn Ommer},
      year={2022},
      eprint={2207.13038},
      archivePrefix={arXiv},
      primaryClass={cs.CV}
}

@misc{intro:sora,
  title={Open-Sora: Democratizing Efficient Video Production for All},
  author={Zangwei Zheng and Xiangyu Peng and Tianji Yang and Chenhui Shen and Shenggui Li and Hongxin Liu and Yukun Zhou and Tianyi Li and Yang You},
  archivePrefix={ArXiv},
  year={2024},
  eprint={2412.20404},
  url={https://api.semanticscholar.org/CorpusID:275133398}
}

@misc{eda:chipnemo,
  title={ChipNeMo: Domain-Adapted LLMs for Chip Design},
  author={Liu, Mingjie and Ene, Teodor-Dumitru and Kirby, Robert and Cheng, Chris and Pinckney, Nathaniel and Liang, Rongjian and Alben, Jonah and Anand, Himyanshu and Banerjee, Sanmitra and Bayraktaroglu, Ismet and others},
  year={2023},
  eprint={2311.00176},
  archivePrefix={arXiv},
  primaryClass={cs.CE}
}

@inproceedings{eda:rtllm,
  title={{RTLLM}: An Open-Source Benchmark for Design {RTL} Generation with Large Language Model},
  author={Lu, Yao and Liu, Shang and Zhang, Qijun and Xie, Zhiyao},
  booktitle={Proceedings of the IEEE/ACM Asia and South Pacific Design Automation Conference},
  pages={473--478},
  year={2023}
}

@inproceedings{eda:verilogeval,
  title={VerilogEval: Evaluating Large Language Models for Verilog Code Generation},
  author={Liu, Mingjie and Pinckney, Nathaniel and Khailany, Brucek and Ren, Haoxing},
  booktitle={Proceedings of the IEEE/ACM International Conference on Computer-Aided Design},
  pages={1--8},
  year={2023}
}

@misc{eda:chipgpt,
  title={ChipGPT: How Far Are We from Natural Language Hardware Design?},
  author={Chang, Kaiyan and Wang, Ying and Ren, Haimeng and Wang, Mengdi and Liang, Shengwen and Han, Yinhe and Li, Huawei and Li, Xiaowei},
  year={2023},
  eprint={2305.14019},
  archivePrefix={arXiv},
  primaryClass={cs.HC}
}

@inproceedings{eda:2023verigen,
  title={VeriGen: A Large Language Model for Verilog Code Generation},
  author={Thakur, Shailja and Ahmad, Baleegh and Pearce, Hammond and Tan, Benjamin and Dolan-Gavitt, Brendan and Karri, Ramesh and Garg, Siddharth},
  booktitle={Proceedings of the Design, Automation and Test in Europe Conference and Exhibition},
  pages={1--6},
  year={2023}
}

@article{eda:chateda,
  title={ChatEDA: A Large Language Model Powered Autonomous Agent for {EDA}},
  author={Wu, Haoyuan and He, Zhuolun and Zhang, Xinyun and Yao, Xufeng and Zheng, Su and Zheng, Haisheng and Yu, Bei},
  journal={IEEE Transactions on Computer-Aided Design of Integrated Circuits and Systems},
  year={2024},
  volume={43},
  number={6},
  pages={1045--1058}
}

@inproceedings{eda:autochip,
  title={AutoChip: Automating {HDL} Generation Using {LLM} Feedback},
  author={Thakur, Shailja and Blocklove, Jason and Pearce, Hammond and Tan, Benjamin and Garg, Siddharth and Karri, Ramesh},
  booktitle={Proceedings of the 61st ACM/IEEE Design Automation Conference},
  pages={1--6},
  year={2024}
}

@inproceedings{eda:rtlfixer,
  title={{RTLFixer}: Automatically Fixing {RTL} Syntax Errors with Large Language Models},
  author={Tsai, Yun-Da and Liu, Mingjie and Ren, Haoxing},
  booktitle={Proceedings of the 61st ACM/IEEE Design Automation Conference},
  pages={1--6},
  year={2024}
}

@inproceedings{eda:verilogreader,
  title={VerilogReader: {LLM}-Aided Hardware Test Generation},
  author={Ma, Ruiyang and Yang, Yuxin and Liu, Ziqian and Zhang, Jiaxi and Li, Min and Huang, Junhua and Luo, Guojie},
  booktitle={Proceedings of the Workshop on Languages, Tools, and Automation for Hardware Design},
  pages={1--6},
  year={2024}
}

@misc{mllm:qwen25vl,
      title={Qwen2.5-VL Technical Report}, 
      author={Shuai Bai and Keqin Chen and Xuejing Liu and Jialin Wang and Wenbin Ge and Sibo Song and Kai Dang and Peng Wang and Shijie Wang and Jun Tang and others},
      year={2025},
      eprint={2502.13923},
      archivePrefix={arXiv},
      primaryClass={cs.CV},
      url={https://arxiv.org/abs/2502.13923}, 
}

@misc{mllm:qwen2vl,
      title={Qwen2-VL: Enhancing Vision-Language Model's Perception of the World at Any Resolution}, 
      author={Peng Wang and Shuai Bai and Sinan Tan and Shijie Wang and Zhihao Fan and Jinze Bai and Keqin Chen and Xuejing Liu and Jialin Wang and Wenbin Ge and others},
      year={2024},
      eprint={2409.12191},
      archivePrefix={arXiv},
      primaryClass={cs.CV},
      url={https://arxiv.org/abs/2409.12191}, 
}

@misc{mllm:internvl25,
      title={Expanding Performance Boundaries of Open-Source Multimodal Models with Model, Data, and Test-Time Scaling}, 
      author={Zhe Chen and Weiyun Wang and Yue Cao and Yangzhou Liu and Zhangwei Gao and Erfei Cui and Jinguo Zhu and Shenglong Ye and Hao Tian and Zhaoyang Liu and others},
      year={2025},
      eprint={2412.05271},
      archivePrefix={arXiv},
      primaryClass={cs.CV},
      url={https://arxiv.org/abs/2412.05271}, 
}

@article{mllm:internvl2,
  title={How far are we to gpt-4v? closing the gap to commercial multimodal models with open-source suites},
  author={Chen, Zhe and Wang, Weiyun and Tian, Hao and Ye, Shenglong and Gao, Zhangwei and Cui, Erfei and Tong, Wenwen and Hu, Kongzhi and Luo, Jiapeng and Ma, Zheng and others},
  journal={Science China Information Sciences},
  volume={67},
  number={12},
  pages={220101},
  year={2024},
  publisher={Springer}
}

@misc{mllm:ovis2,
      title={Ovis: Structural Embedding Alignment for Multimodal Large Language Model}, 
      author={Shiyin Lu and Yang Li and Qing-Guo Chen and Zhao Xu and Weihua Luo and Kaifu Zhang and Han-Jia Ye},
      year={2024},
      eprint={2405.20797},
      archivePrefix={arXiv},
      primaryClass={cs.CV},
      url={https://arxiv.org/abs/2405.20797}, 
}

@article{mllm:llavao,
title={{LL}a{VA}-OneVision: Easy Visual Task Transfer},
author={Bo Li and Yuanhan Zhang and Dong Guo and Renrui Zhang and Feng Li and Hao Zhang and Kaichen Zhang and Peiyuan Zhang and Yanwei Li and Ziwei Liu and Chunyuan Li},
journal={Transactions on Machine Learning Research},
issn={2835-8856},
year={2025},
url={https://openreview.net/forum?id=zKv8qULV6n},
pages = {1--44},
volume = {1},
}

@misc{mllm:nvlm,
      title={NVLM: Open Frontier-Class Multimodal LLMs}, 
      author={Wenliang Dai and Nayeon Lee and Boxin Wang and Zhuolin Yang and Zihan Liu and Jon Barker and Tuomas Rintamaki and Mohammad Shoeybi and Bryan Catanzaro and Wei Ping},
      year={2024},
      eprint={2409.11402},
      archivePrefix={arXiv},
      primaryClass={cs.CL},
      url={https://arxiv.org/abs/2409.11402}, 
}

@misc{mllm:gemini,
      title={Gemini 2.5: Pushing the Frontier with Advanced Reasoning, Multimodality, Long Context, and Next Generation Agentic Capabilities}, 
      author={Gheorghe Comanici and Eric Bieber and Mike Schaekermann and Ice Pasupat and Noveen Sachdeva and Inderjit Dhillon and Marcel Blistein and Ori Ram and Dan Zhang and Evan Rosen and others},
      year={2025},
      eprint={2507.06261},
      archivePrefix={arXiv},
      primaryClass={cs.CL},
      url={https://arxiv.org/abs/2507.06261}, 
}

@misc{mllm:gpt4o,
      title={GPT-4o System Card}, 
      author={OpenAI and Aaron Hurst and Adam Lerer and Adam P. Goucher and Adam Perelman and Aditya Ramesh and Aidan Clark and AJ Ostrow and Akila Welihinda and Alan Hayes and Alec Radford and others},
      year={2024},
      eprint={2410.21276},
      archivePrefix={arXiv},
      primaryClass={cs.CL},
      url={https://arxiv.org/abs/2410.21276}, 
}

@misc{mllm:llama3,
      title={The Llama 3 Herd of Models}, 
      author={Aaron Grattafiori and Abhimanyu Dubey and Abhinav Jauhri and Abhinav Pandey and Abhishek Kadian and Ahmad Al-Dahle and Aiesha Letman and Akhil Mathur and Alan Schelten and Alex Vaughan and others},
      year={2024},
      eprint={2407.21783},
      archivePrefix={arXiv},
      primaryClass={cs.AI},
      url={https://arxiv.org/abs/2407.21783}, 
}

@misc{mllm:phi35,
      title={Phi-3 Technical Report: A Highly Capable Language Model Locally on Your Phone}, 
      author={Marah Abdin and Jyoti Aneja and Hany Awadalla and Ahmed Awadallah and Ammar Ahmad Awan and Nguyen Bach and Amit Bahree and Arash Bakhtiari and Jianmin Bao and Harkirat Behl and others},
      year={2024},
      eprint={2404.14219},
      archivePrefix={arXiv},
      primaryClass={cs.CL},
      url={https://arxiv.org/abs/2404.14219}, 
}

@misc{mllm:claude,
  title={The Claude 3 Model Family: Opus, Sonnet, Haiku},
  author={Anthropic},
  year={2025},
  url={https://www-cdn.anthropic.com/de8ba9b01c9ab7cbabf5c33b80b7bbc618857627/Model_Card_Claude_3.pdf}
}

@misc{mllm:mplugowl3,
      title={mPLUG-Owl3: Towards Long Image-Sequence Understanding in Multi-Modal Large Language Models}, 
      author={Jiabo Ye and Haiyang Xu and Haowei Liu and Anwen Hu and Ming Yan and Qi Qian and Ji Zhang and Fei Huang and Jingren Zhou},
      year={2024},
      eprint={2408.04840},
      archivePrefix={arXiv},
      primaryClass={cs.CV},
      url={https://arxiv.org/abs/2408.04840}, 
}

@misc{mllm:janus,
      title={Janus-Pro: Unified Multimodal Understanding and Generation with Data and Model Scaling}, 
      author={Xiaokang Chen and Zhiyu Wu and Xingchao Liu and Zizheng Pan and Wen Liu and Zhenda Xie and Xingkai Yu and Chong Ruan},
      year={2025},
      eprint={2501.17811},
      archivePrefix={arXiv},
      primaryClass={cs.AI},
      url={https://arxiv.org/abs/2501.17811}, 
}

@misc{gui:aguvis,
      title={Aguvis: Unified Pure Vision Agents for Autonomous GUI Interaction}, 
      author={Yiheng Xu and Zekun Wang and Junli Wang and Dunjie Lu and Tianbao Xie and Amrita Saha and Doyen Sahoo and Tao Yu and Caiming Xiong},
      year={2025},
      eprint={2412.04454},
      archivePrefix={arXiv},
      primaryClass={cs.CL},
      url={https://arxiv.org/abs/2412.04454}, 
}

@inproceedings{gui:osatlas,
  title={OS-ATLAS: Foundation Action Model for Generalist GUI Agents},
  author={Wu, Zhiyong and Wu, Zhenyu and Xu, Fangzhi and Wang, Yian and Sun, Qiushi and Jia, Chengyou and Cheng, Kanzhi and Ding, Zichen and Chen, Liheng and Liang, Paul Pu and others},
  booktitle={The Thirteenth International Conference on Learning Representations},
  year={2025}
}

@misc{gui:uitars,
      title={UI-TARS: Pioneering Automated GUI Interaction with Native Agents}, 
      author={Yujia Qin and Yining Ye and Junjie Fang and Haoming Wang and Shihao Liang and Shizuo Tian and Junda Zhang and Jiahao Li and Yunxin Li and Shijue Huang and others},
      year={2025},
      eprint={2501.12326},
      archivePrefix={arXiv},
      primaryClass={cs.AI},
      url={https://arxiv.org/abs/2501.12326}, 
}

@inproceedings{gui:cogagent,
  title={Cogagent: A visual language model for gui agents},
  author={Hong, Wenyi and Wang, Weihan and Lv, Qingsong and Xu, Jiazheng and Yu, Wenmeng and Ji, Junhui and Wang, Yan and Wang, Zihan and Dong, Yuxiao and Ding, Ming and others},
  booktitle={Proceedings of the IEEE/CVF Conference on Computer Vision and Pattern Recognition},
  pages={14281--14290},
  year={2024}
}

@inproceedings{gui:aria,
  title={Aria-UI: Visual Grounding for GUI Instructions},
  author={Yang, Yuhao and Wang, Yue and Li, Dongxu and Luo, Ziyang and Chen, Bei and Huang, Chao and Li, Junnan},
  booktitle={ICLR 2025 Workshop on Foundation Models in the Wild},
  year={2025}
}

@inproceedings{gui:showui,
  title={Showui: One vision-language-action model for gui visual agent},
  author={Lin, Kevin Qinghong and Li, Linjie and Gao, Difei and Yang, Zhengyuan and Wu, Shiwei and Bai, Zechen and Lei, Stan Weixian and Wang, Lijuan and Shou, Mike Zheng},
  booktitle={Proceedings of the Computer Vision and Pattern Recognition Conference},
  pages={19498--19508},
  year={2025}
}

@inproceedings{gui:seeclick,
  title={SeeClick: Harnessing GUI Grounding for Advanced Visual GUI Agents},
  author={Cheng, Kanzhi and Sun, Qiushi and Chu, Yougang and Xu, Fangzhi and YanTao, Li and Zhang, Jianbing and Wu, Zhiyong},
  booktitle={Proceedings of the 62nd Annual Meeting of the Association for Computational Linguistics (Volume 1: Long Papers)},
  pages={9313--9332},
  year={2024}
}

@misc{gui:osgenesis,
      title={OS-Genesis: Automating GUI Agent Trajectory Construction via Reverse Task Synthesis}, 
      author={Qiushi Sun and Kanzhi Cheng and Zichen Ding and Chuanyang Jin and Yian Wang and Fangzhi Xu and Zhenyu Wu and Chengyou Jia and Liheng Chen and Zhoumianze Liu and others},
      year={2025},
      eprint={2412.19723},
      archivePrefix={arXiv},
      primaryClass={cs.AI},
      url={https://arxiv.org/abs/2412.19723}, 
}

@inproceedings{bench:mmbench,
  title={Mmbench: Is your multi-modal model an all-around player?},
  author={Liu, Yuan and Duan, Haodong and Zhang, Yuanhan and Li, Bo and Zhang, Songyang and Zhao, Wangbo and Yuan, Yike and Wang, Jiaqi and He, Conghui and Liu, Ziwei and others},
  booktitle={European conference on computer vision},
  pages={216--233},
  year={2024},
  organization={Springer}
}

@inproceedings{bench:abench,
  title={A-Bench: Are LMMs Masters at Evaluating AI-generated Images?},
  author={Zhang, Zicheng and Wu, Haoning and Li, Chunyi and Zhou, Yingjie and Sun, Wei and Min, Xiongkuo and Chen, Zijian and Liu, Xiaohong and Lin, Weisi and Zhai, Guangtao},
  booktitle={The Thirteenth International Conference on Learning Representations},
  pages={1--22},
  year={2025}
}

@ARTICLE{bench:rbench,
  author={Li, Chunyi and Zhang, Jianbo and Zhang, Zicheng and Wu, Haoning and Tian, Yuan and Sun, Wei and Lu, Guo and Min, Xiongkuo and Liu, Xiaohong and Lin, Weisi and others},
  journal={IEEE Journal of Selected Topics in Signal Processing}, 
  title={R-Bench: Are your Large Multimodal Model Robust to Real-world Corruptions?}, 
  year={2025},
  pages={1-16},
  doi={10.1109/JSTSP.2025.3558652}}

@inproceedings{bench:cmcbench,
      title={Towards a Cross-Modality Paradigm of Visual Signal Compression}, 
      author={Chunyi Li and Xiele Wu and Haoning Wu and Donghui Feng and Zicheng Zhang and Guo Lu and Xiongkuo Min and Xiaohong Liu and Guangtao Zhai and Weisi Lin},
      year={2025},
      booktitle={ACM International Conference on Multimedia},
}

@inproceedings{bench:info,
  title={Information Density Principle for MLLM Benchmarks},
  author={Li, Chunyi and Li, Xiaozhe and Zhang, Zicheng and Tian, Yuan and Jia, Ziheng and Liu, Xiaohong and Min, Xiongkuo and Wang, Jia and Duan, Haodong and Chen, Kai and others},
  booktitle={Proceedings of the International Conference on Computer Vision},
  pages={4167-4177},
  year={2025}
}

@article{bench:AIBench,
  title   = {AIBench: Towards trustworthy evaluation under the 45° law},
  author  = {Zicheng Zhang and Junying Wang and Yijin Guo and Farong Wen and Zijian Chen and Hanqing Wang and Wenzhe Li and Lu Sun and Yingjie Zhou and Jianbo Zhang and others},
  journal = {Displays},
  year    = {2025},
  pages   = {103255},
  issn    = {0141-9382},
  doi     = {10.1016/j.displa.2025.103255}
}

@article{bench:large,
  author    = {Zhang, Zicheng and Wang, Junying and Wen, Farong and Guo, Yijin and Zhao, Xiangyu and Fang, Xinyu and Ding, Shengyuan and Jia, Ziheng and Xiao, Jiahao and Shen, Ye and others},
  title     = {Large Multimodal Models Evaluation: A Survey},
  journal   = {SCIENCE CHINA Information Sciences},
  year      = {2025},
  number  = {12},
  volume    = {68},
  pages     = {221301-221369},
  doi       = {https://doi.org/10.1007/s11432-025-4676-4}
}

@article{old:instructblip,
  title={Instructblip: Towards general-purpose vision-language models with instruction tuning},
  author={Dai, Wenliang and Li, Junnan and Li, Dongxu and Tiong, Anthony and Zhao, Junqi and Wang, Weisheng and Li, Boyang and Fung, Pascale N and Hoi, Steven},
  journal={Advances in neural information processing systems},
  volume={36},
  pages={49250--49267},
  year={2023}
}

@inproceedings{old:visualglm,
  title={GLM: General Language Model Pretraining with Autoregressive Blank Infilling},
  author={Du, Zhengxiao and Qian, Yujie and Liu, Xiao and Ding, Ming and Qiu, Jiezhong and Yang, Zhilin and Tang, Jie},
  booktitle={Proceedings of the 60th Annual Meeting of the Association for Computational Linguistics (Volume 1: Long Papers)},
  pages={320--335},
  year={2022}
}

\end{document}